\documentclass[10pt]{article} 
\usepackage[accepted]{tmlr}

\usepackage{amsmath,amsfonts,bm}

\def\eqref#1{equation~\ref{#1}}

\def\1{\bm{1}}

\DeclareMathAlphabet{\mathsfit}{\encodingdefault}{\sfdefault}{m}{sl}
\SetMathAlphabet{\mathsfit}{bold}{\encodingdefault}{\sfdefault}{bx}{n}

\usepackage[ruled]{algorithm2e}

\usepackage[utf8]{inputenc} 
\usepackage[T1]{fontenc}    
\usepackage{hyperref}       
\usepackage{url}            
\usepackage{booktabs}       
\usepackage{amsfonts}       
\usepackage{nicefrac}       
\usepackage{microtype}      
\usepackage{xcolor}         

\usepackage{amsmath}
\usepackage{mathtools}
\usepackage{subfigure}
\usepackage{graphicx}
\usepackage{caption}
\usepackage{multirow}
\usepackage{rotating}
\usepackage{float}
\usepackage{wrapfig,lipsum,booktabs}

\newcommand\csm[1]{\textcolor{black}{#1}}
\newcommand\iclr[1]{\textcolor{black}{#1}}

\newcommand\tmlrcsm[1]{\textcolor{black}{#1}}

\title{Hyperparameters in Continual Learning: A Reality Check}

\author{\name Sungmin Cha\thanks{Code is available at: \url{https://github.com/csm9493/GTEP}} \email sungmin.cha@nyu.edu \\
      \addr New York University
      \AND
      \name Kyunghyun Cho \email kyunghyun.cho@nyu.edu\\
\addr New York University \& Genentech}

\begin{document}

\maketitle

\begin{abstract}
Continual learning (CL) aims to train a model on a sequence of tasks (\textit{i.e.}, a CL scenario) while balancing the trade-off between plasticity (learning new tasks) and stability (retaining prior knowledge). The dominantly adopted conventional evaluation protocol for CL algorithms selects the best hyperparameters \tmlrcsm{(\textit{e.g.}, learning rate, mini-batch size, regularization strengths,  etc.)} within a given scenario and then evaluates the algorithms
using these hyperparameters in the same scenario. However, this protocol has significant shortcomings: it overestimates the CL capacity of algorithms and relies on unrealistic hyperparameter tuning, which is not feasible for real-world applications.
From the fundamental principles of evaluation in machine learning, 
we argue that the evaluation of CL algorithms should focus on assessing the generalizability of their CL capacity to unseen scenarios. 
Based on this, we propose the Generalizable Two-phase Evaluation Protocol (GTEP) consisting of hyperparameter tuning and evaluation phases. Both phases share the same scenario configuration (\textit{e.g.}, number of tasks) but are generated from different datasets. Hyperparameters of CL algorithms are tuned in the first phase and applied in the second phase to evaluate the algorithms.
We apply this protocol to class-incremental learning, both with and without pretrained models. Across more than 8,000 experiments, our results show that most state-of-the-art algorithms fail to replicate their reported performance, highlighting that their CL capacity has been significantly overestimated in the conventional evaluation protocol.

\end{abstract}

\section{Introduction}

\begin{figure}[!t]
    \centering
    \includegraphics[width=.98\textwidth]{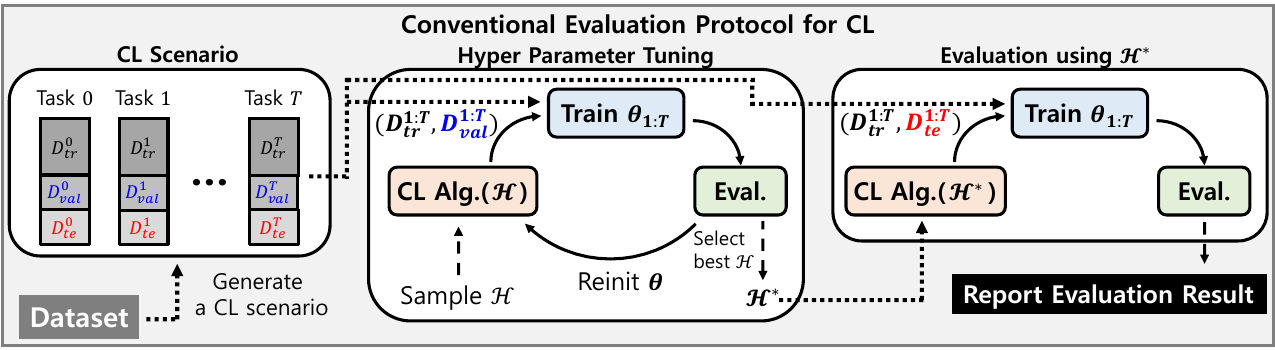}
    \caption{This figure illustrates the conventional evaluation protocol. First, a CL scenario is constructed using a benchmark dataset, where each task has its own training, validation, and test sets. Second, to find the best hyperparameters $\mathcal{H}^*$, a model is sequentially trained up to the final task using  sampled hyperparameters. After training for each task $t$, the model $\theta_t$ is evaluated using the validation dataset. This process is repeated across various hyperparameter settings, and the best hyperparameters $\mathcal{H}^*$ are selected based on a performance metric. Finally, a new model is trained using the CL algorithm with the best hyperparameters $\mathcal{H}^*$ in the same CL scenario, and report the evaluation result on the test dataset. Note that in most studies, results are reported using only $D_{\text{val}}$, without a separate test set (\textit{i.e.}, $D_{\text{te}} = D_{\text{val}}$)~\citep{zhou2023pycil,sun2023pilot}.}
    \label{fig:hp_tuning_intro}
    \vspace{-.2in}
\end{figure}

In recent years, extensive research has been conducted on continual learning (CL) with the goal of effectively learning knowledge from a sequence of tasks~\citep{wang2023comprehensive}. 
\csm{A} neural network model in such CL scenarios faces a crucial trade-off between learning new knowledge from novel tasks (plasticity) and maintaining knowledge on previous tasks (stability)~\citep{(tradeoff)mermillod2013stability}. To address this inherent trade-off, numerous algorithms have been proposed for successful CL in various domains~\citep{wang2023comprehensive}. In these domains, \csm{many CL studies have} focused on classification, 
primarily concentrating on class-incremental learning (class-IL) ~\citep{(cil_survey)masana2020class} without or with pretrained models~\citep{(cil_pre)zhou2024continual}.
However, deploying CL algorithms requires careful hyperparameter tuning. 
Figure \ref{fig:hp_tuning_intro} illustrates the conventional evaluation protocol, including hyperparameter \tmlrcsm{(\textit{e.g.}, learning rate, mini-batch size,  etc.)} tuning, dominantly employed in both offline and online class-incremental learning~\citep{(memo)zhou2022model,boschini2022class,(ease)zhou2024expandable,(coda)smith2023coda,seo2024learning}.
Moreover, similar evaluation protocols have been extensively adopted in other areas of CL, including semantic segmentation~\citep{cha2021ssul,yuan2024survey}, test-time adaptation~\citep{yoo2024and,leebecotta}, federated learning~\citep{piaofederated}, self-supervised learning~\citep{fini2022self,charegularizing}, and large language models~\citep{ke2023continual,wu2024continual}.

Many algorithms have been considered state-of-the-art based on performance validated through the conventional evaluation protocol. However, this raises two issues: First, the hyperparameter tuning method used in this protocol is not applicable to real-world CL scenarios. Second, it results in evaluation overfitting to a given scenario and dataset, which in turn leads to an overestimation of their CL capacity.
In other words, this protocol only assesses performance in a seen scenario but fails to evaluate generalizability to new, unseen ones—an essential aspect for real-world applications. While several alternative evaluation protocols and hyperparameter tuning methods have been proposed, they also have limitations: 1) they require to tune additional hyperparameters for their methods~\citep{(cl_survey2)delange2021continual, liu2023online}, or 2) they are only applied to a few old algorithms, and have not gained widespread acceptance~\citep{(agem)chaudhry2018efficient,(eval1)chen2023forgetting,(eval3)bornschein2023nevis}. As a result, the issues with the conventional evaluation protocol have been largely ignored, and it remains the dominant evaluation protocol for evaluating CL algorithms until now.

In this paper, we aim to reveal the limitation of the conventional evaluation protocol by revisiting the fundamental principles of evaluation in machine learning. 
From this perspective, we argue that the evaluation of CL algorithms should prioritize assessing the generalizability of each algorithm's CL capacity across unseen scenarios. 
\begin{wrapfigure}{r}{0.45\textwidth}
\vspace{-0.1in}
\centering
\includegraphics[width=0.4\textwidth]{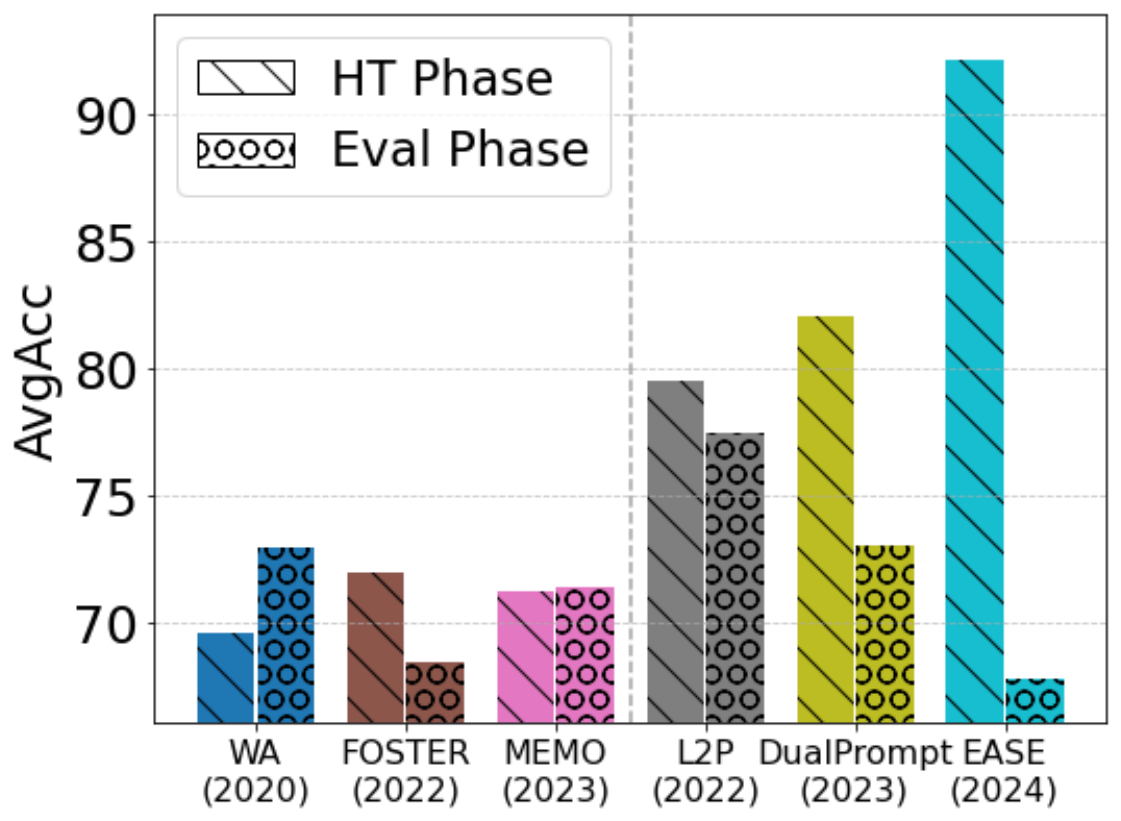}
\vspace{-0.15in}
  \caption{\iclr{Results on both phases.}}\label{figure:intro_figure}
  \vspace{-0.2in}
\end{wrapfigure}
\tmlrcsm{To address this issue, we propose a revised evaluation protocol—the Generalizable Two-phase Evaluation Protocol (GTEP)—which consists of two distinct phases: a hyperparameter tuning phase and an evaluation phase. The core motivation behind GTEP is to assess the generalizability of CL algorithms by first identifying hyperparameters in a seen scenario that closely resembles an unseen scenario, and then evaluating the algorithm's performance in the actual unseen scenario.
Both phases share the same CL scenario configuration (e.g., number of tasks and classes per task), but use disjoint datasets to enable a more realistic assessment. During the hyperparameter tuning phase, models are trained under various hyperparameter configurations, and the best-performing set is selected. These best hyperparameters are then applied in the evaluation phase, where performance is measured on a separate dataset, serving as a reliable indicator of the algorithm’s capacity to generalize in unseen scenarios
As an initial application of GTEP, we focus on the most actively studied setting in CL—class-incremental learning (class-IL)—evaluating algorithms both with and without pretrained models~\citep{wang2023comprehensive}. Based on approximately 8,000 experiments, we report the following key findings:}
\begin{itemize}
  \item {\iclr{First, as illustrated in Figure~\ref{figure:intro_figure}, most state-of-the-art class-IL algorithms achieve strong performance during the hyperparameter tuning phase, which mirrors the conventional evaluation protocol. However, some algorithms exhibit limited generalizability in their CL capacity when evaluated on unseen scenarios (\textit{i.e.}, the evaluation phase). This issue is particularly pronounced in recent algorithms.}}
  \item {Second, further analysis reveals that many of these algorithms suffer from long training times, substantial parameter requirements, or high performance variance across different CL scenarios, indicating that they are less efficient than previously assumed.}
\end{itemize}

Our extensive experimental results using the proposed evaluation protocol expose critical flaws in the conventional evaluation paradigm, which systematically inflates the perceived effectiveness of CL algorithms. These findings call for a fundamental shift in how CL methods are assessed. We urge the community to adopt more rigorous and generalizable evaluation standards across domains to enable truly robust and scalable CL. In this context, we propose our GTEP protocol as a minimal yet essential step toward achieving this goal.    
\section{Related Work}
\noindent{\textbf{Continual learning}} \ \
Continual learning (CL) research has been conducted in various domains~\citep{wang2023comprehensive,(cl_survey1)parisi2019continual,(cl_survey2)delange2021continual,(cil_survey)masana2020class}. In the beginning, \csm{the} CL research focus on task-incremental learning~\citep{(cl_survey1)parisi2019continual,(cl_survey2)delange2021continual}, exploring diverse approaches~\citep{(lwf)li2017learning, (MAS)aljundi2018memory, (Rwalk)chaudhry2018riemannian, (cpr)cha2021cpr, (den)yoon2017lifelong}. 
As the field progressed, attention shifted to the more challenging scenario, class-incremental learning (class-IL)~\citep{(cil_survey)masana2020class}. This shift leads to the investigation of exemplar-based methods, involving the effective utilization of exemplar memory storing a subset of the dataset from previous tasks~\citep{(icarl)rebuffi2017icarl,(wa)zhao2020maintaining, (tbbn)cha2023rebalancing}. 
Since then, using the exemplar memory has become standard, with several methods building on this foundation. Regularization-based methods, which \tmlrcsm{mitigate} catastrophic forgetting by introducing a novel regularization~\citep{(bic)wu2019large,(podnet)douillard2020podnet}\csm{, and m}odel expansion-based methods, which dynamically expand model capacity to balance the trade-off between  stability and plasticity, have become the most powerful approach, achieving state-of-the-art performance~\citep{(foster)wang2022foster,(der)yan2021dynamically,(memo)zhou2022model,wang2022beef}.

Class-IL using pretrained models has recently gained considerable attention for achieving strong performance without relying on the exemplar memory~\citep{(cil_pre)zhou2024continual}. Prompt-based methods enable class-IL through prompt learning while keeping the pretrained model frozen. These approaches have evolved over time, incorporating techniques such as using prompt pool~\citep{(l2p)wang2022learning}, prompt combination~\citep{(dual)wang2022dualprompt}, decomposed prompt~\citep{(coda)smith2023coda}, and prompt generation~\citep{(gen)jung2023generating}.  Additionally, representation-based methods derive class prototypes from the pretrained model and use them for classification~\citep{(adam)zhou2023revisiting}. To enhance the separability of these prototypes, several recent methods have focused on reducing class-wise correlation~\citep{(ranpac)mcdonnell2024ranpac, (ease)zhou2024expandable}.

\noindent{\textbf{Evaluation and hyperparameter tuning of CL}} \ \
Several papers have proposed new evaluation metrics and protocols for the proper assessment of CL algorithms in classification.
Traditionally, accuracy-based metrics (e.g., final and average accuracy) have been used as the primary metrics of evaluating performance of CL algorithms~\citep{(cl_survey1)parisi2019continual,(cil_survey)masana2020class, (Rwalk)chaudhry2018riemannian}. However, recent studies have highlighted limitations of these metrics, particularly regarding computational costs~\citep{prabhu2023computationally} and learned representations~\citep{cha2023objective}.
\cite{(cl_survey2)delange2021continual} introduced a hyperparameter tuning method for task-incremental learning, which involves first conducting a maximum plasticity search and then selecting the best hyperparameters using stability decay. 
Similarly, \cite{liu2023online} proposed a hyperparameter selection method for class-IL based on a bandit algorithm. However, both approaches entail additional training costs and the need to tune extra hyperparameters.

\tmlrcsm{Several prior studies have proposed evaluation protocols similar in spirit to ours~\citep{(agem)chaudhry2018efficient,(eval1)chen2023forgetting,(eval3)bornschein2023nevis, michel2023rethinking, wang2024improving}, incorporating a separate hyperparameter tuning phase. However, while such approaches have advanced evaluation practices in specific contexts, these protocols have predominantly been applied to a limited set of earlier algorithms within narrowly defined domains. 
In particular, they fail to comprehensively address the issues of the hyperparameter tuning in the face of dynamic, real-world conditions. 
Despite these efforts—and ongoing discussions about the need for more rigorous CL evaluation~\citep{mundt2022clevacompass}—the conventional evaluation protocol continues to dominate the assessment of state-of-the-art CL algorithms across a wide range of domains.}

We believe the continued use of the conventional evaluation protocol stems from a lack of awareness in the research community regarding its inherent flaws. In this context, our paper makes two distinct contributions: 1) we introduce a revised evaluation protocol specifically designed to assess the generalizability of each algorithm's CL capacity more accurately, and 2) unlike previous studies, our paper comprehensively exposes the shortcomings of the conventional evaluation protocol, supported by extensive experimental validation.



\section{A Protocol for Evaluating the Generalizability of Continual Learning Algorithms}\label{sec:protocol}

\subsection{Motivation: Limitations of hyperparameter tuning in conventional CL evaluation}

As shown in Figure \ref{fig:hp_tuning_intro}, the primary flaw of the conventional evaluation protocol is that it optimizes an algorithm's hyperparameters in a given CL scenario and then evaluates the algorithm using those hyperparameters in the same scenario.
Surprisingly, many studies have reported their results by directly tuning hyperparameters on validation data without considering a separate test set (\textit{i.e.}, set $D^{HT}_{te} = D^{HT}_{val}$), as seen in studies such as  \cite{(bic)wu2019large,(podnet)douillard2020podnet,(wa)zhao2020maintaining, (der)yan2021dynamically, (foster)wang2022foster, (memo)zhou2022model,wang2022beef, (l2p)wang2022learning,(adam)zhou2023revisiting,(ease)zhou2024expandable}, and others.
Note that this approach is only feasible in experimental scenarios where all task data is always available.
Consequently, this hyperparameter tuning method fails to capture the real challenges of CL and is not applicable to real-world situations. 
While many studies partially address this limitation by reporting robust performance across various experiments with some fixed or minimally adjusted hyperparameters~\citep{wang2022beef, (l2p)wang2022learning, (ease)zhou2024expandable}, these evaluations are still based on given scenarios (\textit{i.e.}, seen scenarios), making it challenging to assess whether the algorithms would perform equally well in unseen scenarios.
Nevertheless, this conventional protocol remains the predominant evaluation protocol for assessing algorithms across most CL domains.

%

\begin{figure*}[t]
\centering
\includegraphics[width=0.98\textwidth]{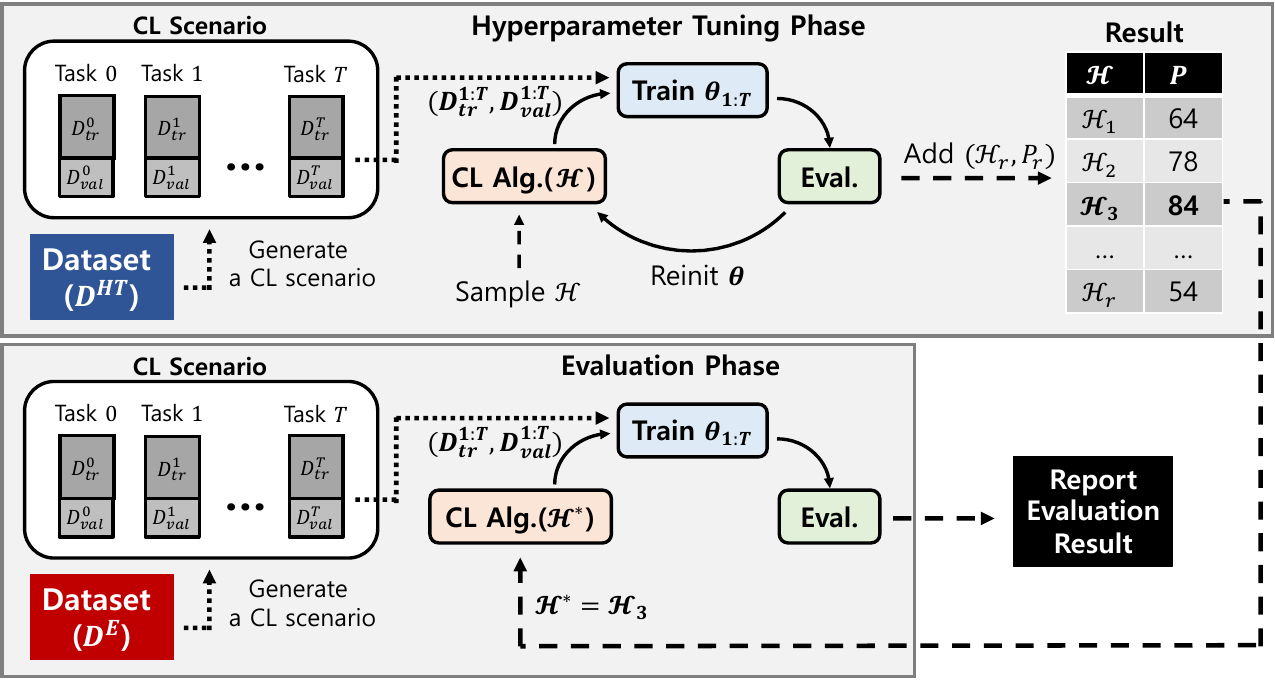}
  \vspace{-0.1in}
  \caption{Illustration of the proposed evaluation protocol. 
  Both phases share the same CL scenario configuration (\textit{e.g.}, the number of tasks and number of classes in each task) but they are generated from distinct datasets ($D^{HT}$ and $D^E$). 
  Best hyperparameters are selected in the hyperparameter tuning phase. Then, the evaluation phase access a CL algorithm by training a model \csm{using} them. \iclr{Note that evaluating an algorithm solely based on the results from the hyperparameter tuning phase is identical to the conventional evaluation protocol without using \tmlrcsm{$D^{E}$}.}}\label{figure:protocol}
  \vspace{-0.25in}
\end{figure*}

\begin{minipage}{\textwidth}
\begin{algorithm}[H]
\caption{\iclr{The Generalizable Two-phase Evaluation Protocol}}
\label{algorithm:protocol}
\textbf{Input} : A CL algorithm $\mathcal{A}$, a model $\theta$, the dataset for the hyperparameter tuning phase $D^{HT}$, the dataset for the evaluation phase  $D^{E}$, the number of random samplings $R$, the number of trials $S$, and the number of hyperparameters $K$.\\
\textbf{Output} : Final evaluation result ($P^{E}$) for a CL algorithm $\mathcal{A}$ in the evaluation phase\\
\begin{enumerate}
\item $\{(\mathcal{H}_i, P^{HT}_{i})\}^{R}_{i=1}$ $\leftarrow$ \texttt{HyperparameterTuning}($\theta, \mathcal{A}, D^{HT}, R, S, K$)
\item $\mathcal{H}^* \leftarrow $ \texttt{SelectBestHyperparameter}($\{(\mathcal{H}_i, P^{HT}_i)\}^{R}_{i=1}$)
\item $P^{E}$ $\leftarrow$ \texttt{Evaluation}($\theta, \mathcal{A}, D^{E}, \mathcal{H}^*, S$)
\end{enumerate}
\end{algorithm}
\vspace{-0.1in}
\end{minipage}

\subsection{\iclr{Generalizable} Two-phase Evaluation Protocol (GTEP) for CL evaluation}


Given the previously discussed issues with the conventional evaluation protocol, the key question becomes: What hyperparameter tuning and evaluation protocol should be used to properly assess CL algorithms? 
\iclr{Note that effective evaluation in machine learning should prioritize realistic methods tailored to each learning scenario, rather than rigidly adhering to assumptions (\textit{e.g.}, i.i.d.) for theoretical convenience. In this \tmlrcsm{regard}, }\tmlrcsm{we argue that evaluating the generalizability of each algorithm's CL capacity is essential.} For example, consider a real-world CL scenario where an algorithm is applied to a CL scenario consisting of a sequence of tasks. 
Since the entire task data would not be fully accessible at once, the conventional hyperparameter tuning method cannot be applied. 
In such cases, a reasonable approach is to construct a simulated CL scenario, reflecting the expected actual CL scenario, using a benchmark or available dataset. This involves identifying the best hyperparameters in the simulated scenario and then applying them to the actual CL scenario.
In other words, one of the basic evaluation protocols—consistent with the fundamental principles of evaluation in machine learning—is to tune hyperparameters in seen scenarios (\textit{e.g.}, simulated scenarios) and test them in unseen scenarios (\textit{e.g.}, actual scenarios). 


\begin{wrapfigure}{r}{8.cm}
\vspace{-0.15in}
\begin{minipage}{0.5\textwidth}
\begin{algorithm}[H]
\caption{Pseudo algorithm of the hyperparameter huning phase}
\label{algorithm:tune}
\textbf{Input} : A CL algorithm $\mathcal{A}$, a model $\theta$, the dataset for the hyperparameter tuning phase $D^{HT}$, the number of random samplings $R$, the number of trials $S$, the number of hyperparameters $K$, and the function that generates a CL scenario $\mathcal{F}$.\\
\textbf{Output} : $\{(\mathcal{H}_i, P^{HT}_i)\}^{R}_{i=1}$ \\
\begin{enumerate}
\item Result $\leftarrow {\{\}}$
\item \textbf{for} $r \leftarrow 1 \ \ to \ \ R \ \ \textbf{do}$
\item \ \ \ \ \textbf{for} $k \leftarrow 1 \ \ to \ \ K \ \ \textbf{do}$
\item \ \ \ \ \ \ \ \ \ $h_k \leftarrow \text{RandomSample}(h^{\text{Set}}_k)$
\item \ \ \ \ $\mathcal{H}_r$ $\leftarrow \tmlrcsm{(h_1, \cdots, h_K)}$
\item \ \ \ \ \textbf{for} $s \leftarrow 1 \ \ to \ \ S \ \ \textbf{do}$
\item \ \ \ \ \ \ \ \ Initialize $\theta$
\item \ \ \ \ \ \ \ \ $D^{HT}_{tr}, D^{HT}_{val} \leftarrow \mathcal{F}$(Shuffle($D^{HT}$))
\item \ \ \ \ \ \ \ \ $P^{HT}_s \leftarrow \text{TrainCL}$($\mathcal{A}, D^{HT}_{tr}, D^{HT}_{val}, \theta, \mathcal{H}_r$)
\item \ \ \ \ $P^{HT}_r \leftarrow \frac{1}{S}\sum^{S}_{s=1}{P^{HT}_s}$
\item \ \ \ \ Add $(\mathcal{H}_r, P^{HT}_r)$ to Result
\end{enumerate}
\end{algorithm}
\end{minipage}
\vspace{-0.3in}
\end{wrapfigure}
Building on the above concept, we propose
a revised evaluation protocol consisting of two phases, 
\iclr{the Generalizable Two-phase Evaluation Protocol (GTEP)}: hyperparameter tuning and evaluation. 
Figure \ref{figure:protocol} and Algorithm \ref{algorithm:protocol} outlines the overall process. The key idea is that CL scenarios for the hyperparameter tuning and evaluation phases are generated from different datasets (\textit{i.e.}, $D^{HT} \neq D^{E}$) but share the same scenario configuration (\textit{e.g.}, the number of tasks and classes per task), based on expectations on the actual scenario.
In the hyperparameter tuning phase, the goal is to identify the best hyperparameters for the CL algorithm. In the evaluation phase, these hyperparameters are applied to assess the algorithm's CL capacity in unseen scenarios, providing a more realistic measure of its generalizability.

The pseudo algorithm of the hyperparameter tuning phase is outlined in Algorithm \ref{algorithm:tune}. 
First, we randomly sample hyperparameters $h_k$ from a predefined set $h_k^{Set}$ and build a list of selected hyperparameters $\mathcal{H}_r$.
Next, we generate a predefined CL scenario using the function $\mathcal{F}$ with shuffled class orderings.
Afterward, the model $\theta$ is trained using the selected hyperparameters $\mathcal{H}_r$, the CL algorithm $\mathcal{A}$, and the training dataset $D^{HT}_{tr}$.
Performance ($P^{HT}$) is then measured on the validation dataset $D^{HT}_{val}$. This phase returns multiple sets of hyperparameters and their corresponding performance. 
Next, using the \texttt{SelectBestHyperparameter} function in Algorithm~\ref{algorithm:protocol}, we select the best hyperparameters, denoted as $\mathcal{H}^{*}$.
Note that the hyperparameter tuning phase is identical to the conventional evaluation protocol. However, we only use the results from this phase to select the best hyperparameters.

In the evaluation phase (shown in Algorithm \ref{algorithm:eval}), we train a model $\theta$ using the CL algorithm with the best hyperparameters $\mathcal{H}^{*}$. The trained model is then tested on the validation dataset $D^{E}_{val}$. The final performance metric is the averaged performance ($P^{E}$) of the trained model across multiple class orderings, which serves as the evaluation criterion for the CL algorithm.

\begin{minipage}{1.0\textwidth}
\begin{algorithm}[H]
\caption{Pseudo algorithm of the evaluation phase}
\label{algorithm:eval}

\textbf{Input} : A CL Algorithm $\mathcal{A}$, a model $\theta$, the dataset for the Eval phase $D^{E}$, the best hyperparameter value $\mathcal{H}^*$, the number of trials $S$, the number of hyperparameters $K$, and the function that generates a CL scenario $\mathcal{F}$.\\
\textbf{Output} : Final evaluation result ($P^{E}$) for $\mathcal{A}$ \\
\begin{enumerate}
\item \textbf{for} $s \leftarrow 1 \ \ to \ \ S \ \ \textbf{do}$
\item \ \ \ \ Initialize $\theta$
\item \ \ \ \ $D^{E}_{tr}, D^{E}_{val} \leftarrow \mathcal{F}$(Shuffle($D^{E}$))
\item \ \ \ \ $P^{E}_s \leftarrow \text{TrainCL}$($\mathcal{A}, D^{E}_{tr}, D^{E}_{val}, \theta, \mathcal{H}^*$)
\item $P^{E} \leftarrow \frac{1}{S}\sum^{S}_{s=1}{P^{E}_s}$
\end{enumerate}
\end{algorithm}
\end{minipage}

To find the best hyperparameters for each algorithm, we optimize both algorithm-specific hyperparameters (\textit{e.g.}, regularization strength) and general hyperparameters (\textit{e.g.}, learning rate and batch size). During the hyperparameter tuning phase, we train the model with $R$ sets of randomly selected hyperparameters  and account for $S$ task orderings per set. In the evaluation phase, we assess the performance across $S$ task orderings as well.
In this paper, we set $R=30$ and $S=5$ for all experiments.
We also take into account various similarity scenarios between the hyperparameter tuning dataset ($D^{HT}$) and the evaluation dataset ($D^{E}$). 
High similarity indicates that the characteristics of the dataset used in the actual scenario are somewhat predictable, allowing us to generate a scenario for the hyperparameter tuning phase using a similar dataset.   Conversely, low similarity suggests unpredictability, indicating that the datasets used to generate scenarios in both phases differ significantly.
\tmlrcsm{Unlike previous studies~\citep{(similarity1)standley2020tasks,(similarity2)fifty2021efficiently,(similarity3)zamir2018taskonomy}, we consider similarity in the simplest form at the dataset level. Specifically, high similarity refers to splitting the same dataset (\textit{e.g.}, CIFAR-100) into two disjoint subsets, which are then used in each phase. On the other hand, low similarity involves using completely different datasets in each phase.}
Evaluating each algorithm under both similarity cases offers a comprehensive understanding of the generalizability of its CL capacity. Furthermore, these efforts toward more rigorous evaluation underscore key methodological differences from previously proposed protocols~\citep{(agem)chaudhry2018efficient,(eval1)chen2023forgetting,(eval3)bornschein2023nevis,mundt2022clevacompass}, further motivating the need for a revised evaluation framework.
\iclr{Note that the high-level concept underlying our proposed protocol is broadly applicable across diverse CL domains. By adapting the CL scenario generation process—denoted as $\mathcal{F}$ in Algorithms~\ref{algorithm:tune} and~\ref{algorithm:eval}—to reflect domain-specific characteristics (e.g., class imbalance within tasks, blurred task boundaries, or alternative CL domains), the protocol can be effectively applied to a wide range of settings that involve CL scenarios.}

\section{Experimental Results}
\iclr{In this section, we present extensive experimental results using our proposed protocol within the most actively studied domain of continual learning (CL)—class-incremental learning (class-IL)~\citep{wang2023comprehensive}—considering both settings without and with pretrained models~\citep{(cil_survey)masana2020class,(cil_pre)zhou2024continual,zhou2023pycil}.
We begin by evaluating whether the best hyperparameters identified during the hyperparameter tuning phase outperform the original ones (\textit{i.e.}, the previously reported best hyperparameters for each algorithm). We then assess the continual learning capacity of each algorithm across different levels of scenario similarity and compare their cost efficiency.}


\subsection{Class-incremental learning without pretrained models}

\noindent{\textbf{Experimental settings}} \ \
We conduct the hyperparameter tuning and evaluation phases
using benchmark datasets, as shown in Table \ref{table:datasets_cil}.
From ImageNet-1k~\citep{(imagenet)deng2009imagenet}, we derive two subsets, ImageNet-100-1 and ImageNet-100-2, each containing 100 randomly selected non-overlapping classes. 
To account for varying dataset similarities, we further divide CIFAR-100~\citep{(cifar)krizhevsky2009learning} and ImageNet-100-1 into disjoint classes, generating CIFAR-50-1, CIFAR-50-2, ImageNet-50-1, and ImageNet-50-2. \iclr{We focus on two primary class-incremental learning (class-IL) scenarios~\citep{(cil_survey)masana2020class}: \textbf{10 Tasks}, where the model learns an equal number of classes from each task, and \textbf{6 Tasks}, where the model learns half of the total classes in the first task then evenly distributes the remaining classes evenly across subsequent tasks. 
Note that evaluating using both scenarios has been widely considered the proper assessment of each algorithm~\citep{(cil_survey)masana2020class,zhou2023pycil}} The table presents the configuration of the number classes (C) and tasks (T) for each scenario.
\begin{wraptable}{r}{0.45\textwidth} 
\vspace{-0.15in}
\caption{\iclr{Scenarios and datasets.}}
\vspace{-0.1in}
\centering
\begin{minipage}[b]{0.45\textwidth}
\resizebox{.98\linewidth}{!}{
\begin{tabular}{c||c|c}
\hline
Scenario                                                                           & $D^{HT}$                                                                             & $D^{E}$                                                                              \\ \hline \hline
\begin{tabular}[c]{@{}c@{}}10 Tasks\\ (C10$\times$T10)\end{tabular}                & \multirow{2}{*}{ImageNet-100-1}                                                      & \multirow{2}{*}{ImageNet-100-2}                                                      \\ \cline{1-1}
\begin{tabular}[c]{@{}c@{}}6 Tasks\\ (C50$\times$T1 +  C10$\times$T5)\end{tabular} &                                                                                      &                                                                                      \\ \hline
\begin{tabular}[c]{@{}c@{}}10 Tasks\\ (C5$\times$T10)\end{tabular}                 & \multirow{2}{*}{\begin{tabular}[c]{@{}c@{}}ImageNet-50-1,\\ CIFAR-50-1\end{tabular}} & \multirow{2}{*}{\begin{tabular}[c]{@{}c@{}}ImageNet-50-2,\\ CIFAR-50-2\end{tabular}} \\ \cline{1-1}
\begin{tabular}[c]{@{}c@{}}6 Tasks\\ (C25$\times$T1 +  C5$\times$T5)\end{tabular}  &                                                                                      &                                                                                      \\ \hline
\end{tabular}
}
\end{minipage}\label{table:datasets_cil}
\vspace{-0.15in}
\end{wraptable}
We conduct experiments using ResNet~\citep{(resnet)he2016deep}. 
We employ two key performance metrics commonly used for evaluating class-IL algorithms~\citep{(cil_survey)masana2020class}: {$\textbf{Acc}$} is final classification accuracy for the entire validation dataset after training the final task, and $\textbf{AvgAcc} = \frac{1}{T} \sum^{T}_{t=1}{Acc_t}$, where $Acc_t$ denotes accuracy on the validation data up to task $t$.
The hyperparameters that yield the highest \textbf{harmonic mean} of Acc and AvgAcc are selected during the hyperparameter tuning phase\tmlrcsm{, as most studies evaluate algorithms by simultaneously considering both metrics.}

\noindent{\textbf{Baselines}} \ \ 
We evaluate nine major class-IL algorithms, including replay-based methods (Replay, iCaRL~\citep{(icarl)rebuffi2017icarl}, \csm{and} WA~\citep{(wa)zhao2020maintaining}) and regularization-based methods (BiC~\citep{(bic)wu2019large} \csm{and} PODNet~\citep{(podnet)douillard2020podnet}) and expansion-based methods (DER~\citep{(der)yan2021dynamically}, FOSTER~\citep{(foster)wang2022foster}, \csm{and} BEEF~\citep{wang2022beef}). \iclr{Note that we use the partially implemented DER, as neither PyCIL nor the official DER code includes the implementation details for masking and pruning}. Replay serves as a naive baseline, where a model is fine-tuned using both the exemplar memory and the current task's dataset.
Note that these algorithms have demonstrated progressively improved performance in the order of their publication. Among them, FOSTER, BEEF, and MEMO are recognized as the current \textit{state-of-the-art}, reporting superior performance that surpasses DER by a small margin.
We conduct experiments using the implementation code proposed in PyCIL~\citep{zhou2023pycil}.
The size of the exemplar memory is set to 2000 for ImageNet-100, and 1000 for ImageNet-50 and CIFAR-50 variants.
More details on settings, predefined hyperparameter sets and selected hyperparameters are presented in Section \ref{appendix:environment} of the Appendix.

\begin{figure*}[t]
\centering 
\subfigure[Results on $D^{HT}= \ \text{ImageNet-100-1}$]
{\includegraphics[width=0.60\linewidth]{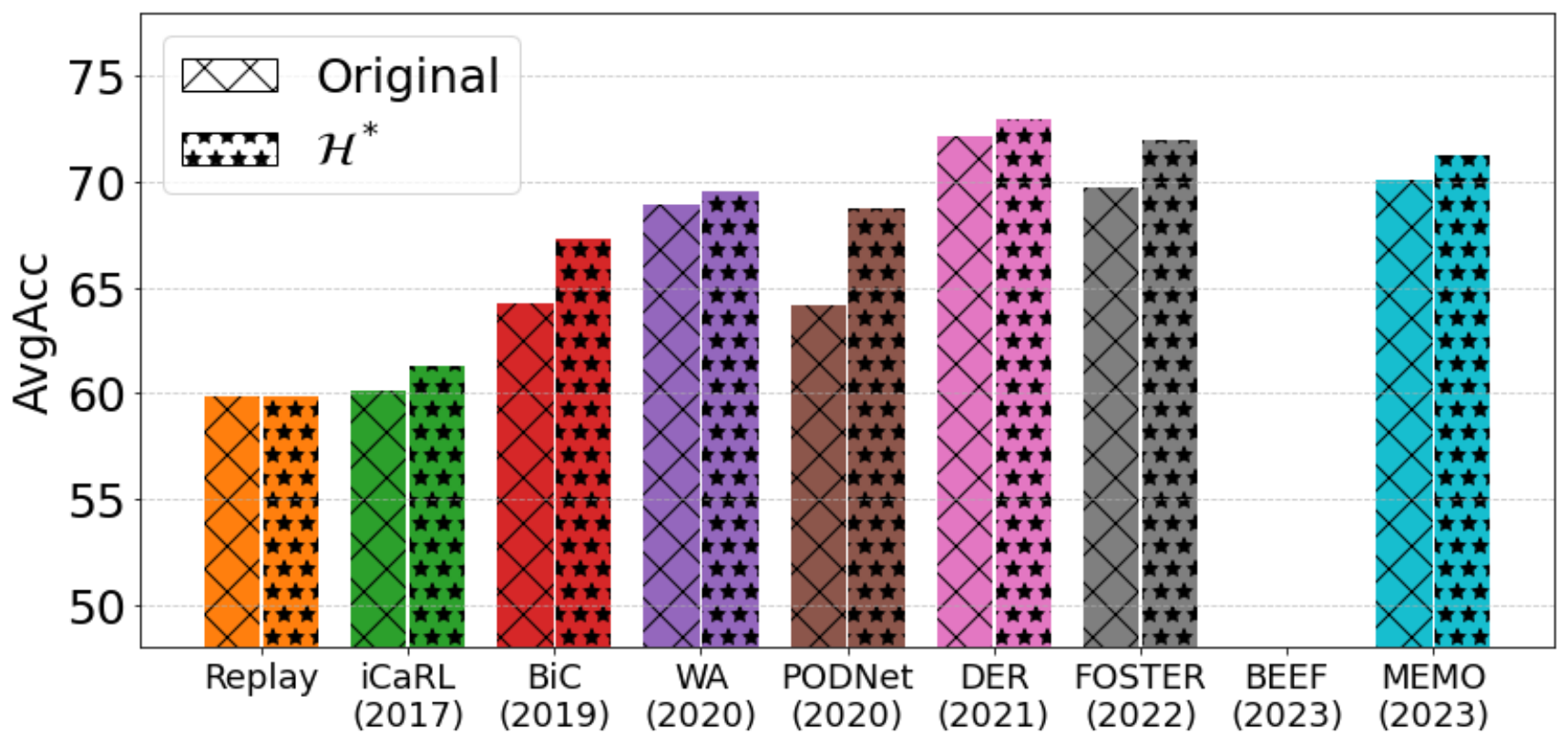}\label{figure:original_result_cil}}
\subfigure[Comparing results on $D^{HT}$ and $D^{E}$]
{\includegraphics[width=0.39
\linewidth]{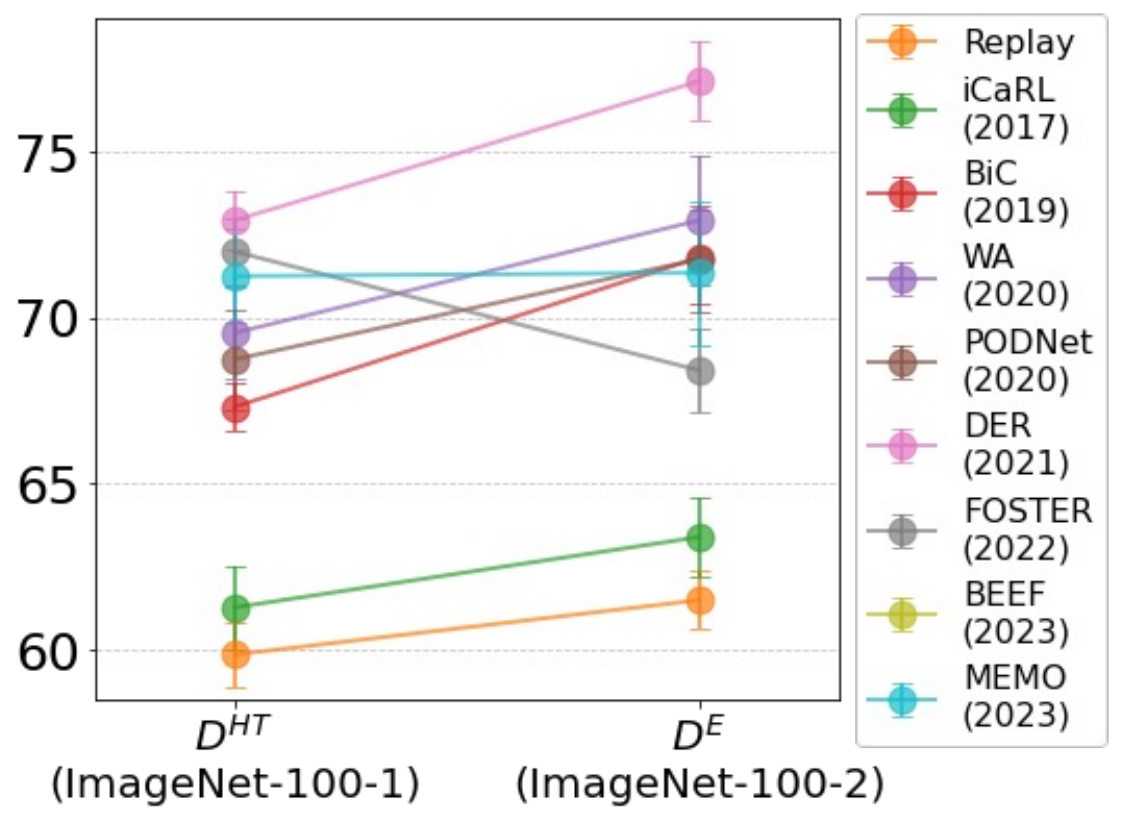}\label{figure:compare_cil}}
\vspace{-.2in}
\caption{Experimental results (AvgAcc) on the 10 Tasks scenario using ImageNet-100-1 for $D^{HT}$ and ImageNet-100-2 for $D^{E}$ (high similarity). The term Original and $\mathcal{H}^*$ refer to the use of reported hyperparameters and hyperparameters selected from our protocol, respectively. BEEF constantly returns NaN in training loss at specific seeds so we could not report its performance.}
\label{figure:each_task_graph}
 \end{figure*}

\noindent{\textbf{Experiments using original and selected hyperparameters}} \ \
To demonstrate whether the hyperparameters identified during the hyperparameter tuning phase achieve better performance than those previously reported, we conduct experiments with both sets of hyperparameters.
Figure \ref{figure:original_result_cil} presents results on $D^{HT} = \text{ImageNet-100-1}$, showing that using the best hyperparameters ($\mathcal{H}^*$) generally outperforms the \csm{original} ones across all algorithms except BEEF.
Note that the performance differences among DER, FOSTER, and MEMO are within their respective standard deviations.
Considering the hyperparameter tuning phase aligns with the conventional evaluation protocol, this graph indicates that each algorithm reflects the performance trends observed in their respective papers, gradually improving over time in accordance with the order of publication.
On the other hand, we confirm that BEEF is significantly sensitive to hyperparameters, as it occasionally results in NaN (Not a Number)  in training loss for specific seeds, even when using the \csm{original} hyperparameters.

In the evaluation phase, we apply the best hyperparameters to assess the CL capacity in unseen scenarios generated by $D^{E}$.
\iclr{Note that, due to differences in the datasets between these phases, the final performance may vary across phases, even when using identical hyperparameters for each algorithm.}
Figure \ref{figure:compare_cil} presents experimental results. The graph shows that the CL capacity of the state-of-the-art algorithms (\textit{i.e.}, FOSTER, BEEF, and MEMO) is significantly inferior to that of older algorithms, such as WA, BiC and PODNet. Additionally, BEEF again produces NaN values for certain seeds.
In contrast, DER demonstrates superior generalizability of its CL capacity, consistently maintaining strong performance in both phases.

\begin{figure*}[t]
\vspace{-.05in}
\centering 
{\includegraphics[width=0.95\linewidth]{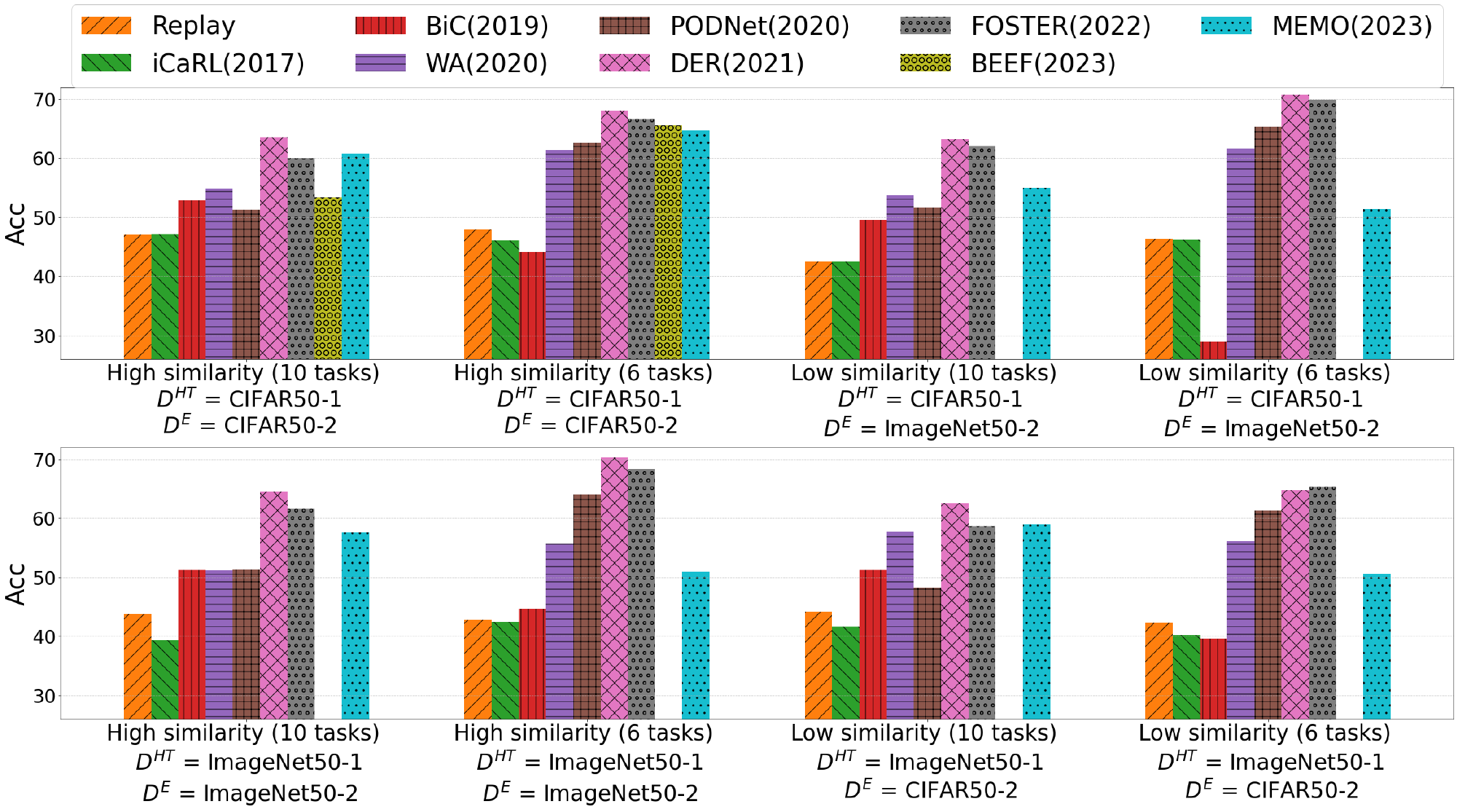}}
\vspace{-.15in}
\caption{Bar graphs depict the experimental results from the evaluation phase. The Y-axis represents final classification accuracy (Acc). The parentheses next to each algorithm indicate the publication year. The bar graphs in the first row show the experimental results using the best hyerparameters selected in the hyperparameter tuning phase with $D^{HT}= \text{CIFAR-50-1}$ , while the graphs in the second row show the results using $D^{HT}= \text{ImageNet-50-1}$ . In cases of using ImageNet-50-1 or ImageNet-50-2, we encountered challenges in obtaining results for BEEF due to NaN issues.}
\label{figure:overall_performance}
\vspace{-.25in}
 \end{figure*}
 
\noindent{\textbf{Experiments across diverse similarity cases}} \
To broadly assess the generalizability of each algorithm’s CL capacity, we conduct experiments across various similarity cases. The bar graphs in the first row of Figure \ref{figure:overall_performance} display results for both high and low similarity cases, using the best hyperparameters selected during the hyperparameter tuning phase using $D^{HT}=\text{CIFAR-50-1}$.
In most cases, iCaRL performs worse than Replay, and BiC also struggles in some cases (\textit{e.g.}, 6 tasks in both high and low similarity settings). Additionally, both WA and PODNet outperform other regularization-based methods, with PODNet particularly excelling in the 6 Tasks. Lastly, the current state-of-the-art methods—FOSTER, BEEF, and MEMO—exhibit lower performance compared to DER, with BEEF again showing significant sensitivity, especially on ImageNet-50-2.

The second row of Figure \ref{figure:overall_performance} presents results using the best hyperparameters selected based on $D^{HT} = \text{ImageNet-50-1}$. The trends are consistent with previous experiments: DER maintains superior performance in most cases, although FOSTER surpasses DER in the low similarity case (6 tasks). Additionally, BEEF suffers from NaN issues  in training loss for certain seeds. \tmlrcsm{Additionally, from the results of the high- and low-similarity cases, we observe that BiC and MEMO are particularly sensitive to hyperparameters.}

\begin{figure*}[h!]
\vspace{-.15in}
\centering 
\subfigure[Performance graph]
{\includegraphics[width=0.34\linewidth]{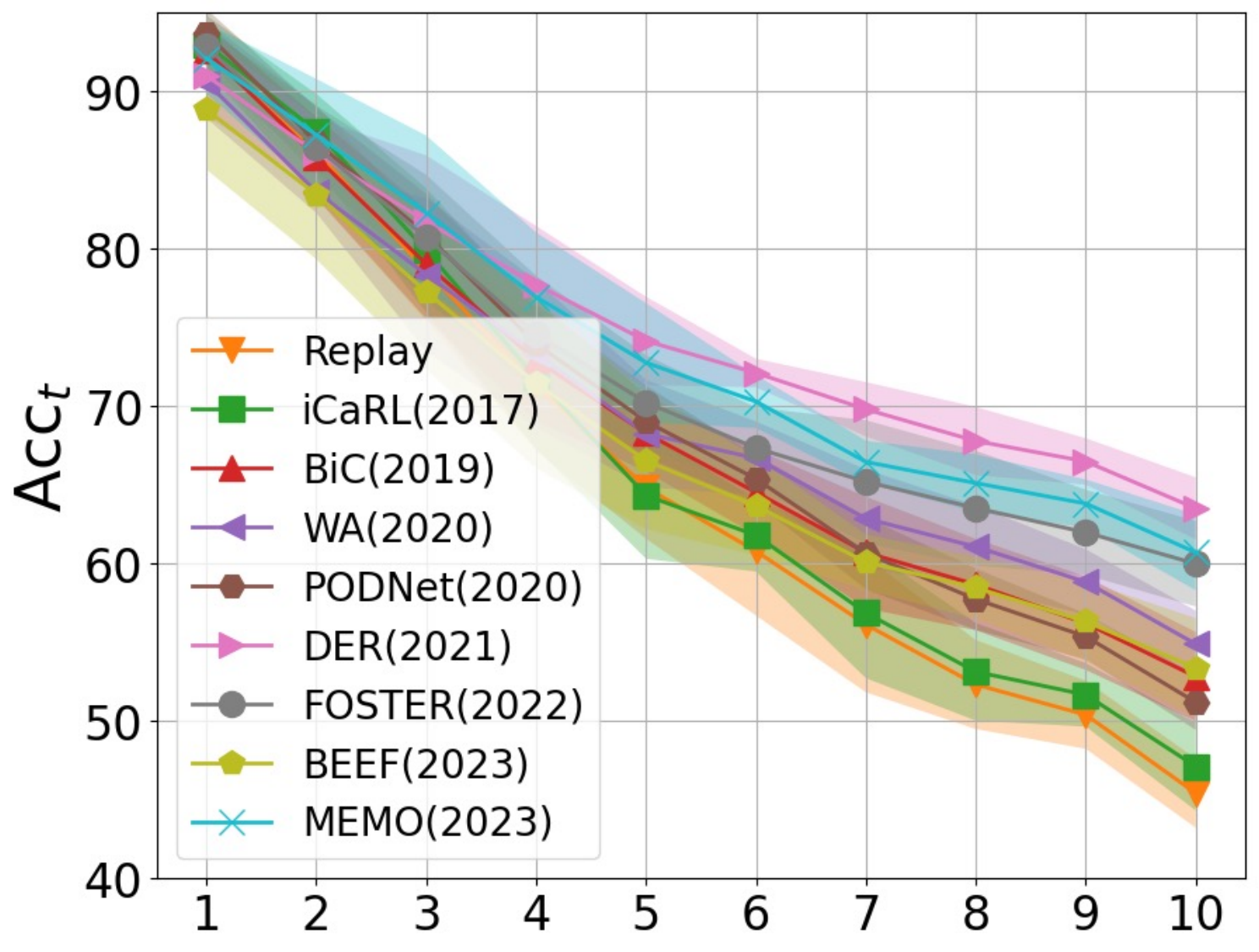}\label{figure:cifar50_1}}
\subfigure[Number of parameters]
{\includegraphics[width=0.31\linewidth]{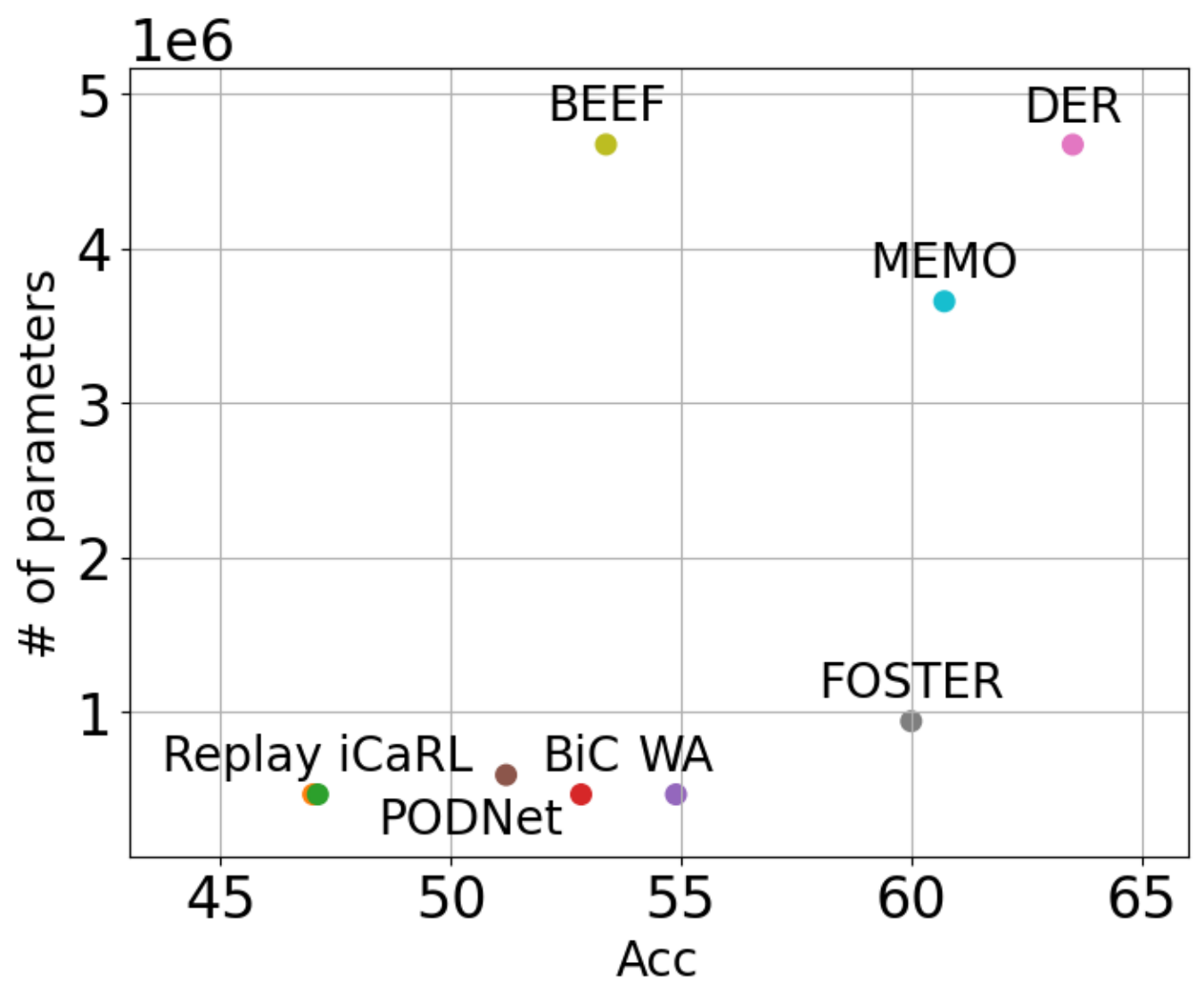}\label{figure:parameter}}
\subfigure[Total training time]
{\includegraphics[width=0.31\linewidth]{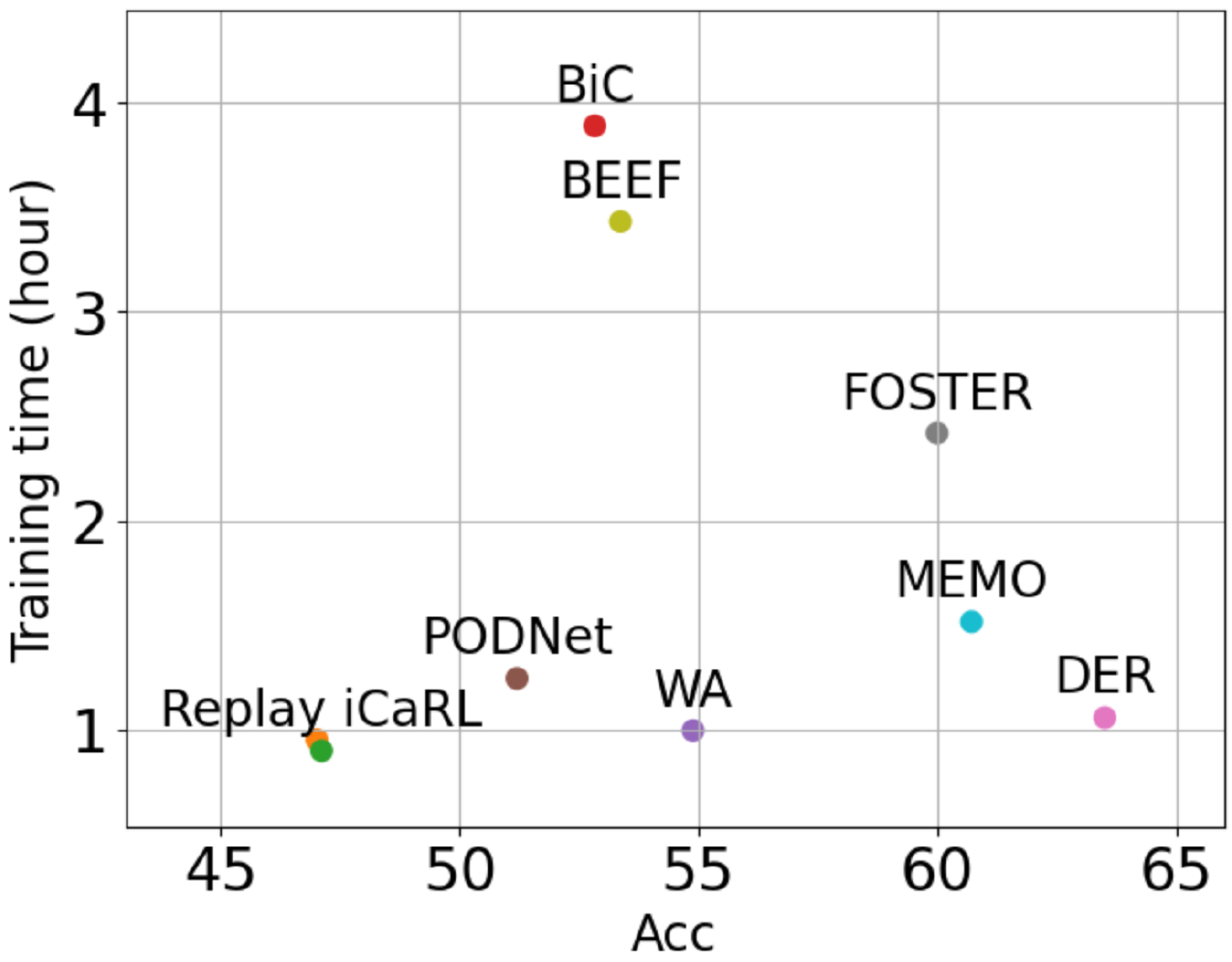}\label{figure:training_time}}
\vspace{-.15in}
\caption{Experimental analysis in the evaluation phase. All experimental results are obtained by first identifying the best hyperparameters using CIFAR-50-1 (10 Tasks) in the hyperparameter tuning phase, then evaluating each algorithm using CIFAR-50-2 (10 Tasks) in the evaluation phase. (b) and (c) show results after training up to the final task.}
\label{figure:analysis_cil}
\vspace{-.15in}
 \end{figure*}

\noindent{\textbf{Additional analysis}} \ \ 
Figure \ref{figure:cifar50_1} shows the evaluation results for each task $t$ in the evaluation phase, with shaded areas representing the standard deviation across 5 trials. From these graphs, it is evident that DER consistently outperforms current state-of-the-art algorithms (\textit{i.e.}, FOSTER, BEEF and MEMO). Considering the standard deviation, the performances of FOSTER and MEMO are nearly indistinguishable. Among the remaining algorithms, WA demonstrates relatively better performance while BEEF performs similarly to the order algorithms. 

\iclr{Recent studies have increasingly focused on evaluating CL algorithms based on their training costs, particularly in terms of GPU usage and energy consumption~\citep{prabhu2023computationally, chavan2023towards}. However, these evaluations were often conducted by either limiting the number of training iterations or comparing costs under a fixed number of training epochs. Building on this, we extend the analysis by examining the final model size and total training time for each algorithm, using their best hyperparameters to ensure a fair and comprehensive comparison of efficiency.} Figures  \ref{figure:parameter} and \ref{figure:training_time} present scatter plots showing achieved accuracy, total parameter counts, and training times. 
DER performs the best and requires relatively less training time. 
Nevertheless, it exhibits considerable inefficiency in the total number of parameters, which increases linearly with the number of tasks, raising concerns about its actual cost-efficiency as a CL algorithm. 
On the other hand, BiC, BEEF, and MEMO fail to demonstrate superior performance while requiring similar or longer training times compared to DER, highlighting their serious inefficiency.


\subsection{Class-incremental learning with pretrained models}

\noindent{\textbf{Experimental details}} \ \
We conduct both the hyperparameter tuning and evaluation phases
\begin{wraptable}{r}{0.4\textwidth} 
\vspace{-0.15in}
\caption{\iclr{Scenarios and datasets.}}
\vspace{-0.1in}
\centering
\begin{minipage}[b]{0.4\textwidth}
\resizebox{.98\linewidth}{!}{
\begin{tabular}{c||c|c}
\hline
Scenario                                                         & $D^{HT}$                                                                           & $D^{E}$                                                                                            \\ \hline \hline
\begin{tabular}[c]{@{}c@{}}20 Tasks\\ (C10$\times$T20)\end{tabular} & \multirow{2}{*}{\begin{tabular}[c]{@{}c@{}}CUB-200,\\ ImageNet-R\end{tabular}}     & \multirow{2}{*}{\begin{tabular}[c]{@{}c@{}}ImageNet-R,\\ ImageNet-A\end{tabular}}                  \\ \cline{1-1}
\begin{tabular}[c]{@{}c@{}}10 Tasks\\ (C20$\times$T10)\end{tabular} &                                                                                    &                                                                                                    \\ \hline
\begin{tabular}[c]{@{}c@{}}20 Tasks\\ (C5$\times$T20)\end{tabular}  & \multirow{2}{*}{\begin{tabular}[c]{@{}c@{}}CUB-100-1,\\ ImageNet-R-1\end{tabular}} & \multirow{2}{*}{\begin{tabular}[c]{@{}c@{}}CUB-100-2,\\ ImageNet-R-2,\\ ImageNet-A-2\end{tabular}} \\ \cline{1-1}
\begin{tabular}[c]{@{}c@{}}10 Tasks\\ (C10$\times$T10)\end{tabular} &                                                                                    &                                                                                                    \\ \hline
\end{tabular}
}
\end{minipage}\label{table:datasets_cil_pretrained}
\vspace{-0.15in}
\end{wraptable}
 using widely used datasets in class-incremental learning (class-IL) with pretrained models, including CUB-200~\citep{(cub200)wah2011caltech}, ImageNet-R~\citep{(imagenetr)hendrycks2021many}, and ImageNet-A~\citep{(imageneta)hendrycks2021natural}, all of which contain 200 classes. 
To explore diverse similarity cases, we divide these datasets into two disjoint subsets, as outlined in Table \ref{table:datasets_cil_pretrained}.
\iclr{Following \cite{sun2023pilot}, we consider two major class-IL scenarios: \textbf{20 Tasks} and \textbf{10 Tasks}, where the model learns an equal number of classes in each task. Note that the 20 Tasks scenario has been commonly regarded as the standard for better evaluating algorithm performance due to the need to handle more tasks.}
For all experiments, we employ the ViT B16 model, which is pretrained on ImageNet~\citep{(vit)dosovitskiy2020image}. 
The best hyperparameters are selected based on the same metrics: the \textbf{harmonic mean} of \textbf{Acc} and \textbf{AvgAcc}. 


\noindent{\textbf{Baselines}} \ \ 
We select six major algorithms: prompt-based methods (L2P~\citep{(l2p)wang2022learning}, DualPrompt~\citep{(dual)wang2022dualprompt} and CODA-Prompt~\citep{(coda)smith2023coda}) and representation-based methods (Adam-Adapter~\citep{(adam)zhou2023revisiting}, Ranpac~\citep{(ranpac)mcdonnell2024ranpac} and EASE~\citep{(ease)zhou2024expandable}). Within each category, CODA-Prompt and EASE represent current \textit{state-of-the-art }algorithms. Although DAP~\citep{(gen)jung2023generating} reports better performance within the prompt-based method category, we exclude it due to fairness issues in comparison, as mentioned in \cite{(cil_pre)zhou2024continual}. All experiments are conducted using code implemented in PILOT~\citep{sun2023pilot}. 
Details on experimental settings, predefined hyperparameter sets \tmlrcsm{and} the best hyperparameters are proposed in Section \ref{appendix:environment_pretrained} of the Appendix.

\begin{figure*}[h]
\vspace{-.1in}
\centering 
\subfigure[Results on $D^{HT}= \ \text{CUB-200}$]
{\includegraphics[width=0.60\linewidth]{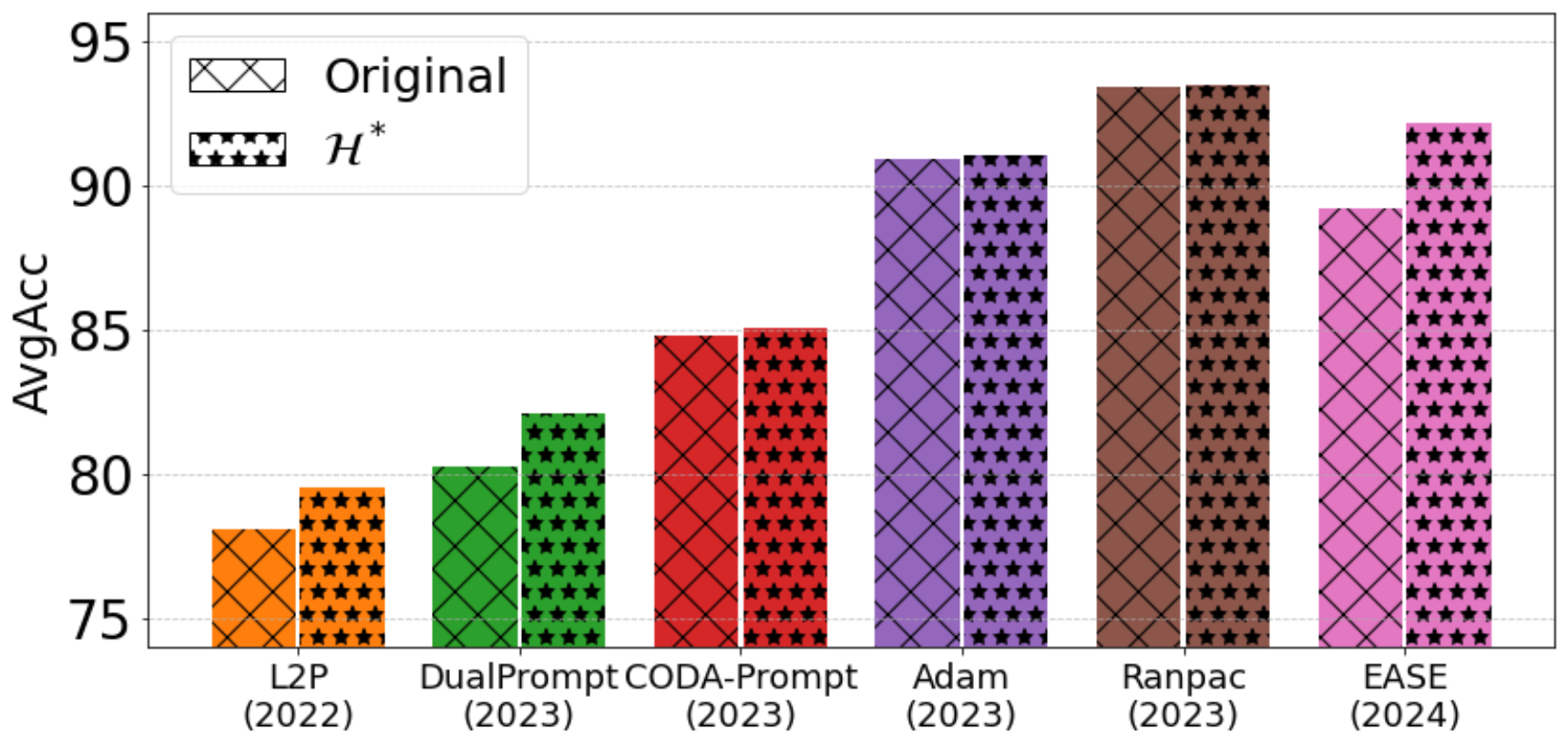}\label{figure:original_result_cil_pretrained}}
\subfigure[Comparing results on $D^{HT}$ and $D^{E}$]
{\includegraphics[width=0.39
\linewidth]{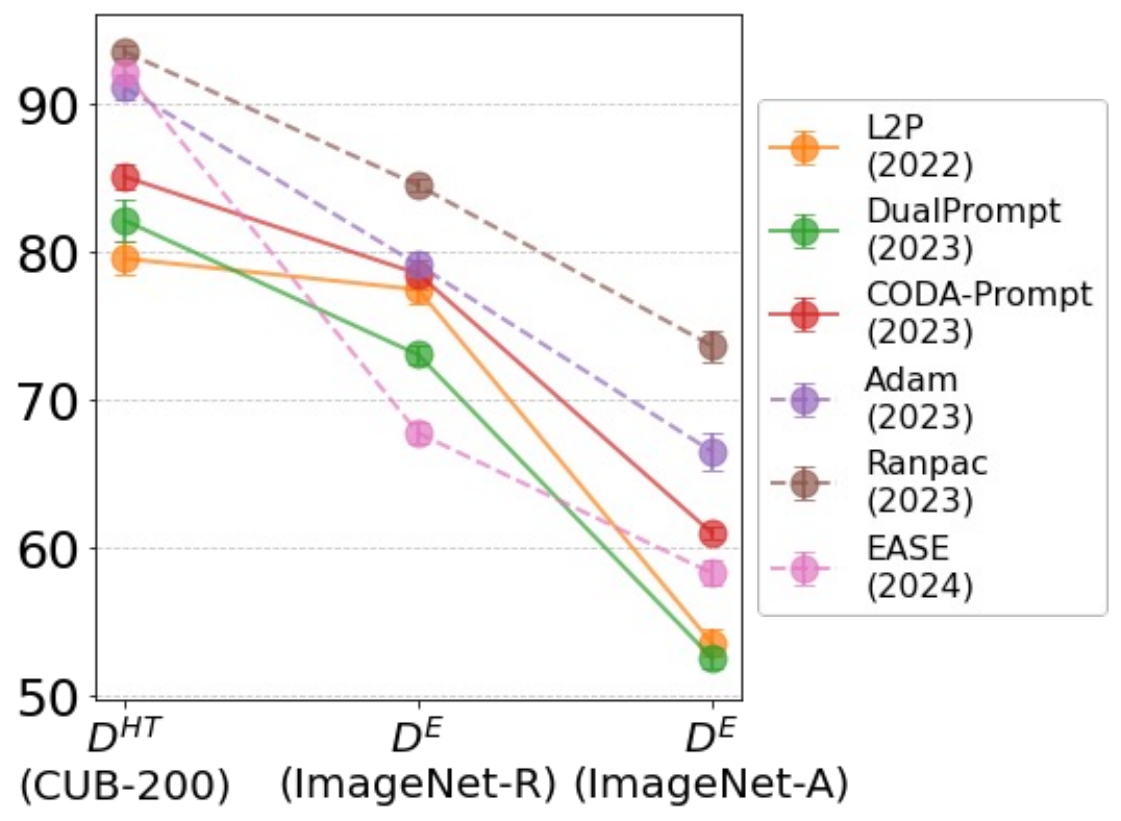}\label{figure:compare_cil_pretrained}}
\vspace{-.2in}
\caption{Experimental results (AvgAcc) for 10 Tasks scenario using CUB-200 for $D^{HT}$, ImageNet-R, and ImageNet-A for $D^{E}$ (low similarity). The term Original and $\mathcal{H}^*$ refer to the use of original hyperparameters and the hyperparameters selected from our protocol, respectively.}

 \end{figure*}
\begin{figure*}[t]
\vspace{-.15in}
\centering 
{\includegraphics[width=0.95\linewidth]{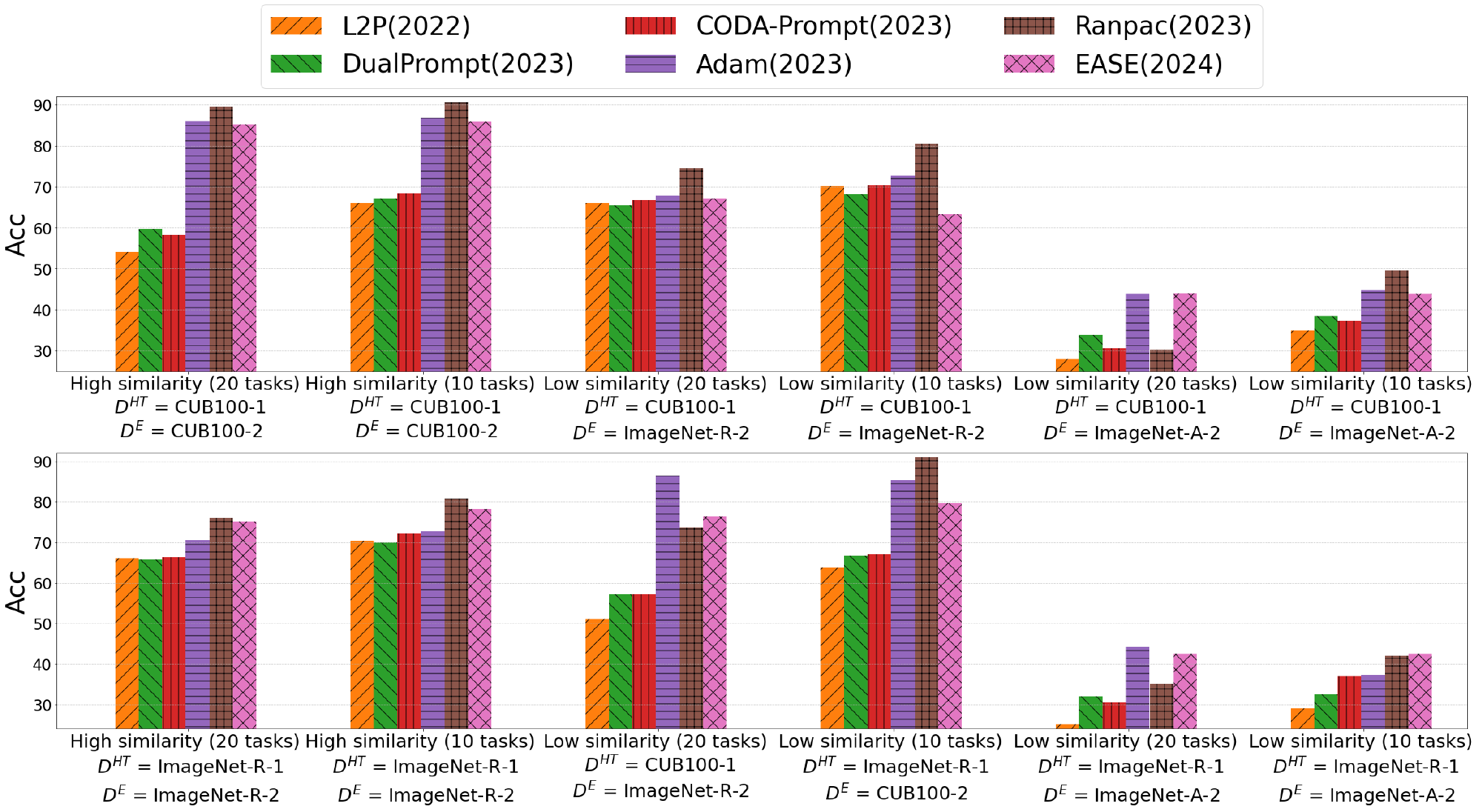}}
\vspace{-.15in}
\caption{Bar graphs depict the experimental results from the evaluation phase. The Y-axis represents final accuracy (Acc). In the legend, the parentheses next to each algorithm indicate the publication year. The bar graphs in the first row show the experimental results using the best hyerparameters selected in the hyperparameter tuning phase with $D^{HT}= \text{CUB100-1}$ , while the graphs in the second row display the results using $D^{HT}= \text{ImageNet-R-1}$ .}\label{figure:overall_performance2}
 \vspace{-.3in}
 \end{figure*}

\begin{figure*}[h!]
\vspace{-.05in}
\centering 
\subfigure[Performance graph]
{\includegraphics[width=0.32\linewidth]{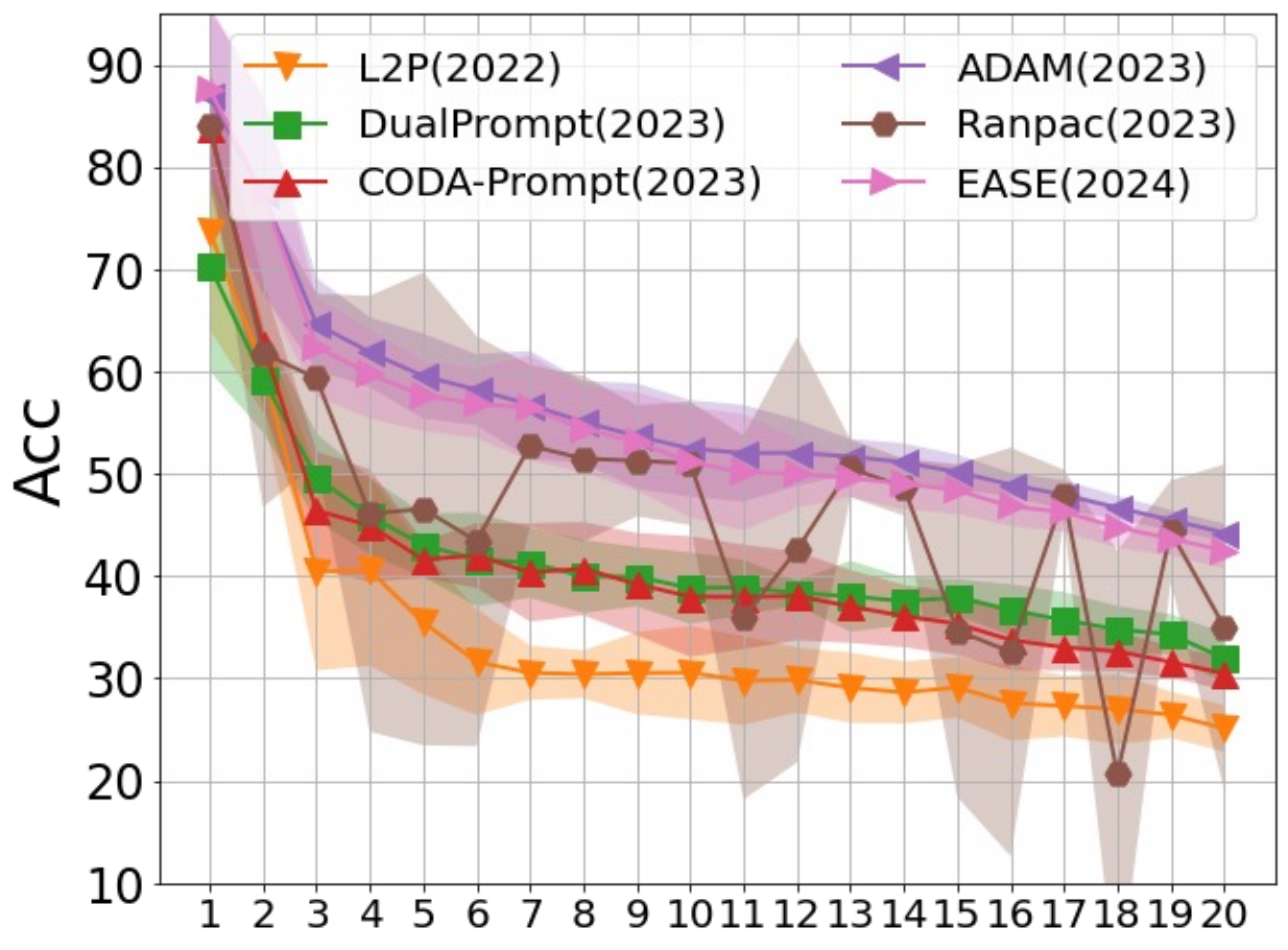}\label{figure:pretrained_2}}
\subfigure[\# of trainable params]
{\includegraphics[width=0.32\linewidth]{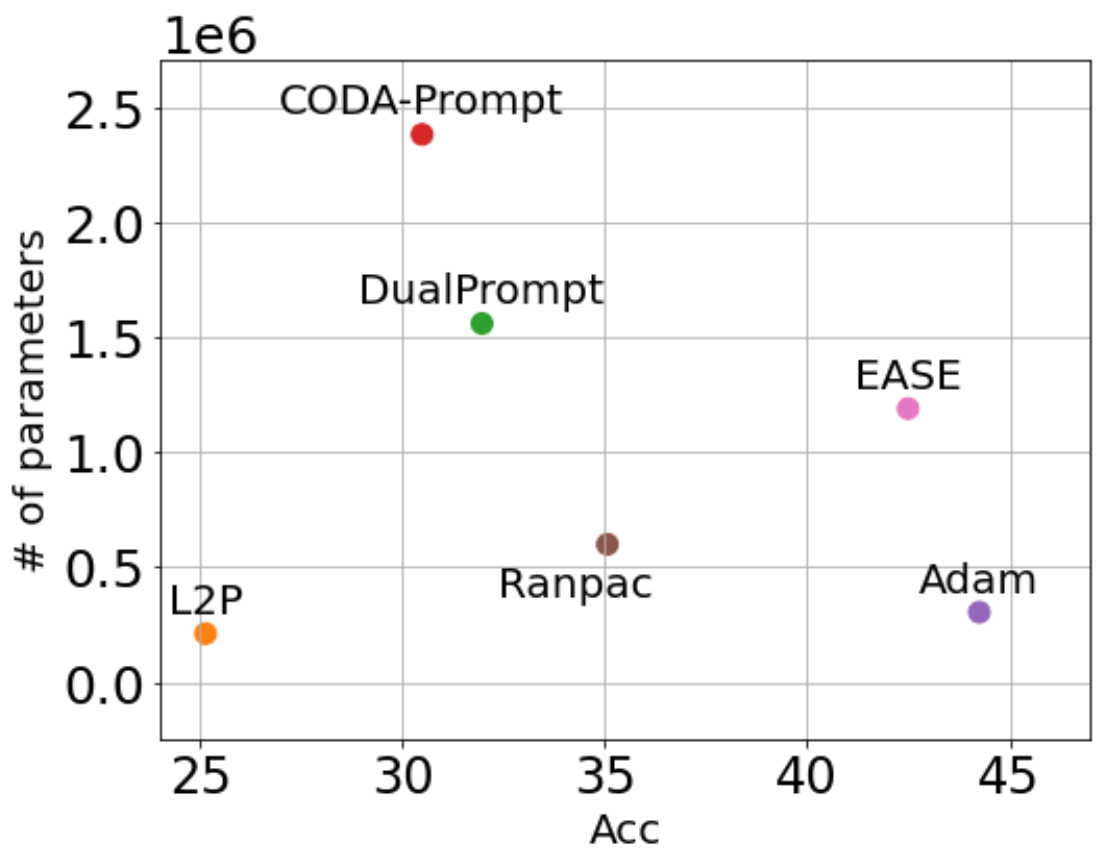}\label{parameters_pretrained}}
\subfigure[Training time]
{\includegraphics[width=0.32\linewidth]{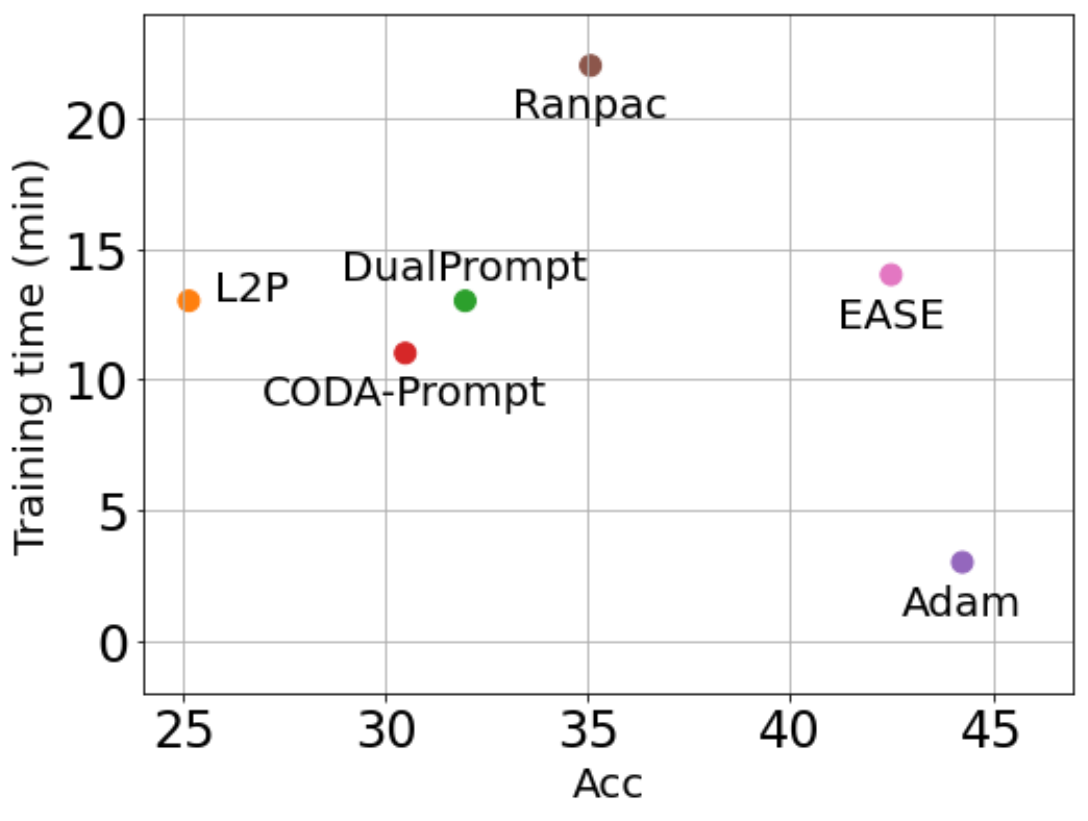}\label{figure:training_time_pretrained}}
\vspace{-.15in}
\caption{Experimental analyses in the evaluation phase. All experimental results are obtained by first identifying the best hyperparameters using ImageNet-R-1 (20 Tasks) in the hyperparameter tuning phase, then evaluating each algorithm using ImageNet-A-2 (20 Tasks) in the evaluation phase. (b) and (c) show the results after training up to the final task.}
\label{figure:analysis_cil_pretrained}
 \vspace{-.25in}
 \end{figure*}

 \noindent{\textbf{Experiments using original and selected hyperparameters}} \ \
 To verify best hyperparameters selected in the hyperparameter tuning phase, we conduct experiments on $D^{HT}=\text{CUB-200}$ using both the original and selected hyperparameters of each algorithm. Figure \ref{figure:original_result_cil_pretrained} demonstrates that using the selected hyperparameters leads to better performance across all algorithms. Additionally, the performance of each algorithm gradually improves in accordance with their publication order, as reported in the respective papers. However, Ranpac and EASE achieve similar performance, with differences falling within their standard deviations.

 Following our evaluation protocol, we apply the best hyperparameters for each algorithm in the evaluation phase. We conduct experiments for two evaluation phases using ImageNet-R and ImageNet-A as $D^{E}$ and Figure \ref{figure:compare_cil_pretrained} shows experimental results. From these results, we confirm the following observations: First, among the prompt-based algorithms (solid lines), DualPrompt exhibits degraded performance compared to L2P in both evaluation phases. Additionally, CODA-Prompt demonstrates superior performance in all cases, although it shows nearly identical performance to L2P in the ImageNet-R. In the case of the representation-based algorithms (dashed lines), Ranpac consistently demonstrates superior performance across all datasets; however, we observe some instability in specific scenarios, as will be shown in the following analysis. Furthermore, EASE, recognized as the current state-of-the-art, shows significantly poorer performance in both evaluation phases. 


\noindent{\textbf{Experiments across diverse similarity cases}} \ \ Figure \ref{figure:overall_performance2} presents the experimental results evaluated in the evaluation phase. 
Similar to trends reported in \cite{(cil_pre)zhou2024continual}, representation-based methods generally outperform prompt-based methods. However, significant differences are observed under the proposed evaluation protocol:
First, the prompt-based methods have reported substantial performance improvements over previous algorithms (\textit{e.g.},  7-10\% increases on the CUB200 dataset for each algorithm~\citep{(cil_pre)zhou2024continual}). However, the proposed evaluation protocol reveals either no significant performance difference between them (\textit{e.g}., low similarity (20 tasks) using ImageNet-R-2 in the first row of the graph) or cases where an \tmlrcsm{older} algorithm outperforms a newer one  (\textit{e.g.},  high similarity (20 tasks) using CUB100-2 in the first row of the graph).
Second, the current state-of-the-art representation-based method, EASE, often underperforms compared to Ranpac, especially in low similarity cases (\textit{e.g.}, low similarity (10 tasks) using ImageNet-R-2 in the first row of the graph).
Lastly, while Ranpac achieves the best performance in most cases, it exhibits significantly degraded performance in several low similarity cases (\textit{e.g.}, low similarity (20 tasks) using ImageNet-A-2 in the first row of the graph). This degradation is attributed to considerable performance instability in certain tasks.
\tmlrcsm{Furthermore, by comparing the results of the high- and low-similarity cases, we observe that Adam and Ranpac are highly sensitive to hyperparameters. Specifically, while Ranpac generally achieves strong performance in the high-similarity case, its performance is sometimes reversed in the low-similarity case.}

\noindent{\textbf{Additional analysis}} \ \ 
As we already confirmed in the previous experiments, Figure \ref{figure:pretrained_2} illustrates that Ranpac suffers from significant instability in certain tasks, resulting in a substantial increase in standard deviation (shaded area).
Furthermore, we observe that the state-of-the-art algorithms, EASE and CODA-Prompt in their respective categories, do not consistently outperform baseline algorithms like Adam and DualPrompt in many cases, highlighting a lack of generalizability in their CL capacity.

Figures \ref{parameters_pretrained} and \ref{figure:training_time_pretrained} display the number of trainable parameters and training times with the best hyperparameters. For prompt-based algorithms, training times are comparable; however, CODA-Prompt requires more parameters while delivering lower performance compared to DualPrompt.
Among representation-based methods, the oldest algorithm \csm{(\textit{i.e.}, Adam)} achieves the best performance with minimal costs in terms of trainable parameters and training time. 

In Section \ref{appendix:additional_results} of the Appendix, we present additional experimental results, including training graphs and numerical data related to the results discussed in the manuscript.

\section{Concluding Remarks}

\noindent\textbf{Limitations of the conventional evaluation protocol} \ \
The conventional evaluation protocol, which remains the standard for assessing continual learning (CL) algorithms, exhibits critical shortcomings. Most notably, it assumes access to repeated training within the same scenario for hyperparameter tuning—an unrealistic assumption in real-world applications. This approach not only fails to reflect practical CL settings but also systematically overestimates the actual CL capacity of algorithms.
According to fundamental principles of machine learning evaluation, the generalizability of a model—particularly to unseen scenarios—should be the primary metric of interest. In this context, we introduced the Generalizable Two-phase Evaluation Protocol (GTEP), which separates hyperparameter tuning and evaluation across different scenarios, thereby enabling a more faithful assessment of generalizability across varying degrees of similarity.

\noindent\textbf{Summary of experimental findings} \ \
Our experiments yield three key insights. First, many algorithms show strong performance in seen scenarios but struggle to generalize to unseen ones, indicating that their success under the conventional protocol often stems from overfitting. Second, several algorithms are highly sensitive to hyperparameters, leading to failures on certain task orders or unstable performance across tasks. Third, even algorithms that generalize well under our protocol (GTEP) often require substantial training time or parameter usage, undermining the cost-efficiency goal of CL. Although our experiments focus on class-incremental learning, these limitations likely extend to other CL domains that rely on the same conventional evaluation practices.


\noindent\textbf{Toward more realistic evaluation: key takeaways} \ \ 
We argue that the proposed GTEP offers a principled and realistic framework for evaluating CL algorithms. 
To advance CL research meaningfully, we recommend that future evaluations across all CL domains incorporate the following minimum criteria:

\begin{itemize}
\item{\textbf{Generalization Check using GTEP}: Does the algorithm outperform baselines in the evaluation phase when using hyperparameters selected from a separate hyperparameter tuning phase?}
\item{\textbf{Efficiency and Stability Check}: Is the algorithm more cost-efficient than the baselines (\textit{e.g.}, in terms of training/inference time, parameter count, or GPU usage)? Does it consistently maintain stable performance across different tasks and scenarios (\textit{e.g.}, varying task orders)?}
\end{itemize}

\section{Limitations and Future Work}
Our study is not without limitations. First, implementing GTEP requires repeated training trials, which can be computationally demanding. In our experiments, we conducted 30 random trials per algorithm with 5 seeds per trial. Encouragingly, we observed that most algorithms converged to their optimal hyperparameters within 20 trials, with negligible gains beyond that point. Nevertheless, the development of more sample-efficient tuning strategies—such as targeted search based on hyperparameter importance—remains an important direction for future research.
Second, our experiments assume predictable CL scenarios, where the number of tasks and class distributions are known in advance, reflecting structured and practical deployments. However, evaluating the robustness of algorithms in unpredictable or adaptive scenarios (e.g., unknown task boundaries or shifting data distributions) is a critical next step, particularly for adaptive methods that dynamically adjust hyperparameters.
Lastly, our experiments focus on offline class-incremental learning. Extending GTEP to online CL settings and broader CL domains—such as class incremental semantic segmentation, continual self-supervised learning continual reinforcement learning, and continual learning for LLMs—will be essential for validating the scalability and practical utility of CL algorithms in each domain.

\newpage

\section{Acknowledgement}

This work was supported by the Institute of Information \& Communications Technology Planning \& Evaluation (IITP) with a grant funded by the Ministry of Science and ICT (MSIT) of the Republic of Korea in connection with the Global AI Frontier Lab International Collaborative Research. This work was also supported by the Samsung Advanced Institute of Technology (under the project Next Generation Deep Learning: From Pattern Recognition to AI) and the National Science Foundation (under NSF Award 1922658) This work was supported in part through the NYU IT High Performance Computing resources, services, and staff expertise. We would especially like to thank Shenglong Wang for his generous support in setting up the experimental environment, which was instrumental in enabling our extensive experiments.




\bibliography{references}
\bibliographystyle{tmlr}


\newpage

\appendix

\newpage

\section{Additional Details on Experimental Settings}






\subsection{Class-incremental learning without a pretrained model}\label{appendix:environment}
\noindent\textbf{Experimental details} \ \ We conduct all experiments using PyCIL~\citep{zhou2023pycil} in the following environment: Python 3.8, PyTorch 1.13.1, and CUDA 11.7. We use ResNet-18 and ResNet-32 architectures for our experiments. For class-incremental learning without a pretrained model, we employ the SGD optimizer with a momentum of 0.9 across all methods, consistent with their respective implementations. Other hyperparameters, however, are sampled during the hyperparameter tuning phase.

\begin{table}[!h]
\caption{Hyperparameters for training the first task.}
\centering
\smallskip\noindent
\resizebox{0.6\linewidth}{!}{
\begin{tabular}{|c||c|}
\hline
Hyperparameters          & Values                                                                                                    \\ \hline \hline
Init epochs              & 200                                                                                                       \\ \hline
Init learning rate       & 0.1                                                                                                       \\ \hline
Init milestones          & \begin{tabular}[c]{@{}c@{}}{[}60, 120, 170{]}\\ (Only applied when `StepLR' is selected)\end{tabular} \\ \hline
Init learning rate decay & 0.1                                                                                                       \\ \hline
Init weight decay        & 0.0005                                                                                                    \\ \hline
\end{tabular}\label{table:first_hyperparameters}
}
\end{table}

\noindent\textbf{Predefiend hyperparameters} \ \ Recent studies have demonstrated that newer algorithms perform better when trained for more epochs on the first task and fewer epochs on subsequent tasks~\citep{(cil_survey)masana2020class}. Additionally, it is known that performance on the first task significantly impacts overall performance~\citep{cha2023objective}. To apply this approach consistently across all algorithms, we train a model on the first task using the hyperparameters listed in Table \ref{table:first_hyperparameters}. Subsequently, we train that model with randomly sampled hyperparameters starting from the second task.

\begin{figure}[h]
\vspace*{-0.1in}
\centering
\includegraphics[width=0.6\textwidth]{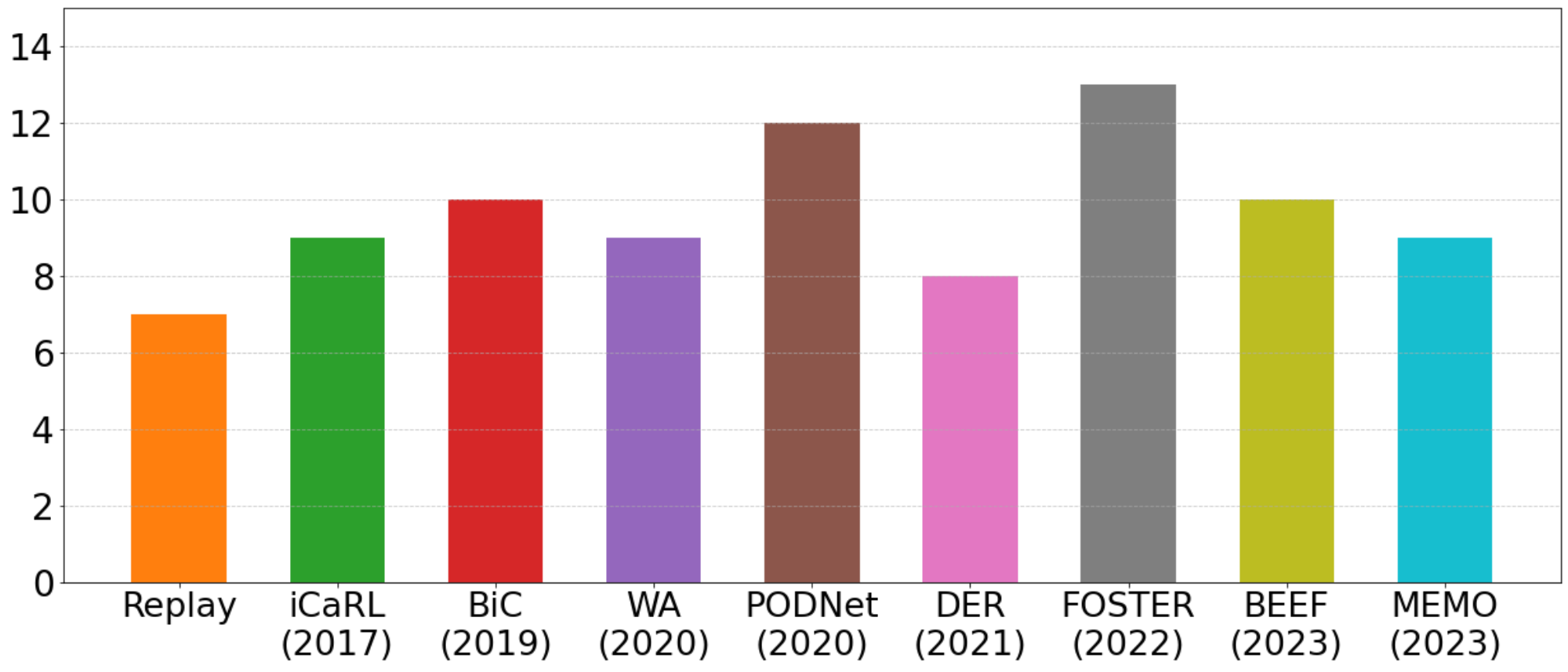}
\vspace*{-0.1in}
\caption{\# of hyperparameters.}\label{fig:hyperparameters_cil}
\vspace{-.1in}
\end{figure}

Figure \ref{fig:hyperparameters_cil} shows the number of hyperparameters for each algorithm. 
We consider both algorithm-specific and general hyperparameters in the hyperparameter tuning phase. 
Table \ref{table:predefined_hyperparameters} presents the sets of predefined hyperparameters considered for each algorithm. 
Note that `Epoch', `Num milestones', `LR decay', `Batch size', `Weight decay', and `LR scheduler' are commonly considered hyperparameters for all algorithms. Additionally, both `Num milestones' and `Lr decay' are applicable only when `StepLR' is selected as a scheduler. The others are specific hyperparameters of each algorithm.
We consider all the hyperparameters necessary for implementing each algorithm. For instance, even if a specific algorithm uses the same value for a particular hyperparameter across all experiments (\textit{e.g.}, fixing the strength of an additional regularization to 1), we aimed to find the best hyperparameter for it (\textit{e.g.}, setting the strength as $\alpha$ and finding the best value of it in the hyperparameter tuning phase).
We determine the range of values for the predefined hyperparameters based on the following criteria. First, for general hyperparameters, we establish the range to include all optimal values reported by each algorithm. For specific hyperparameters related to each algorithm, we not only include the optimal values report in the papers but also considered the full range of values that were explored during their hyperparameter searches.

When the LR scheduler is set to StepLR, the milestones must be determined. To achieve this, we generalize the process of random sampling based on the milestones used in existing algorithms. First, we randomly sample num\_milestones. Based on this sampling, the milestones for the StepLR are set according to the following rule: For example, if Num\_milestones is set to 2, the milestones are defined as [epoch*(2/5), epoch*(4/5)]. If set to 3, the milestones become [epoch*(2/7), epoch*(4/7), epoch*(6/7)]. Similarly, for 4 milestones, the values are [epoch*(2/9), epoch*(4/9), epoch*(6/9), epoch*(8/9)].
However, note that the num\_milestones is ignored when another LR schduler is selected.

\begin{table}[!h]
\caption{The predefined set of hyperparametes for class-IL without a pretrained model.}
\centering
\smallskip\noindent
\resizebox{0.9\linewidth}{!}{
\begin{tabular}{|c|c||c|}
\hline
Algorithm                       & Hyperparameter Name                                           & $h^{Set}$                                                                                                         \\ \hline \hline
\multirow{7}{*}{All algorithms} & Epoch                                                         & {[}30, 70, 120, 160, 200{]}                                                                                       \\ \cline{2-3} 
                                & LR                                                            & {[}0.05, 0.1, 0.15, 0.2, 0.3{]}                                                                                   \\ \cline{2-3} 
                                & \begin{tabular}[c]{@{}c@{}}Num\\ milestones\end{tabular}      & {[}2, 3, 4{]}                                                                                                     \\ \cline{2-3} 
                                & \begin{tabular}[c]{@{}c@{}}LR\\ decay\end{tabular}            & {[}0.1, 0.3, 0.5{]}                                                                                               \\ \cline{2-3} 
                                & \begin{tabular}[c]{@{}c@{}}Batch\\ size\end{tabular}          & {[}32, 64, 128, 256, 512{]}                                                                                       \\ \cline{2-3} 
                                & \begin{tabular}[c]{@{}c@{}}Weigh\\ decay\end{tabular}         & {[}0.0001, 0.0005, 0.001, 0.005{]}                                                                                \\ \cline{2-3} 
                                & \begin{tabular}[c]{@{}c@{}}LR\\ Scheduler\end{tabular}        & {[}'StepLR', 'Cosine'{]}                                                                                          \\ \hline
iCaRL, BiC, WA and FOSTER       & \begin{tabular}[c]{@{}c@{}}T\\ (KD)\end{tabular}              & {[}0.5, 1, 1.5, 2, 2.5{]}                                                                                         \\ \hline
BiC, WA and FOSTER              & \begin{tabular}[c]{@{}c@{}}$\lambda$\\ (KD)\end{tabular}      & {[}0.5, 1, 1.5, 2, 3{]}                                                                                           \\ \hline
BiC                             & \begin{tabular}[c]{@{}c@{}}Split\\ ratio\end{tabular}         & {[}0.05, 0.1, 0.15, 0.2, 0.3{]}                                                                                   \\ \hline
iCaRL, PODNet, DER and MEMO     & \begin{tabular}[c]{@{}c@{}}$\lambda$\\ (Aux)\end{tabular}     & {[}0.5, 1, 1.5, 2, 3{]}                                                                                           \\ \hline
FOSTER                          & \begin{tabular}[c]{@{}c@{}}$\lambda$\\ (FE)\end{tabular}      & {[}0.5, 1, 1.5, 2, 3{]}                                                                                           \\ \hline
FOSTER                          & $\beta_1$                                                     & {[}0.93, 0.95, 0.97, 0.99{]}                                                                                      \\ \hline
FOSTER                          & $\beta_2$                                                     & {[}0.93, 0.95, 0.97, 0.99{]}                                                                                      \\ \hline
PODNet                          & \begin{tabular}[c]{@{}c@{}}Num\\ proxy\end{tabular}           & {[}10, 20, 30, 50, 100{]}                                                                                         \\ \hline
PODNet, FOSTER and BEEF         & \begin{tabular}[c]{@{}c@{}}Post FT\\ epochs\end{tabular}      & \begin{tabular}[c]{@{}c@{}}{[}5, 10, 20, 30, 50{]}\\ / {[}30, 70, 120, 160, 200{]} (FOSTER and BEEF)\end{tabular} \\ \hline
PODNet                          & \begin{tabular}[c]{@{}c@{}}Post FT\\ LR\end{tabular}          & {[}0.001, 0.003, 0.005, 0.007, 0.01{]}                                                                            \\ \hline
PODNet                          & Adaptive factor                                               & {[}True, False{]}    \\ \hline
BEEF                            & \begin{tabular}[c]{@{}c@{}}Energy\\ weight\end{tabular}       & {[}0.001, 0.005, 0.01, 0.02, 0.05{]}                                                                                                              \\ \hline
BEEF                            & \begin{tabular}[c]{@{}c@{}}Logit\\ alignment\end{tabular}     & {[}1.1, 1.4, 1.7, 2.0, 2.3{]}                                                                                     \\ \hline
MEMO                            & \begin{tabular}[c]{@{}c@{}}Exemplar\\ batch size\end{tabular} & {[}16, 32, 64, 128, 256{]}                                                                                        \\ \hline
\end{tabular}\label{table:predefined_hyperparameters}
}
\end{table}

\newpage

\noindent\textbf{Original hyperparameters} \ \ 
The following shows the original hyperparameters of each algorithm reported in PyCIL.
\begin{itemize}
  \item {Replay: ep\_70\_lr\_0.1\_lr\_decay\_0.1\_batch\_128\_w\_decay\_0.0002\_scheduler\_steplr}
\item {BiC:	ep\_170\_lr\_0.1\_lr\_decay\_0.1\_batch\_128\_w\_decay\_0.0002\_scheduler\_steplr \\ T\_2\_lambda\_kd\_0\_split\_ratio\_0.1}
\item {PODNet:	ep\_160\_milestone\_2\_lr\_0.1\_lr\_decay\_0.1\_batch\_128\_w\_decay\_0.0005\_scheduler\_cosine \\ lambda\_c\_5\_lambda\_f\_1.0\_nb\_proxy\_10\_ft\_epochs\_20\_ft\_lrate\_0.005\_adaptive\_factor\_True}
\item {FOSTER:	ep\_170\_lr\_0.1\_lr\_decay\_0.1\_batch\_128\_w\_decay\_0.0005\_scheduler\_cosine\\ T\_2\_lambda\_kd\_1\_fe\_1\_beta\_0.96\_0.97\_comp\_ep\_130}
\item {MEMO:	ep\_170\_milestone\_3\_lr\_0.1\_lr\_decay\_0.1\_batch\_128\_w\_decay\_0.0002\_scheduler\_steplr\\ lambda\_aux\_1\_examplar\_bs\_64}
\item {iCaRL:	ep\_170\_lr\_0.1\_lr\_decay\_0.1\_batch\_128\_w\_decay\_0.0002\_scheduler\_steplr\\ T\_2\_lambda\_aux\_1}
\item {WA:	ep\_170\_milestone\_3\_lr\_0.1\_lr\_decay\_0.1\_batch\_128\_w\_decay\_0.0002\_scheduler\_steplr\\ T\_2.0\_lambda\_kd\_0}
\item {DER:	ep\_170\_lr\_0.1\_lr\_decay\_0.1\_batch\_128\_w\_decay\_0.0002\_scheduler\_steplr\\ lambda\_aux\_1}
\item {BEEF:	ep\_170\_milestone\_4\_lr\_0.1\_lr\_decay\_0.1\_batch\_128\_w\_decay\_0.0005\_scheduler\_cosine\\ fusion\_ep\_60\_energy\_w\_0.01\_logits\_align\_1.7}
\end{itemize}

Note that setting `lambda\_kd = 0' for both BiC and WA indicates the use of their adaptive rule.

\newpage

\noindent\textbf{Best hyperparameters (ImageNet-100, 10 Tasks)} \ \ 
The following represents the best hyperparameters of each algorithm selected in the hyperparameter tuning phase using ImageNet-100 (10 Tasks).
\begin{itemize}
  \item {Replay:	ep\_70\_milestone\_3\_lr\_0.2\_lr\_decay\_0.1\_batch\_64\_w\_decay\_0.0001\_scheduler\_steplr}
\item {BiC:	ep\_120\_milestone\_3\_lr\_0.1\_lr\_decay\_0.1\_batch\_32\_w\_decay\_0.0001\_scheduler\_steplr\\ T\_1\_lambda\_kd\_3.0\_split\_ratio\_0.1}
\item {PODNet:	ep\_30\_milestone\_4\_lr\_0.05\_lr\_decay\_0.1\_batch\_64\_w\_decay\_0.0001\_scheduler\_steplr\\ lambda\_c\_3\_lambda\_f\_1.5\_nb\_proxy\_20\_ft\_epochs\_5\_ft\_lrate\_0.005\_adaptive\_factor\_False}
\item {FOSTER:	ep\_30\_milestone\_4\_lr\_0.05\_lr\_decay\_0.1\_batch\_64\_w\_decay\_0.0001\_scheduler\_steplr\\ T\_1\_lambda\_kd\_1.5\_fe\_1\_beta\_0.93\_0.97\_comp\_ep\_160}
\item {MEMO:	ep\_120\_milestone\_3\_lr\_0.15\_lr\_decay\_0.1\_batch\_512\_w\_decay\_0.001\_scheduler\_steplr\\ ambda\_aux\_0.5\_examplar\_bs\_32}
\item {iCaRL:	ep\_200\_milestone\_3\_lr\_0.15\_lr\_decay\_0.1\_batch\_64\_w\_decay\_0.0001\_scheduler\_cosine\\ T\_2.5\_lambda\_aux\_2}
\item {WA:	ep\_120\_milestone\_3\_lr\_0.1\_lr\_decay\_0.1\_batch\_32\_w\_decay\_0.0001\_scheduler\_steplr\\ T\_1\_lambda\_kd\_3.0\_split\_ratio\_0.1}
\item {DER:	ep\_200\_milestone\_3\_lr\_0.15\_lr\_decay\_0.1\_batch\_64\_w\_decay\_0.0001\_scheduler\_cosine\\ lambda\_aux\_3}
\item {BEEF:	ep\_120\_milestone\_2\_lr\_0.2\_lr\_decay\_0.3\_batch\_128\_w\_decay\_0.0001\_scheduler\_steplr\\ fusion\_ep\_30\_energy\_w\_0.02\_logits\_align\_2.3}

\end{itemize}

\noindent\textbf{Best hyperparameters (ImageNet-100, 6 Tasks)} \ \ 
The following represents the best hyperparameters of each algorithm selected in the hyperparameter tuning phase using ImageNet-100 (6 Tasks).
\begin{itemize}
  \item {Replay: ep\_70\_milestone\_3\_lr\_0.2\_lr\_decay\_0.1\_batch\_64\_w\_decay\_0.0001\_scheduler\_steplr}
\item {BiC: ep\_30\_milestone\_4\_lr\_0.05\_lr\_decay\_0.1\_batch\_64\_w\_decay\_0.0001\_scheduler\_steplr\\ T\_1\_lambda\_kd\_1.5\_split\_ratio\_0.1}
\item {PODNet: 	ep\_30\_milestone\_2\_lr\_0.15\_lr\_decay\_0.1\_batch\_128\_w\_decay\_0.001\_scheduler\_steplr\\ lambda\_c\_9\_lambda\_f\_0.5\_nb\_proxy\_100\_ft\_epochs\_10\_ft\_lrate\_0.007\_adaptive\_factor\_False}
\item {FOSTER: 	ep\_70\_milestone\_3\_lr\_0.05\_lr\_decay\_0.1\_batch\_512\_w\_decay\_0.0001\_scheduler\_cosine\\ T\_2.5\_lambda\_kd\_0.5\_fe\_3\_beta\_0.95\_0.93\_comp\_ep\_30}
\item {MEMO: 	ep\_120\_milestone\_3\_lr\_0.15\_lr\_decay\_0.1\_batch\_512\_w\_decay\_0.001\_scheduler\_steplr\\ lambda\_aux\_0.5\_examplar\_bs\_32}
\item {iCaRL: 	ep\_200\_milestone\_3\_lr\_0.15\_lr\_decay\_0.1\_batch\_64\_w\_decay\_0.0001\_scheduler\_cosine\\ T\_2.5\_lambda\_aux\_2}
\item {WA: 	ep\_170\_lr\_0.1\_lr\_decay\_0.1\_batch\_128\_w\_decay\_0.0002\_scheduler\_steplr\\ T\_2\_lambda\_kd\_1}
\item {DER: 	ep\_200\_milestone\_3\_lr\_0.15\_lr\_decay\_0.1\_batch\_64\_w\_decay\_0.0001\_scheduler\_cosine\\ lambda\_aux\_3}
\item {BEEF: 	ep\_30\_milestone\_4\_lr\_0.05\_lr\_decay\_0.1\_batch\_128\_w\_decay\_0.0001\_scheduler\_steplr\\ fusion\_ep\_70\_energy\_w\_0.01\_logits\_align\_1.4}

\end{itemize}

\noindent\textbf{Best hyperparameters (CIFAR-50, 10 Tasks)} \ \ 
The following represents the best hyperparameters of each algorithm selected in the hyperparameter tuning phase using CIFAR-50 (10 Tasks).
\begin{itemize}
  \item {Replay:	ep\_160\_milestone\_3\_lr\_0.15\_lr\_decay\_0.3\_batch\_32\_w\_decay\_0.0001\_scheduler\_cosine}
\item {BiC:	ep\_200\_milestone\_2\_lr\_0.1\_lr\_decay\_0.1\_batch\_32\_w\_decay\_0.0001\_scheduler\_cosine\\ T\_0.5\_lambda\_kd\_0.5\_split\_ratio\_0.2}
\item {PODNet:	ep\_70\_milestone\_2\_lr\_0.1\_lr\_decay\_0.1\_batch\_32\_w\_decay\_0.0001\_scheduler\_steplr\\ lambda\_c\_1\_lambda\_f\_1\_nb\_proxy\_10\_ft\_epochs\_30\_ft\_lrate\_0.007\_adaptive\_factor\_False}
\item {FOSTER:	ep\_120\_milestone\_3\_lr\_0.1\_lr\_decay\_0.5\_batch\_32\_w\_decay\_0.0001\_scheduler\_steplr\\ T\_2\_lambda\_kd\_1.5\_fe\_0.5\_beta\_0.97\_0.93\_comp\_ep\_160}
\item {MEMO:	ep\_120\_milestone\_4\_lr\_0.05\_lr\_decay\_0.3\_batch\_32\_w\_decay\_0.0005\_scheduler\_steplr\\ lambda\_aux\_0.5\_examplar\_bs\_16}
\item {iCaRL:	ep\_70\_milestone\_3\_lr\_0.05\_lr\_decay\_0.3\_batch\_32\_w\_decay\_0.001\_scheduler\_cosine\\ T\_2.5\_lambda\_aux\_1}
\item {WA:	ep\_160\_milestone\_4\_lr\_0.05\_lr\_decay\_0.1\_batch\_64\_w\_decay\_0.001\_scheduler\_cosine\\ T\_2\_lambda\_kd\_3}
\item {DER:	ep\_200\_milestone\_3\_lr\_0.2\_lr\_decay\_0.1\_batch\_256\_w\_decay\_0.001\_scheduler\_cosine\\ lambda\_aux\_2}
\item {BEEF:	ep\_200\_milestone\_3\_lr\_0.15\_lr\_decay\_0.1\_batch\_128\_w\_decay\_0.0001\_scheduler\_cosine\\ fusion\_ep\_200\_energy\_w\_0.02\_logits\_align\_2.3}
\end{itemize}

\noindent\textbf{Best hyperparameters (CIFAR-50, 6 Tasks)} \ \ 
The following represents the best hyperparameters of each algorithm selected in the hyperparameter tuning phase using CIFAR-50 (6 Tasks).
\begin{itemize}
  \item {Replay:	ep\_70\_milestone\_2\_lr\_0.05\_lr\_decay\_0.1\_batch\_32\_w\_decay\_0.0001\_scheduler\_cosine}
\item {BiC:	ep\_120\_milestone\_2\_lr\_0.05\_lr\_decay\_0.3\_batch\_32\_w\_decay\_0.0001\_scheduler\_cosine\\ T\_2.5\_lambda\_kd\_1.5\_split\_ratio\_0.3}
\item {PODNet:	ep\_30\_milestone\_3\_lr\_0.05\_lr\_decay\_0.5\_batch\_64\_w\_decay\_0.0005\_scheduler\_cosine\\ lambda\_c\_1\_lambda\_f\_3\_nb\_proxy\_30\_ft\_epochs\_50\_ft\_lrate\_0.003\_adaptive\_factor\_False}
\item {FOSTER:	ep\_70\_milestone\_2\_lr\_0.05\_lr\_decay\_0.1\_batch\_64\_w\_decay\_0.0005\_scheduler\_steplr\\ T\_1.5\_lambda\_kd\_1\_fe\_3\_beta\_0.97\_0.93\_comp\_ep\_200}
\item {MEMO:	ep\_160\_milestone\_4\_lr\_0.05\_lr\_decay\_0.1\_batch\_32\_w\_decay\_0.001\_scheduler\_cosine\\ lambda\_aux\_0.5\_examplar\_bs\_256}
\item {iCaRL:	ep\_120\_milestone\_2\_lr\_0.05\_lr\_decay\_0.1\_batch\_32\_w\_decay\_0.0005\_scheduler\_steplr\\ T\_1\_lambda\_aux\_1}
\item {WA:	ep\_160\_milestone\_3\_lr\_0.05\_lr\_decay\_0.1\_batch\_64\_w\_decay\_0.0001\_scheduler\_cosine\\ T\_2\_lambda\_kd\_1.5}
\item {DER:	ep\_120\_milestone\_3\_lr\_0.05\_lr\_decay\_0.5\_batch\_64\_w\_decay\_0.001\_scheduler\_cosine\\ lambda\_aux\_1.5}
\item {BEEF:	ep\_30\_milestone\_4\_lr\_0.05\_lr\_decay\_0.1\_batch\_128\_w\_decay\_0.0001\_scheduler\_steplr\\ fusion\_ep\_70\_energy\_w\_0.01\_logits\_align\_1.4}
\end{itemize}

\noindent\textbf{Best hyperparameters (ImageNet-50, 10 Tasks)} \ \ 
The following represents the best hyperparameters of each algorithm selected in the hyperparameter tuning phase using ImageNet-50 (10 Tasks).
\begin{itemize}
  \item {Replay:	ep\_70\_milestone\_3\_lr\_0.2\_lr\_decay\_0.1\_batch\_64\_w\_decay\_0.0001\_scheduler\_steplr}
\item {BiC:	ep\_120\_milestone\_3\_lr\_0.1\_lr\_decay\_0.1\_batch\_32\_w\_decay\_0.0001\_scheduler\_steplr\\ T\_1\_lambda\_kd\_3\_split\_ratio\_0.1}
\item {PODNet:	ep\_30\_milestone\_4\_lr\_0.2\_lr\_decay\_0.3\_batch\_64\_w\_decay\_0.0005\_scheduler\_cosine\\ lambda\_c\_7\_lambda\_f\_1\_nb\_proxy\_50\_ft\_epochs\_20\_ft\_lrate\_0.007\_adaptive\_factor\_True}
\item {FOSTER:	ep\_120\_milestone\_3\_lr\_0.1\_lr\_decay\_0.1\_batch\_32\_w\_decay\_0.0001\_scheduler\_steplr\\ T\_1\_lambda\_kd\_3\_fe\_1\_beta\_0.99\_0.93\_comp\_ep\_160}
\item {MEMO:	ep\_120\_milestone\_3\_lr\_0.1\_lr\_decay\_0.1\_batch\_32\_w\_decay\_0.0001\_scheduler\_steplr\\ lambda\_aux\_1\_examplar\_bs\_256}
\item {iCaRL:	ep\_200\_milestone\_3\_lr\_0.15\_lr\_decay\_0.1\_batch\_64\_w\_decay\_0.0001\_scheduler\_cosine\\ T\_2.5\_lambda\_aux\_2}
\item {WA:	ep\_200\_milestone\_3\_lr\_0.15\_lr\_decay\_0.1\_batch\_64\_w\_decay\_0.0001\_scheduler\_cosine\\ T\_2.5\_lambda\_kd\_2}
\item {DER:	ep\_200\_milestone\_3\_lr\_0.15\_lr\_decay\_0.1\_batch\_64\_w\_decay\_0.0001\_scheduler\_cosine\\ lambda\_aux\_3}
\item {BEEF	NaN}
\end{itemize}

\noindent\textbf{Best hyperparameters (ImageNet-50, 6 Tasks)} \ \ 
The following represents the best hyperparameters of each algorithm selected in the hyperparameter tuning phase using ImageNet-50 (6 Tasks).
\begin{itemize}
    \item {Replay:	ep\_200\_milestone\_2\_lr\_0.2\_lr\_decay\_0.3\_batch\_32\_w\_decay\_0.0001\_scheduler\_steplr}
\item {BiC:	ep\_120\_milestone\_3\_lr\_0.1\_lr\_decay\_0.1\_batch\_32\_w\_decay\_0.0001\_scheduler\_steplr\\ T\_1\_lambda\_kd\_3\_split\_ratio\_0.1}
\item {PODNet:	ep\_30\_milestone\_4\_lr\_0.2\_lr\_decay\_0.3\_batch\_64\_w\_decay\_0.0005\_scheduler\_cosine\\ lambda\_c\_7\_lambda\_f\_1\_nb\_proxy\_50\_ft\_epochs\_20\_ft\_lrate\_0.007\_adaptive\_factor\_True}
\item {FOSTER:	ep\_30\_milestone\_4\_lr\_0.2\_lr\_decay\_0.3\_batch\_64\_w\_decay\_0.0005\_scheduler\_cosine\\ T\_2\_lambda\_kd\_1\_fe\_2\_beta\_0.97\_0.99\_comp\_ep\_120}
\item {MEMO:	ep\_120\_milestone\_3\_lr\_0.15\_lr\_decay\_0.1\_batch\_512\_w\_decay\_0.001\_scheduler\_steplr\\ lambda\_aux\_0.5\_examplar\_bs\_32}
\item {iCaRL:	ep\_200\_milestone\_3\_lr\_0.15\_lr\_decay\_0.1\_batch\_64\_w\_decay\_0.0001\_scheduler\_cosine\\ T\_2.5\_lambda\_aux\_2}
\item {WA:	ep\_120\_milestone\_4\_lr\_0.1\_lr\_decay\_0.5\_batch\_64\_w\_decay\_0.0005\_scheduler\_steplr\\ T\_1\_lambda\_kd\_1}
\item {DER:	ep\_120\_milestone\_4\_lr\_0.3\_lr\_decay\_0.5\_batch\_128\_w\_decay\_0.0005\_scheduler\_cosine\\ lambda\_aux\_0.5}
\item {BEEF:	NaN}
\end{itemize}


\newpage

\subsection{Experimental settings for class-incremental learning with a pretrained model}\label{appendix:environment_pretrained}
\noindent\textbf{Experimental details} \ \ For experiments using the proposed evaluation protocol on class-incremental learning algorithms with a pretrained model, we employ the PILOT~\citep{sun2023pilot} code for each algorithm. The experimental setup closely followed PILOT's environment, using Python 3.8, PyTorch 2.0.1, and CUDA 11.7.

\begin{figure}[h]
\vspace*{-0.1in}
\centering
\includegraphics[width=0.6\textwidth]{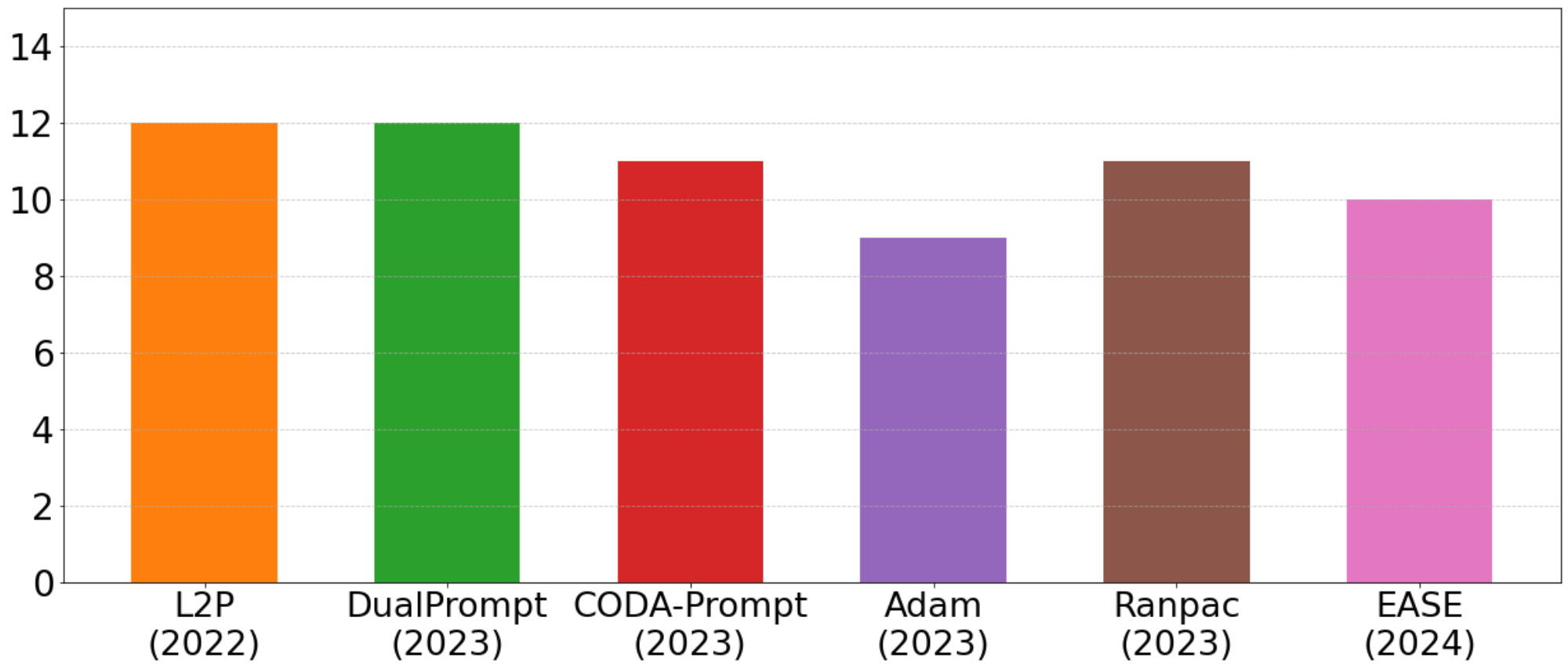}
\vspace*{-0.1in}
\caption{\# of hyperparameters.}\label{fig:hyperparameters_cil_pretrained}
\vspace{-.1in}
\end{figure}

\noindent\textbf{Pretrained hyperparameters} \ \ The process of selecting hyperparameters for algorithms using a pretrained model is similar to the previous experiments. We comprehensively consider both general hyperparameters and algorithm-specific ones, finding the best hyperparameters during the tuning phase. Figure \ref{fig:hyperparameters_cil_pretrained} shows the number of hyperparameters for each algorithm. The predefined hyperparameters used for this process are listed in Table \ref{table:predefined_hyperparameters_pretrained}. Using the selected hyperparameters, we train each algorithm across the entire CL scenario. The range of each hyperparameter is set based on values reported in previous work for each type of algorithm. Unlike the algorithms without pretrained models, which use the same optimizer (\textit{i.e.}, SGD), different optimizers have been used across algorithms in this case, so we also perform sampling for the optimizer. For hyperparameters of the optimizer that were not sampled, we use the default values provided in PyTorch.

\begin{table}[!h]
\caption{The predefined set of hyperparametes for class-IL with a pretrained model.}
\centering
\smallskip\noindent
\resizebox{0.99\linewidth}{!}{
\begin{tabular}{|c|c||c|}
\hline
Algorithm                       & Hyperparameter Name                                      & $h^{Set}$                                                                                                                                                                                                      \\ \hline \hline
\multirow{8}{*}{All algorithms} & Epoch                                                    & \begin{tabular}[c]{@{}c@{}}{[}3, 5, 10, 15, 20, 25{]} \\ (for L2P, DualPrompt. CODA-Pormpt) \\ / {[}5, 10, 15, 20, 25, 30{]} \\ (for Adam-Adapter, Ranpac, EASE)\end{tabular}                                  \\ \cline{2-3} 
                                & LR                                                       & \begin{tabular}[c]{@{}c@{}}{[}0.000875, 0.001375, 0.001875, 0.002375, 0.0025{]} \\ (for L2P, DualPrompt. CODA-Pormpt) \\ / {[}0.01, 0.02, 0.03, 0.04, 0.05{]} \\ (for Adam-Adapter, Ranpac, EASE)\end{tabular} \\ \cline{2-3} 
                                & \begin{tabular}[c]{@{}c@{}}Num\\ milestones\end{tabular} & {[}2, 3, 4{]}                                                                                                                                                                                                  \\ \cline{2-3} 
                                & \begin{tabular}[c]{@{}c@{}}LR\\ decay\end{tabular}       & {[}0.1, 0.3, 0.5{]}                                                                                                                                                                                            \\ \cline{2-3} 
                                & \begin{tabular}[c]{@{}c@{}}Batch\\ size\end{tabular}     & \begin{tabular}[c]{@{}c@{}}{[}8, 16, 24, 48, 64, 128{]} \\ (for L2P, DualPrompt, CODA-Prompt, Adam-Adapter\end{tabular}                                                                                        \\ \cline{2-3} 
                                & \begin{tabular}[c]{@{}c@{}}Weigh\\ decay\end{tabular}    & \begin{tabular}[c]{@{}c@{}}{[}0, 0.0001, 0.0005{]} \\ (for L2P, DualPrompt, CODA-Prompt) \\ / {[}0.0001, 0.0005, 0.001, 0.005{]} \\ (for Adam-Adapter, Ranpac, EASE)\end{tabular}                              \\ \cline{2-3} 
                                & \begin{tabular}[c]{@{}c@{}}LR\\ Scheduler\end{tabular}   & {[}'steplr', 'cosine', 'constant'{]}                                                                                                                                                                           \\ \cline{2-3} 
                                & Optimizer                                                & {[}'sgd', 'adam', 'adamw'{]}                                                                                                                                                                                   \\ \hline
L2P, DualPrompt                 & M Size                                                   & {[}10, 15, 20, 25, 30{]}                                                                                                                                                          \\ \hline
L2P                             & Length ($L_p$)                                           & {[}2, 4, 6, 8, 10{]}                                                                                                                                                                                           \\ \hline
L2P                             & Top k                                                    & {[}2, 4, 6, 8, 10{]}                                                                                                                                                                                           \\ \hline
L2P, DualPrompt                 & $\lambda$                                                & {[}0.1, 0.3, 0.5{]}                                                                                                                                                                                            \\ \hline
DualPrompt                      & Prompt length of g ($L_g$)                               & {[}5, 10, 15, 20, 30{]}                                                                                                                                                                                        \\ \hline
DualPrompt                      & Length ($L_e$)                                           & {[}5, 10, 15, 20, 30{]}                                                                                                                                                                                        \\ \hline
CODA-Prompt                     & Pool size                                                & {[}30, 50, 100, 200, 300{]}                                                                                                                                                                                    \\ \hline
CODA-Prompt                     & Prompt length                                            & {[}4, 8, 16, 24, 32{]}                                                                                                                                                                                         \\ \hline
CODA-Prompt                     & Orthogonality Mu                                         & {[}0.2, 0.1, 0.01, 0.001, 0{]}                                                                                                                                                                                 \\ \hline
Adam-Adapter, Ranpac, EASE      & FFN num                                                  & {[}4,8,16,32,64{]}                                                                                                                                                                                             \\ \hline
Ranpac                          & M                                                        & {[}5000, 10000, 15000, 20000{]}                                                                                                                                                                                \\ \hline
Ranpac                          & Prompt token num                                         & {[}3, 5, 10, 20, 30, 50{]}                                                                                                                                                                                     \\ \hline
EASE                            & $\alpha$                                                 & {[}0.01, 0.05, 0.1, 0.15, 0.2{]}                                                                                                                                                                               \\ \hline
\end{tabular}\label{table:predefined_hyperparameters_pretrained}
}
\end{table}

\newpage

\noindent\textbf{Original hyperparameters} \ \ 
The following shows the original hyperparameters of each algorithm reported in PILOT.
\begin{itemize}
  \item {L2P\_ep\_10\_milestone\_3\_lr\_0.001875\_lr\_decay\_0\_batch\_32\_w\_decay\_0\\ scheduler\_constant\_optimizer\_adam\_size\_10\_length\_5\_top\_k\_5\_lamb\_0.1  }
\item {DualPrompt\_ep\_10\_milestone\_4\_lr\_0.001\_lr\_decay\_0.0\_batch\_24\_w\_decay\_0.0\\ scheduler\_constant\_optimizer\_adam\_size\_10\_L\_e\_5\_L\_g\_5\_top\_k\_1\_lamb\_0.1  }
\item CODA-{Prompt\_ep\_50\_milestone\_2\_lr\_0.001\_lr\_decay\_0.0\_batch\_128\_w\_decay\_0.0\\ scheduler\_cosine\_optimizer\_adam\_e\_pool\_size\_100\_e\_p\_length\_8\_ortho\_mu\_0.0  }
\item {Adam\_ep\_10\_milestone\_3\_lr\_0.05\_lr\_decay\_0.0\_batch\_16\_w\_decay\_0.005\\ scheduler\_constant\_optimizer\_sgd\_ffn\_num\_100  }
\item {Ranpac\_ep\_10\_milestone\_2\_lr\_0.05\_lr\_decay\_0.0\_batch\_16\_w\_decay\_0.005\\ scheduler\_constant\_optimizer\_sgd\_ffn\_num\_64\_M\_10000\_pt\_num\_30  }
\item {EASE\_ep\_20\_milestone\_4\_lr\_0.05\_lr\_decay\_0.0\_batch\_16\_w\_decay\_0.005\\ scheduler\_cosine\_optimizer\_sgd\_ffn\_num\_64\_alpha\_0.1}
\end{itemize}

\noindent\textbf{Best hyperparameters (CUB-200, 20 Tasks)} \ \ 
The following represents the best hyperparameters of each algorithm selected in the hyperparameter tuning phase using CUB-200 (20 Tasks).
\begin{itemize}
  \item {L2P: ep\_20\_milestone\_2\_lr\_0.002375\_lr\_decay\_0.5\_batch\_64\_w\_decay\_0.0001\\ scheduler\_constant\_optimizer\_adamw\_size\_15\_length\_6\_top\_k\_4\_lamb\_0.1  }
\item {DualPrompt: ep\_25\_milestone\_3\_lr\_0.000875\_lr\_decay\_0.1\_batch\_48\_w\_decay\_0.0005\\ scheduler\_constant\_optimizer\_adamw\_size\_20\_L\_e\_5\_L\_g\_30\_top\_k\_1\_lamb\_0.3  }
\item {CODA-Prompt: ep\_25\_milestone\_2\_lr\_0.000875\_lr\_decay\_0.3\_batch\_24\_w\_decay\_0.0005\\ scheduler\_steplr\_optimizer\_sgd\_e\_pool\_size\_30\_e\_p\_length\_4\_ortho\_mu\_0.01  }
\item {Adam: ep\_15\_milestone\_4\_lr\_0.05\_lr\_decay\_0.5\_batch\_48\_w\_decay\_0.0005\\ scheduler\_cosine\_optimizer\_sgd\_ffn\_num\_8  }
\item {Ranpac: ep\_30\_milestone\_4\_lr\_0.01\_lr\_decay\_0.1\_batch\_8\_w\_decay\_0.0005\\ scheduler\_cosine\_optimizer\_sgd\_ffn\_num\_32\_M\_20000\_pt\_num\_5  }
\item {EASE: ep\_15\_milestone\_4\_lr\_0.02\_lr\_decay\_0.5\_batch\_128\_w\_decay\_0.001\\ scheduler\_cosine\_optimizer\_sgd\_ffn\_num\_8\_alpha\_0.01}
\end{itemize}

\noindent\textbf{Best hyperparameters (CUB-200, 10 Tasks)} \ \ 
The following represents the best hyperparameters of each algorithm selected in the hyperparameter tuning phase using CUB-200 (10 Tasks).
\begin{itemize}
  \item {L2P: ep\_25\_milestone\_2\_lr\_0.001875\_lr\_decay\_0.3\_batch\_64\_w\_decay\_0.0005\\ scheduler\_cosine\_optimizer\_adamw\_size\_10\_length\_6\_top\_k\_6\_lamb\_0.5  }
\item {DualPrompt: ep\_25\_milestone\_3\_lr\_0.0025\_lr\_decay\_0.5\_batch\_128\_w\_decay\_0.0005\\ scheduler\_steplr\_optimizer\_sgd\_size\_20\_L\_e\_10\_L\_g\_10\_top\_k\_1\_lamb\_0.5  }
\item {CODA-Prompt: ep\_25\_milestone\_3\_lr\_0.0025\_lr\_decay\_0.3\_batch\_64\_w\_decay\_0\\ scheduler\_cosine\_optimizer\_adamw\_e\_pool\_size\_100\_e\_p\_length\_8\_ortho\_mu\_0  }
\item {Adam: ep\_20\_milestone\_3\_lr\_0.04\_lr\_decay\_0.3\_batch\_8\_w\_decay\_0.0005\\ scheduler\_steplr\_optimizer\_sgd\_ffn\_num\_32  }
\item {Ranpac: ep\_30\_milestone\_4\_lr\_0.02\_lr\_decay\_0.3\_batch\_16\_w\_decay\_0.0001\\ scheduler\_steplr\_optimizer\_sgd\_ffn\_num\_64\_M\_10000\_pt\_num\_3  }
\item {EASE: ep\_15\_milestone\_4\_lr\_0.01\_lr\_decay\_0.3\_batch\_64\_w\_decay\_0.0005\\ scheduler\_steplr\_optimizer\_sgd\_ffn\_num\_8\_alpha\_0.05
}
\end{itemize}

\noindent\textbf{Best hyperparameters (ImageNet-R, 20 Tasks)} \ \ 
The following represents the best hyperparameters of each algorithm selected in the hyperparameter tuning phase using ImageNet-R (20 Tasks).
\begin{itemize}
  \item {L2P\_ep\_25\_milestone\_3\_lr\_0.000875\_lr\_decay\_0.5\_batch\_64\_w\_decay\_0\\ scheduler\_steplr\_optimizer\_adam\_size\_10\_length\_10\_top\_k\_4\_lamb\_0.5 } 
\item {DualPrompt: ep\_15\_milestone\_4\_lr\_0.001875\_lr\_decay\_0.5\_batch\_128\_w\_decay\_0\\ scheduler\_steplr\_optimizer\_adam\_size\_20\_L\_e\_30\_L\_g\_5\_top\_k\_1\_lamb\_0.5  }
\item {CODA-Prompt: ep\_15\_milestone\_2\_lr\_0.002375\_lr\_decay\_0.1\_batch\_48\_w\_decay\_0.0001\\ scheduler\_cosine\_optimizer\_adamw\_e\_pool\_size\_300\_e\_p\_length\_32\_ortho\_mu\_0.001  }
\item {Adam: ep\_25\_milestone\_3\_lr\_0.05\_lr\_decay\_0.5\_batch\_64\_w\_decay\_0.001\\ scheduler\_constant\_optimizer\_sgd\_ffn\_num\_64  }
\item {Ranpac: ep\_20\_milestone\_2\_lr\_0.05\_lr\_decay\_0.3\_batch\_24\_w\_decay\_0.0005\\ scheduler\_constant\_optimizer\_sgd\_ffn\_num\_16\_M\_15000\_pt\_num\_20  }
\item {EASE: ep\_15\_milestone\_4\_lr\_0.04\_lr\_decay\_0.1\_batch\_24\_w\_decay\_0.0001\\ scheduler\_constant\_optimizer\_sgd\_ffn\_num\_16\_alpha\_0.15}
\end{itemize}

\noindent\textbf{Best hyperparameters (ImageNet-R, 10 Tasks)} \ \ 
The following represents the best hyperparameters of each algorithm selected in the hyperparameter tuning phase using ImageNet-R (10 Tasks).
\begin{itemize}
  \item {L2P: ep\_25\_milestone\_3\_lr\_0.001375\_lr\_decay\_0.5\_batch\_128\_w\_decay\_0\\ scheduler\_constant\_optimizer\_adamw\_size\_20\_length\_6\_top\_k\_10\_lamb\_0.3  }
\item {DualPrompt: ep\_25\_milestone\_2\_lr\_0.001375\_lr\_decay\_0.3\_batch\_128\_w\_decay\_0.0005\\ scheduler\_constant\_optimizer\_adamw\_size\_30\_L\_e\_20\_L\_g\_20\_top\_k\_1\_lamb\_0.3  }
\item {CODA-Prompt: ep\_20\_milestone\_2\_lr\_0.001375\_lr\_decay\_0.1\_batch\_48\_w\_decay\_0\\ scheduler\_steplr\_optimizer\_adam\_e\_pool\_size\_300\_e\_p\_length\_8\_ortho\_mu\_0  }
\item {Adam: ep\_30\_milestone\_2\_lr\_0.05\_lr\_decay\_0.1\_batch\_64\_w\_decay\_0.001\\ scheduler\_cosine\_optimizer\_sgd\_ffn\_num\_32  }
\item {Ranpac: ep\_20\_milestone\_3\_lr\_0.03\_lr\_decay\_0.1\_batch\_24\_w\_decay\_0.0001\\ scheduler\_steplr\_optimizer\_sgd\_ffn\_num\_64\_M\_20000\_pt\_num\_20  }
\item {EASE: ep\_30\_milestone\_4\_lr\_0.05\_lr\_decay\_0.3\_batch\_128\_w\_decay\_0.001\\ scheduler\_cosine\_optimizer\_adam\_ffn\_num\_16}
\end{itemize}

\noindent\textbf{Best hyperparameters (CUB-100-1, 20 Tasks)} \ \ 
The following represents the best hyperparameters of each algorithm selected in the hyperparameter tuning phase using CUB-100-1 (20 Tasks).
\begin{itemize}
  \item {L2P: ep\_20\_milestone\_3\_lr\_0.002375\_lr\_decay\_0.3\_batch\_128\_w\_decay\_0.0005\\ scheduler\_constant\_optimizer\_adamw\_size\_20\_length\_8\_top\_k\_4\_lamb\_0.5  }
\item {DualPrompt: ep\_25\_milestone\_4\_lr\_0.001375\_lr\_decay\_0.1\_batch\_128\_w\_decay\_0\\ scheduler\_constant\_optimizer\_adam\_size\_15\_L\_e\_15\_L\_g\_20\_top\_k\_1\_lamb\_0.5  }
\item {CODA-Prompt: ep\_10\_milestone\_4\_lr\_0.0025\_lr\_decay\_0.3\_batch\_64\_w\_decay\_0\\ scheduler\_constant\_optimizer\_adam\_e\_pool\_size\_200\_e\_p\_length\_4\_ortho\_mu\_0.001  }
\item {Adam: ep\_5\_milestone\_2\_lr\_0.05\_lr\_decay\_0.5\_batch\_16\_w\_decay\_0.0005\\ scheduler\_constant\_optimizer\_sgd\_ffn\_num\_4  }
\item {Ranpac: ep\_15\_milestone\_4\_lr\_0.04\_lr\_decay\_0.1\_batch\_128\_w\_decay\_0.0001\\ scheduler\_cosine\_optimizer\_sgd\_ffn\_num\_4\_M\_15000\_pt\_num\_30  }
\item {EASE: ep\_10\_milestone\_4\_lr\_0.01\_lr\_decay\_0.1\_batch\_16\_w\_decay\_0.001\\ scheduler\_constant\_optimizer\_sgd\_ffn\_num\_32\_alpha\_0.05}
\end{itemize}

\noindent\textbf{Best hyperparameters (CUB-100-1, 10 Tasks)} \ \ 
The following represents the best hyperparameters of each algorithm selected in the hyperparameter tuning phase using CUB-100-1 (10 Tasks).
\begin{itemize}
  \item {L2P: ep\_25\_milestone\_2\_lr\_0.0025\_lr\_decay\_0.3\_batch\_128\_w\_decay\_0\\ scheduler\_cosine\_optimizer\_adam\_size\_20\_length\_4\_top\_k\_6\_lamb\_0.5} 
\item {DualPrompt: ep\_25\_milestone\_4\_lr\_0.002375\_lr\_decay\_0.3\_batch\_128\_w\_decay\_0.0005\\ scheduler\_cosine\_optimizer\_adamw\_size\_15\_L\_e\_30\_L\_g\_15\_top\_k\_1\_lamb\_0.5  }
\item {CODA-Prompt: ep\_25\_milestone\_3\_lr\_0.001375\_lr\_decay\_0.1\_batch\_64\_w\_decay\_0.0001\\ scheduler\_cosine\_optimizer\_adamw\_e\_pool\_size\_50\_e\_p\_length\_4\_ortho\_mu\_0  }
\item {Adam: ep\_25\_milestone\_2\_lr\_0.05\_lr\_decay\_0.3\_batch\_24\_w\_decay\_0.0001\\ scheduler\_steplr\_optimizer\_sgd\_ffn\_num\_32  }
\item {Ranpac: ep\_25\_milestone\_3\_lr\_0.02\_lr\_decay\_0.3\_batch\_16\_w\_decay\_0.0005\\ scheduler\_constant\_optimizer\_sgd\_ffn\_num\_16\_M\_10000\_pt\_num\_20  }
\item {EASE: ep\_15\_milestone\_3\_lr\_0.03\_lr\_decay\_0.5\_batch\_128\_w\_decay\_0.0001\\ scheduler\_constant\_optimizer\_sgd\_ffn\_num\_64\_alpha\_0.05}
\end{itemize}

\noindent\textbf{Best hyperparameters (ImageNet-R-1, 20 Tasks)} \ \ 
The following represents the best hyperparameters of each algorithm selected in the hyperparameter tuning phase using ImageNet-R-1 (20 Tasks).
\begin{itemize}
  \item {L2P: ep\_15\_milestone\_2\_lr\_0.002375\_lr\_decay\_0.5\_batch\_48\_w\_decay\_0\\ scheduler\_cosine\_optimizer\_adamw\_size\_25\_length\_6\_top\_k\_10\_lamb\_0.3}  
\item {DualPrompt: ep\_20\_milestone\_3\_lr\_0.001875\_lr\_decay\_0.1\_batch\_128\_w\_decay\_0\\ scheduler\_steplr\_optimizer\_adam\_size\_10\_L\_e\_30\_L\_g\_30\_top\_k\_1\_lamb\_0.1  }
\item {CODA-Prompt: ep\_15\_milestone\_2\_lr\_0.002375\_lr\_decay\_0.5\_batch\_64\_w\_decay\_0.0001\\ scheduler\_cosine\_optimizer\_adamw\_e\_pool\_size\_100\_e\_p\_length\_4\_ortho\_mu\_0.01  }
\item {Adam: ep\_25\_milestone\_3\_lr\_0.04\_lr\_decay\_0.3\_batch\_24\_w\_decay\_0.001\\ scheduler\_constant\_optimizer\_sgd\_ffn\_num\_16  }
\item {Ranpac: ep\_10\_milestone\_2\_lr\_0.02\_lr\_decay\_0.3\_batch\_8\_w\_decay\_0.0001\\ scheduler\_cosine\_optimizer\_sgd\_ffn\_num\_8\_M\_20000\_pt\_num\_10  }
\item {EASE: ep\_15\_milestone\_4\_lr\_0.03\_lr\_decay\_0.5\_batch\_16\_w\_decay\_0.0005\\ scheduler\_constant\_optimizer\_sgd\_ffn\_num\_64\_alpha\_0.05}
\end{itemize}

\noindent\textbf{Best hyperparameters (ImageNet-R-1, 10 Tasks)} \ \ 
The following represents the best hyperparameters of each algorithm selected in the hyperparameter tuning phase using ImageNet-R-1 (10 Tasks).
\begin{itemize}
  \item {L2P: ep\_20\_milestone\_3\_lr\_0.000875\_lr\_decay\_0.1\_batch\_24\_w\_decay\_0.0001\\ scheduler\_cosine\_optimizer\_adamw\_size\_20\_length\_8\_top\_k\_10\_lamb\_0.5  }
\item {DualPrompt: ep\_25\_milestone\_4\_lr\_0.000875\_lr\_decay\_0.3\_batch\_64\_w\_decay\_0\\ scheduler\_cosine\_optimizer\_adam\_size\_15\_L\_e\_30\_L\_g\_20\_top\_k\_1\_lamb\_0.1  }
\item {CODA-Prompt: ep\_15\_milestone\_3\_lr\_0.001375\_lr\_decay\_0.5\_batch\_64\_w\_decay\_0.0005\\ scheduler\_constant\_optimizer\_adamw\_e\_pool\_size\_300\_e\_p\_length\_4\_ortho\_mu\_0.01  }
\item {Adam: ep\_25\_milestone\_3\_lr\_0.04\_lr\_decay\_0.3\_batch\_24\_w\_decay\_0.001\\ scheduler\_constant\_optimizer\_sgd\_ffn\_num\_16  }
\item {Ranpac: ep\_25\_milestone\_4\_lr\_0.05\_lr\_decay\_0.1\_batch\_24\_w\_decay\_0.0001\\ scheduler\_constant\_optimizer\_sgd\_ffn\_num\_64\_M\_20000\_pt\_num\_10  }
\item {EASE: ep\_10\_milestone\_4\_lr\_0.04\_lr\_decay\_0.1\_batch\_24\_w\_decay\_0.001\\ scheduler\_constant\_optimizer\_sgd\_ffn\_num\_64\_alpha\_0.2}
\end{itemize}

\newpage

\section{Additional Experimental Results on the Evaluation Phase}\label{appendix:additional_results}

\subsection{Discussion on the number of hyperparameters}

\tmlrcsm{We believe that the number of hyperparameters can influence the evaluation protocol proposed in this paper. This is because a larger number of hyperparameters allows an algorithm to fit more aggressively during the hyperparameter tuning phase, making it easier to achieve higher performance. However, such overfitting can lead to poor generalization in the evaluation phase. This trend can be observed in Figure \ref{figure:compare_cil}. Among the evaluated algorithms, Foster has the largest number of hyperparameters (see Figure \ref{fig:hyperparameters_cil}). The experimental results show that while Foster reports outstanding (overfitted) performance during the hyperparameter tuning phase, its performance significantly deteriorates in the actual evaluation phase. We believe this finding further underscores the necessity of our proposed protocol for properly assessing each CL algorithm.}

\subsection{Instability issues with BEEF baseline}

\tmlrcsm{During our experiments with the BEEF baseline, we encountered persistent NaN issues when training on ImageNet-scale datasets using ResNet-18. First, we verified the hyperparameters used in our experiments to ensure they matched those specified in the original BEEF paper (Section C.2). Specifically, we confirmed that the learning rate of 0.1 (with a StepLR scheduler) and a mini-batch size of 256 were consistent with the settings in the original experiments. After confirming that there were no discrepancies in the hyperparameters, we investigated community reports related to the issue. In a relevant thread on the PyCIL repository, another user had reported similar NaN problems and suggested lowering the learning rate as a potential solution. Following this advice, we conducted additional experiments using lower learning rates, testing values of [0.001, 0.005, 0.01, 0.015, 0.02]. Despite these adjustments, none of the configurations avoided NaN results across all seeds. Representative results for two configurations that showed relatively better performance are presented in Table \ref{appendix:beef_table}. These findings suggest that the NaN issue is not due to hyperparameter misconfiguration but rather an intrinsic instability in the BEEF algorithm. }

\begin{table}[ht]
\centering
\begin{tabular}{|c||c|c|c|c|c|c|}
\hline
{Hyperparameters} & {Acc / AvgAcc} & {Seed 0} & {Seed 1} & {Seed 2} & {Seed 3} & {Seed 4} \\
\hline \hline
{BEEF (HP1)} & 49.52 / 65.57 & 49.24 / 59.22 & NaN & NaN & NaN & NaN \\
\hline
{BEEF (HP2)} & 48.62 / 65.14 & 46.40 / 58.31 & NaN & NaN & NaN & NaN \\
\hline
\end{tabular}
\caption{Performance of BEEF with different hyperparameters on ImageNet-100 dataset.}\label{appendix:beef_table}
\end{table}

\tmlrcsm{Further investigation into the BEEF code within the PyCIL implementation revealed that the instability originates from the adversarial learning process used by BEEF. The process generates adversarial examples to compute the energy loss, which leads to extreme value growth in the feature maps. Specifically, the feature maps from the copied model (\_network\_copy) undergo unbounded amplification due to adversarial samples, ultimately resulting in NaN values during training. This instability appears to be an algorithmic flaw rather than a minor numerical issue. To mitigate this, we suggest applying regularization techniques, such as L2 or L1 regularization, during the adversarial example generation process. However, we acknowledge that such regularization may affect the performance of BEEF, as its reported success on CIFAR and ImageNet datasets could depend on the current unregulated adversarial process. Given that BEEF exhibits instability across multiple seeds and hyperparameter configurations, we conclude that the NaN issues are an inherent limitation of the algorithm, particularly when applied to large-scale datasets like ImageNet. As a result, we report these findings to ensure transparency and will continue to investigate potential solutions in future work.}

\subsection{Result tables}

\noindent\textbf{Class-IL without a pretrained model ($D^{HT}$ = \text{ImageNet-100-1})}

\begin{table}[h!]
\caption{The experimental results of class-IL without a pretrained model (using original hyperparameters). \tmlrcsm{Each result represents the Final Accuracy and Average Accuracy (Acc / AvgAcc).} The values in parentheses represent the standard deviation.}
\centering
\smallskip\noindent
\resizebox{.8\linewidth}{!}{
\begin{tabular}{|c||c|c|}
\hline
\begin{tabular}[c]{@{}c@{}}10 Tasks\\ (Acc / AvgAcc)\end{tabular} & \multicolumn{1}{l|}{$D^{HT}$ = ImageNet-100-1} & \multicolumn{1}{l|}{$D^{E}$ = ImageNet-100-2} \\ \hline \hline
Replay                                                            & 41.21(1.06) / 59.82(1.48)                      & 41.00(1.46) / 61.73(1.32)                        \\ \hline
iCaRL                                                             & 40.50(1.19) / 60.12(1.41)                      & 41.24(1.31) / 62.87(1.43)                        \\ \hline
BiC                                                               & 39.61(2.39) / 64.27(1.59)                      & 37.62(3.73) / 66.25(1.17)                        \\ \hline
WA                                                                & 53.34(1.39) / 68.92(1.54)                      & 57.81(1.01) / 73.68(1.72)                        \\ \hline
PODNet                                                            & 46.66(1.11) / 64.13(1.20)                      & 48.86(1.15) / 67.38(1.92)                        \\ \hline
DER                                                               & 61.96(1.04) / 72.10(1.41)                      & 65.73(1.06) / 76.12(1.10)                        \\ \hline
FOSTER                                                            & 60.68(0.71) / 69.97(1.70)                      & 63.93(1.06) / 73.95(1.50)                        \\ \hline
BEEF                                                              & NaN                                            & NaN                                              \\ \hline
MEMO                                                              & 59.59(1.29) / 70.04(1.62)                      & 63.42(0.58) / 75.25(1.21)                        \\ \hline
\end{tabular}
}
\end{table}


\begin{table}[h!]
\caption{{The experimental results of class-IL without a pretrained model (using $D^{HT}$ = \text{ImageNet-100-1}) in the hyperparameter tuning phase.} \tmlrcsm{Each result represents the Final Accuracy and Average Accuracy (Acc / AvgAcc).} The values in parentheses represent the standard deviation.}
\centering
\smallskip\noindent
\resizebox{.8\linewidth}{!}{
\begin{tabular}{|c||c|c|}
\hline
\begin{tabular}[c]{@{}c@{}}10 Tasks\\ (Acc / AvgAcc)\end{tabular} & $D^{HT}$ = ImageNet-100-1              & $D^{E}$ = ImageNet-100-2              \\ \hline \hline
Replay                                                            & 44.78(1.19) / 59.85(0.95) & 44.27(1.05) / 61.49(0.87) \\ \hline
iCaRL                                                             & 42.58(1.06) / 61.27(1.26) & 42.44(1.50) / 63.39(1.18) \\ \hline
BiC                                                               & 54.22(1.27) / 67.31(0.74) & 58.77(0.96) / 71.81(1.42) \\ \hline
WA                                                                & 54.67(0.60) / 69.54(1.41) & 59.89(1.18) / 72.93(1.94) \\ \hline
PODNet                                                            & 55.35(0.93) / 68.74(1.52) & 57.48(0.94) / 71.76(1.62) \\ \hline
DER                                                               & 63.31(0.42) / 72.93(0.87) & 70.23(0.46) / 77.12(1.20) \\ \hline
FOSTER                                                            & 58.36(0.85) / 71.99(0.98) & 61.46(0.98) / 68.41(1.23) \\ \hline
BEEF                                                              & NaN                       & NaN                       \\ \hline
MEMO                                                              & 57.91(0.54) / 71.25(1.41) & 61.94(0.78) / 71.35(2.17) \\ \hline
\end{tabular}
}
\end{table}

\noindent\textbf{Performance comparison between original and our found Hyperparameters} \ \
\tmlrcsm{Based on the above experimental results, we can make the following observations. First, some algorithms achieve better performance with the original hyperparameters (\textit{e.g.}, FOSTER and MEMO). However, in contrast, there are cases where the best hyperparameters we identified lead to better performance. These results demonstrate that using the original hyperparameters does not always guarantee optimal performance across all CL scenarios. Furthermore, they show that performing hyperparameter tuning using the evaluation protocol we propose is a more appropriate and practical approach for such CL scenarios.}

\begin{table}[h!]
\caption{{The experimental results of class-IL without a pretrained model (using $D^{HT}$ = \text{ImageNet-100-1}) in the hyperparameter tuning phase.} \tmlrcsm{Each result represents the Final Accuracy and Average Accuracy (Acc / AvgAcc).} The values in parentheses represent the standard deviation.}
\centering
\smallskip\noindent
\resizebox{.8\linewidth}{!}{
\begin{tabular}{|c||c|c|}
\hline
\begin{tabular}[c]{@{}c@{}}6 Tasks\\ (Acc / AvgAcc)\end{tabular} & $D^{HT}$ = ImageNet-100   & $D^{E}$ = ImageNet-200    \\ \hline \hline
Replay                                                           & 42.93(2.41) / 53.81(1.72) & 43.26(1.38) / 49.28(0.53) \\ \hline
iCaRL                                                            & 46.62(1.54) / 57.27(0.73) & 45.64(1.49) / 59.18(0.54) \\ \hline
BiC                                                              & 37.14(1.62) / 36.42(1.89) & 38.43(2.53) / 40.89(3.07) \\ \hline
WA                                                               & 58.72(1.02) / 65.58(1.55) & 60.58(1.35) / 69.47(1.71) \\ \hline
PODNet                                                           & 67.22(0.67) / 75.05(1.16) & 65.51(1.83) / 75.82(1.03) \\ \hline
DER                                                              & 72.20(0.51) / 77.68(1.08) & 75.83(0.64) / 81.19(0.70) \\ \hline
FOSTER                                                           & 69.48(0.50) / 74.59(1.18) & 71.62(1.08) / 78.29(1.14) \\ \hline
BEEF                                                             & 74.67(0.14) / 78.92(0.54) & 75.09(0.29) / 81.31(0.50) \\ \hline
MEMO                                                             & 59.91(0.87) / 67.22(1.63) & 62.80(3.16) / 68.77(6.26) \\ \hline
\end{tabular}
}
\end{table}

\newpage

\noindent\textbf{Class-IL without a pretrained model ($D^{HT}$ = \text{CIFAR-50-1})}

\begin{table}[h!]
\caption{The experimental results of class-IL without a pretrained model (using $D^{HT}$ = \text{CIFAR-50-1}) in the hyperparameter tuning phase. \tmlrcsm{Each result represents the Final Accuracy and Average Accuracy (Acc / AvgAcc).} The values in parentheses represent the standard deviation.}
\centering
\smallskip\noindent
\resizebox{.8\linewidth}{!}{
\begin{tabular}{|c||c|c|}
\hline
\begin{tabular}[c]{@{}c@{}}10 Tasks\\ (Acc / AvgAcc)\end{tabular} & $D^{E}$ = CIFAR-50-2      & $D^{E}$ = ImageNet-50-2   \\ \hline \hline
Replay                                                            & 45.42(2.19) / 65.88(1.97) & 42.51(0.47) / 60.72(1.58) \\ \hline
iCaRL                                                             & 47.12(2.80) / 66.71(2.07) & 42.44(1.00) / 61.55(1.64) \\ \hline
BiC                                                               & 52.83(2.83) / 69.16(2.30) & 49.52(1.16) / 67.09(1.74) \\ \hline
WA                                                                & 54.89(2.13) / 69.85(2.32) & 53.64(1.47) / 67.75(1.90) \\ \hline
PODNet                                                            & 51.20(1.76) / 69.47(0.13) & 51.70(1.19) / 67.86(1.67) \\ \hline
DER                                                               & 63.51(1.98) / 75.04(1.24) & 63.40(1.02) / 72.67(1.62) \\ \hline
FOSTER                                                            & 60.00(2.72) / 72.29(2.09) & 62.09(1.83) / 70.24(1.50) \\ \hline
BEEF                                                              & 57.24(1.48) / 72.26(2.05) & NaN                       \\ \hline
MEMO                                                              & 60.72(2.41) / 73.78(1.99) & 54.91(1.59) / 68.06(2.10) \\ \hline
\end{tabular}
}
\end{table}

\begin{table}[h!]
\caption{The experimental results of class-IL without a pretrained model (using $D^{HT}$ = \text{CIFAR-50-1}) in the hyperparameter tuning phase. \tmlrcsm{Each result represents the Final Accuracy and Average Accuracy (Acc / AvgAcc).} The values in parentheses represent the standard deviation.}
\centering
\smallskip\noindent
\resizebox{.8\linewidth}{!}{
\begin{tabular}{|c||c|c|}
\hline
\begin{tabular}[c]{@{}c@{}}6 Tasks\\ (Acc / AvgAcc)\end{tabular} & $D^{E}$ = CIFAR-50-2      & $D^{E}$ = ImageNet-50-2   \\ \hline \hline
Replay                                                           & 48.00(1.98) / 59.86(1.03) & 46.30(1.31) / 55.67(0.64) \\ \hline
iCaRL                                                            & 46.09(1.51) / 59.14(1.39) & 46.21(1.72) / 57.79(1.06) \\ \hline
BiC                                                              & 58.22(1.20) / 68.16(1.96) & 46.26(3.26) / 59.07(3.87) \\ \hline
WA                                                               & 61.37(1.02) / 70.56(0.51) & 61.47(0.72) / 69.67(0.63) \\ \hline
PODNet                                                           & 62.62(0.39) / 72.62(0.75) & 64.30(0.78) / 73.56(1.01) \\ \hline
DER                                                              & 67.98(1.34) / 75.88(0.78) & 70.68(0.75) / 76.56(0.95) \\ \hline
FOSTER                                                           & 66.45(0.55) / 73.93(0.77) & 69.86(0.45) / 75.27(0.83) \\ \hline
BEEF                                                             & 65.51(1.29) / 72.98(0.50) & NaN                       \\ \hline
MEMO                                                             & 64.64(1.54) / 73.50(0.83) & 51.40(3.39) / 62.11(3.33) \\ \hline
\end{tabular}
}
\end{table}

\newpage

\noindent\textbf{Class-IL without a pretrained model ($D^{HT}$ = \text{ImageNet-50-1})}

\begin{table}[h!]
\caption{The experimental results of class-IL without a pretrained model (using $D^{HT}$ = \text{ImageNet-50-1}) in the hyperparameter tuning phase. \tmlrcsm{Each result represents the Final Accuracy and Average Accuracy (Acc / AvgAcc).} The values in parentheses represent the standard deviation.}
\centering
\smallskip\noindent
\resizebox{.8\linewidth}{!}{
\begin{tabular}{|c||c|c|}
\hline
\begin{tabular}[c]{@{}c@{}}10 Tasks\\ (Acc / AvgAcc)\end{tabular} & $D^{E}$ = ImageNet-50-2      & $D^{E}$ = CIFAR-50-2   \\ \hline \hline
Replay                                                            & 43.71(0.81) / 58.75(1.60) & 44.19(2.17) / 63.57(1.50) \\ \hline
iCaRL                                                             & 39.41(1.46) / 59.51(1.70) & 41.59(3.10) / 62.42(2.85) \\ \hline
BiC                                                               & 51.26(1.39) / 65.33(2.48) & 51.22(3.67) / 66.41(2.92) \\ \hline
WA                                                                & 51.85(0.79) / 67.23(1.79) & 57.72(1.92) / 71.39(2.00) \\ \hline
PODNet                                                            & 51.31(1.24) / 67.28(1.53) & 48.19(1.17) / 65.77(1.29) \\ \hline
DER                                                               & 64.89(1.16) / 74.15(1.56) & 63.64(1.32) / 75.32(1.21) \\ \hline
FOSTER                                                            & 61.57(0.70) / 72.38(1.20) & 58.64(2.15) / 72.89(1.81) \\ \hline
BEEF                                                              & NaN                       & NaN                       \\ \hline
MEMO                                                              & 57.56(1.24) / 68.36(2.27) & 58.99(1.01) / 72.43(1.81) \\ \hline
\end{tabular}
}
\end{table}

\newpage

\begin{table}[h!]
\caption{The experimental results of class-IL without a pretrained model (using $D^{HT}$ = \text{ImageNet-50-1}) in the hyperparameter tuning phase. \tmlrcsm{Each result represents the Final Accuracy and Average Accuracy (Acc / AvgAcc).} The values in parentheses represent the standard deviation.}
\centering
\smallskip\noindent
\resizebox{.8\linewidth}{!}{
\begin{tabular}{|c||c|c|}
\hline
\begin{tabular}[c]{@{}c@{}}6 Tasks\\ (Acc / AvgAcc)\end{tabular} & $D^{E}$ = ImageNet-50-2      & $D^{E}$ = CIFAR-50-2   \\ \hline \hline
Replay                                                           & 42.82(1.43) / 53.50(1.54) & 42.28(0.71) / 52.18(1.31) \\ \hline
iCaRL                                                            & 42.47(1.73) / 54.65(1.85) & 40.24(2.64) / 52.89(2.14) \\ \hline
BiC                                                              & 44.68(2.81) / 54.19(2.93) & 39.65(1.32) / 49.49(1.46) \\ \hline
WA                                                               & 55.68(0.07) / 64.69(0.72) & 56.14(1.99) / 64.08(1.60) \\ \hline
PODNet                                                           & 64.10(0.80) / 72.50(0.81) & 61.33(0.54) / 71.27(1.07) \\ \hline
DER                                                              & 70.28(0.98) / 76.14(1.00) & 64.76(1.06) / 72.89(1.28) \\ \hline
FOSTER                                                           & 68.40(1.08) / 75.02(0.94) & 65.31(0.26) / 73.80(0.68) \\ \hline
BEEF                                                             & {NaN}              & {NaN}              \\ \hline
MEMO                                                             & 50.92(1.25) / 60.93(1.67) & 50.58(2.62) / 60.66(2.65) \\ \hline
\end{tabular}
}
\end{table}


\noindent\textbf{Class-IL with a pretrained model ($D^{HT}$ = \text{CUB-200})}

\begin{table}[h!]
\caption{The experimental results of class-IL with a pretrained model (using original hyperparameters). \tmlrcsm{Each result represents the Final Accuracy and Average Accuracy (Acc / AvgAcc).} The values in parentheses represent the standard deviation.}
\centering
\smallskip\noindent
\resizebox{.5\linewidth}{!}{
\begin{tabular}{|c||c|}
\hline
\begin{tabular}[c]{@{}c@{}}10 Tasks\\ (Acc / AvgAcc)\end{tabular} & $D^{HT}$ = CUB-200        \\ \hline \hline
L2P                                                               & 72.32(0.62) / 76.82(0.30)  \\ \hline
DualPrompt                                                        & 68.74(0.54) / 74.39(0.68) \\ \hline
CODA-Prompt                                                       & 75.19(0.33) / 80.27(0.93) \\ \hline
Adam                                                              & 71.21(1.06) / 77.52(1.24) \\ \hline
Ranpac                                                            & 78.27(0.57) / 83.24(0.44) \\ \hline
EASE                                                              & 77.07(0.19) / 82.65(0.68) \\ \hline
\end{tabular}
}
\end{table}

\begin{table}[h!]
\caption{{The experimental results of class-IL with a pretrained model (using $D^{HT}$ = \text{CUB-200}) in the hyperparameter tuning phase.} \tmlrcsm{Each result represents the Final Accuracy and Average Accuracy (Acc / AvgAcc).} The values in parentheses represent the standard deviation.}
\centering
\smallskip\noindent
\resizebox{.8\linewidth}{!}{
\begin{tabular}{|c||c|c|}
\hline
\begin{tabular}[c]{@{}c@{}}20 Tasks\\ (Acc / AvgAcc)\end{tabular} & $D^{E}$ = ImageNet-R      & $D^{E}$ = ImageNet-A      \\ \hline \hline
L2P                                                               & 69.93(0.39) / 75.90(0.23)  & 40.92(1.53) / 51.24(1.39) \\ \hline
DualPrompt                                                        & 67.20(0.78) / 73.79(0.64)  & 44.00(1.07) / 54.12(0.96)  \\ \hline
CODA-Prompt                                                       & 68.63(0.64) / 74.61(0.84) & 48.20(1.05) / 57.94(0.87)  \\ \hline
Adam                                                              & 67.70(1.38) / 74.45(1.35)  & 49.61(0.29) / 59.67(0.80)  \\ \hline
Ranpac                                                            & 78.72(0.40) / 83.71(0.56)  & 62.95(1.41) / 68.64(2.58) \\ \hline
EASE                                                              & 61.94(0.06) / 68.36(0.63) & 49.37(0.12) / 59.48(0.75) \\ \hline
\end{tabular}
}
\end{table}

\newpage

\begin{table}[h!]
\caption{{The experimental results of class-IL with a pretrained model (using $D^{HT}$ = \text{CUB-200}) in the hyperparameter tuning phase.} \tmlrcsm{Each result represents the Final Accuracy and Average Accuracy (Acc / AvgAcc).} The values in parentheses represent the standard deviation.}
\centering
\smallskip\noindent
\resizebox{.8\linewidth}{!}{
\begin{tabular}{|c||c|c|}
\hline
\begin{tabular}[c]{@{}c@{}}10 Tasks\\ (Acc / AvgAcc)\end{tabular} & $D^{E}$ = ImageNet-R      & $D^{E}$ = ImageNet-A      \\ \hline \hline
L2P                                                               & 71.86(0.66) / 77.42(0.92) & 45.13(1.25) / 53.57(0.92) \\ \hline
DualPrompt                                                        & 66.33(0.42) / 73.03(0.60)  & 39.97(2.32) / 52.58(0.70)  \\ \hline
CODA-Prompt                                                       & 72.86(0.44) / 78.49(0.99) & 51.63(0.50) / 61.00(0.47)   \\ \hline
Adam                                                              & 72.68(0.77) / 79.09(0.89) & 57.03(0.47) / 66.50(1.22)  \\ \hline
Ranpac                                                            & 79.59(0.29) / 84.46(0.41) & 66.14(0.40) / 73.63(1.05)  \\ \hline
EASE                                                              & 61.96(0.06) / 67.74(0.67) & 49.32(0.48) / 58.30(0.86)  \\ \hline
\end{tabular}
}
\end{table}

\noindent\textbf{Class-IL with a pretrained model ($D^{HT}$ = \text{ImageNet-R})}

\begin{table}[h!]
\caption{{The experimental results of class-IL with a pretrained model (using $D^{HT}$ = \text{ImageNet-R}) in the hyperparameter tuning phase.} \tmlrcsm{Each result represents the Final Accuracy and Average Accuracy (Acc / AvgAcc).} The values in parentheses represent the standard deviation.}
\centering
\smallskip\noindent
\resizebox{.8\linewidth}{!}{
\begin{tabular}{|c||c|c|}
\hline
\begin{tabular}[c]{@{}c@{}}20 Tasks\\ (Acc / AvgAcc)\end{tabular} & $D^{E}$ = CUB-200         & $D^{E}$ = ImageNet-A       \\ \hline \hline
L2P                                                               & 63.76(1.81) / 76.59(1.48) & 36.97(1.31) / 46.78(0.71)  \\ \hline
DualPrompt                                                        & 68.78(0.78) / 79.67(1.04) & 47.54(0.79) / 55.91(0.84)  \\ \hline
CODA-Prompt                                                       & 67.92(2.11) / 79.65(1.93) & 50.07(0.29) / 59.76(0.58)  \\ \hline
Adam                                                              & 85.38(0.19) / 90.87(0.90)  & 53.86(1.44) / 63.99(2.61)  \\ \hline
Ranpac                                                            & 89.86(0.22) / 93.44(0.78) & 38.53(31.11) / 67.65(3.37) \\ \hline
EASE                                                              & 79.89(1.22) / 87.58(1.19) & 53.99(1.05) / 64.11(0.78)  \\ \hline
\end{tabular}
}
\end{table}

\begin{table}[h!]
\caption{{The experimental results of class-IL with a pretrained model (using $D^{HT}$ = \text{ImageNet-R}) in the hyperparameter tuning phase.} \tmlrcsm{Each result represents the Final Accuracy and Average Accuracy (Acc / AvgAcc).} The values in parentheses represent the standard deviation.}
\centering
\smallskip\noindent
\resizebox{.8\linewidth}{!}{
\begin{tabular}{|c||c|c|}
\hline
\begin{tabular}[c]{@{}c@{}}10 Tasks\\ (Acc / AvgAcc)\end{tabular} & $D^{E}$ = CUB-200         & $D^{E}$ = ImageNet-A      \\ \hline \hline
L2P                                                               & 69.75(1.79) / 79.92(1.24) & 43.50(0.99) / 50.06(1.18)  \\ \hline
DualPrompt                                                        & 71.74(1.01) / 82.22(1.10)  & 39.47(0.79) / 50.63(0.94) \\ \hline
CODA-Prompt                                                       & 72.30(1.11) / 83.00(1.35)   & 52.39(0.38) / 61.87(1.01) \\ \hline
Adam                                                              & 85.90(0.17) / 90.93(0.89)  & 56.63(0.78) / 65.94(1.45) \\ \hline
Ranpac                                                            & 89.99(0.29) / 93.36(0.83) & 63.78(1.52) / 71.70(1.88)  \\ \hline
EASE                                                              & 74.00(0.78) / 83.69(0.74)  & 54.76(1.36) / 66.14(1.65) \\ \hline
\end{tabular}
}
\end{table}

\newpage

\noindent\textbf{Class-IL with a pretrained model ($D^{HT}$ = \text{CUB-100-1})}

\begin{table}[h!]
\caption{{The experimental results of class-IL with a pretrained model (using $D^{HT}$ = \text{CUB-100-1}) in the hyperparameter tuning phase.} \tmlrcsm{Each result represents the Final Accuracy and Average Accuracy (Acc / AvgAcc).} The values in parentheses represent the standard deviation.}
\centering
\smallskip\noindent
\resizebox{.95\linewidth}{!}{
\begin{tabular}{|c||c|c|c|}
\hline
\begin{tabular}[c]{@{}c@{}}20 Tasks\\ (Acc / AvgAcc)\end{tabular} & $D^{E}$ = CUB-100-2       & $D^{E}$ = ImageNet-R-2    & ImageNet-A-2               \\ \hline \hline
L2P                                                               & 54.12(3.59) / 68.33(3.73) & 66.01(0.74) / 72.17(1.04) & 28.08(2.38) / 39.18(2.75)  \\ \hline
DualPrompt                                                        & 59.83(1.63) / 73.54(2.68) & 65.51(0.32) / 71.58(0.68) & 33.90(2.26) / 44.84(2.25)  \\ \hline
CODA-Prompt                                                       & 58.16(1.88) / 71.05(2.68) & 66.73(0.61) / 73.06(0.46) & 30.62(0.82) / 41.70(1.70)  \\ \hline
Adam                                                              & 85.95(0.08) / 90.56(0.24) & 67.77(0.84) / 74.53(1.74) & 43.93(0.09) / 55.63(2.69)  \\ \hline
Ranpac                                                            & 89.52(0.35) / 90.52(2.96) & 74.53(0.28) / 79.80(0.81) & 30.30(22.41) / 45.87(4.57) \\ \hline
EASE                                                              & 85.19(0.49) / 89.91(0.74) & 67.17(0.29) / 73.61(0.75) & 44.11(0.29) / 55.42(2.83)  \\ \hline
\end{tabular}
}
\end{table}

\begin{table}[h!]
\caption{{The experimental results of class-IL with a pretrained model (using $D^{HT}$ = \text{CUB-100-1}) in the hyperparameter tuning phase.} \tmlrcsm{Each result represents the Final Accuracy and Average Accuracy (Acc / AvgAcc).} The values in parentheses represent the standard deviation.}
\centering
\smallskip\noindent
\resizebox{.95\linewidth}{!}{
\begin{tabular}{|c||c|c|c|}
\hline
\begin{tabular}[c]{@{}c@{}}10 Tasks\\ (Acc / AvgAcc)\end{tabular} & $D^{E}$ = CUB-100-2       & $D^{E}$ = ImageNet-R-2    & ImageNet-A-2              \\ \hline \hline
L2P                                                               & 66.15(1.41) / 76.68(1.49) & 70.11(0.53) / 75.61(0.87) & 34.96(0.92) / 44.98(2.26) \\ \hline
DualPrompt                                                        & 67.20(2.59) / 78.28(1.68) & 68.29(0.49) / 74.32(0.89) & 38.43(1.52) / 49.15(2.43) \\ \hline
CODA-Prompt                                                       & 68.37(2.71) / 78.93(2.57) & 70.35(0.81) / 75.59(0.90) & 37.23(1.87) / 47.48(1.85) \\ \hline
Adam                                                              & 86.76(0.21) / 90.75(0.46) & 72.73(0.27) / 79.42(0.59) & 44.81(0.85) / 55.08(2.22) \\ \hline
Ranpac                                                            & 90.60(0.36) / 93.08(0.65)  & 80.40(0.3) / 85.00(0.47)    & 49.56(2.52) / 57.60(1.96)  \\ \hline
EASE                                                              & 85.86(0.10) / 90.11(0.26)  & 63.36(0.03) / 69.36(0.95) & 43.88(0.15) / 54.49(2.64) \\ \hline
\end{tabular}
}
\end{table}

\newpage

\noindent\textbf{Class-IL with a pretrained model ($D^{HT}$ = \text{ImageNet-R-1})}

\begin{table}[h!]
\caption{{The experimental results of class-IL with a pretrained model (using $D^{HT}$ = \text{ImageNet-R-1}) in the hyperparameter tuning phase.} \tmlrcsm{Each result represents the Final Accuracy and Average Accuracy (Acc / AvgAcc).} The values in parentheses represent the standard deviation.}
\centering
\smallskip\noindent
\resizebox{.95\linewidth}{!}{
\begin{tabular}{|c||c|c|c|}
\hline
\begin{tabular}[c]{@{}c@{}}20 Tasks\\ (Acc / AvgAcc)\end{tabular} & $D^{E}$ = ImageNet-R-2    & $D^{E}$ = CUB-100-2        & ImageNet-A-2               \\ \hline \hline
L2P                                                               & 66.15(0.85) / 71.93(1.13) & 51.04(1.45) / 66.04(1.71)  & 25.13(2.27) / 34.21(2.51)  \\ \hline
DualPrompt                                                        & 65.77(0.78) / 71.83(1.17) & 57.13(3.40) / 71.15(2.25)  & 31.96(2.49) / 41.71(1.76)  \\ \hline
CODA-Prompt                                                       & 66.44(0.66) / 72.62(0.36) & 57.24(1.90) / 71.27(1.95)  & 30.48(1.62) / 41.30(2.56)   \\ \hline
Adam                                                              & 70.69(0.73) / 77.86(0.51) & 86.35(0.14) / 90.83(0.56)  & 44.25(0.86) / 55.84(2.75)  \\ \hline
Ranpac                                                            & 76.15(0.93) / 81.68(0.94) & 73.73(31.52) / 89.58(2.03) & 35.06(15.86) / 47.04(6.07) \\ \hline
EASE                                                              & 75.16(0.68) / 81.68(0.71) & 76.36(2.61) / 84.35(2.55)  & 42.49(1.76) / 54.40(3.21)   \\ \hline
\end{tabular}
}
\end{table}

\begin{table}[h!]
\caption{{The experimental results of class-IL with a pretrained model (using $D^{HT}$ = \text{ImageNet-R-1}) in the hyperparameter tuning phase.} \tmlrcsm{Each result represents the Final Accuracy and Average Accuracy (Acc / AvgAcc).} The values in parentheses represent the standard deviation.}
\centering
\smallskip\noindent
\resizebox{.95\linewidth}{!}{
\begin{tabular}{|c||c|c|c|}
\hline
\begin{tabular}[c]{@{}c@{}}10 Tasks\\ (Acc / AvgAcc)\end{tabular} & $D^{E}$ = ImageNet-R-2    & $D^{E}$ = CUB-100-2       & ImageNet-A-2              \\ \hline \hline
L2P                                                               & 70.35(0.64) / 75.66(0.30) & 63.71(2.33) / 74.62(1.61) & 29.10(1.24) / 38.80(1.44)  \\ \hline
DualPrompt                                                        & 69.97(0.25) / 75.93(0.62) & 66.66(1.12) / 78.11(1.43) & 32.42(0.68) / 42.31(2.02) \\ \hline
CODA-Prompt                                                       & 72.17(0.46) / 77.80(0.50) & 66.98(1.3) / 78.70(0.98)  & 37.04(1.49) / 46.47(2.45) \\ \hline
Adam                                                              & 72.84(0.67) / 79.69(0.86) & 85.26(0.41) / 89.77(0.45) & 37.36(2.72) / 48.62(4.07) \\ \hline
Ranpac                                                            & 80.70(0.50) / 85.28(0.46)   & 91.09(0.51) / 91.63(3.51) & 41.98(19.61) / 58.79(4.70) \\ \hline
EASE                                                              & 78.33(0.41) / 83.82(0.71) & 79.70(1.47) / 86.23(1.59)  & 42.49(0.69) / 53.69(2.61) \\ \hline
\end{tabular}
}
\end{table}

\newpage

\subsection{Training graphs}

\noindent\textbf{Class-IL without a pretrained model ($D^{HT} = \text{CIFAR50-1}, D^{E} = \text{CUB50-2}$)}

\begin{figure*}[!h]
\vspace{-.12in}
\centering 
\subfigure[10 tasks]
{\includegraphics[width=0.40\linewidth]{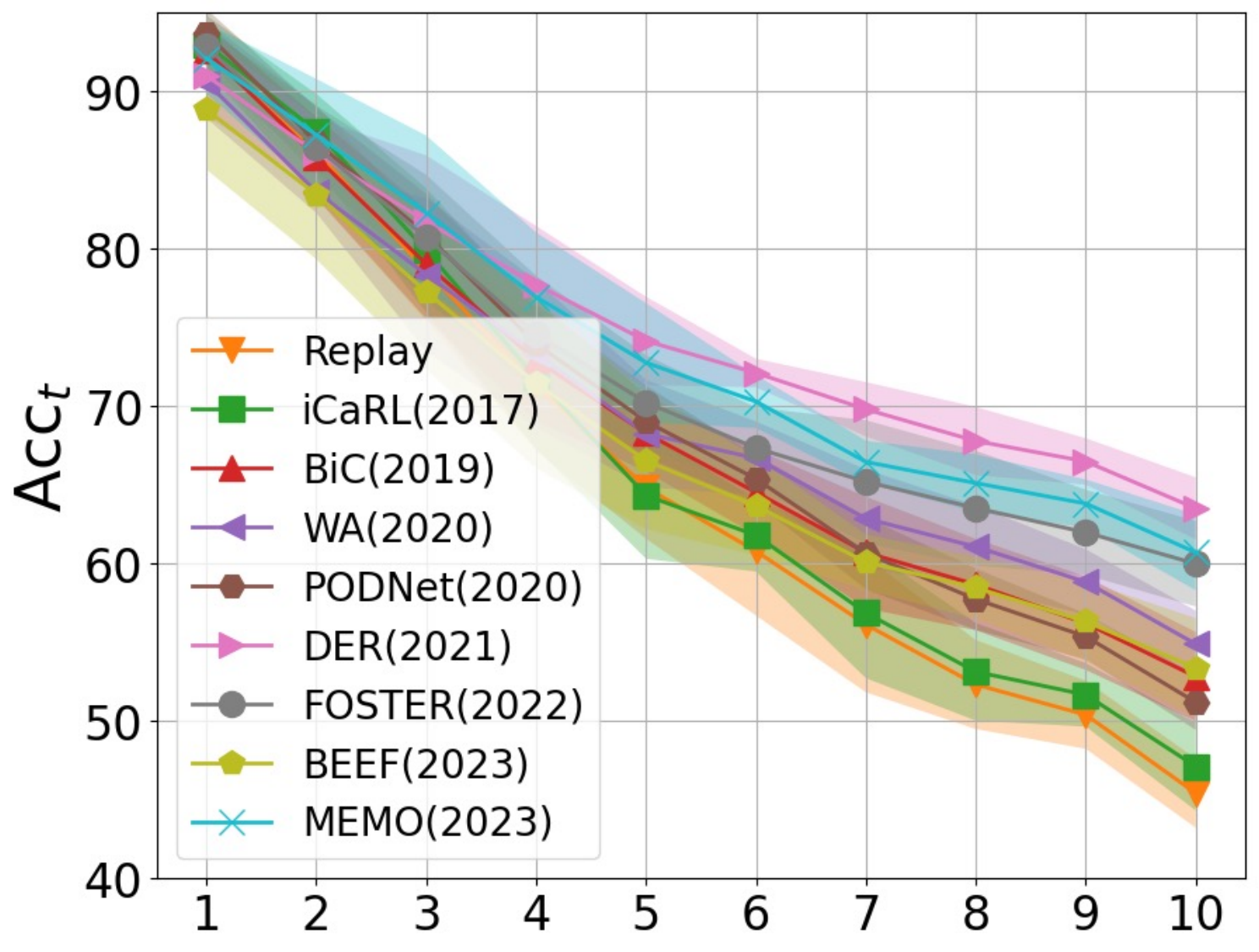}}
\subfigure[6 tasks]
{\includegraphics[width=0.40\linewidth]{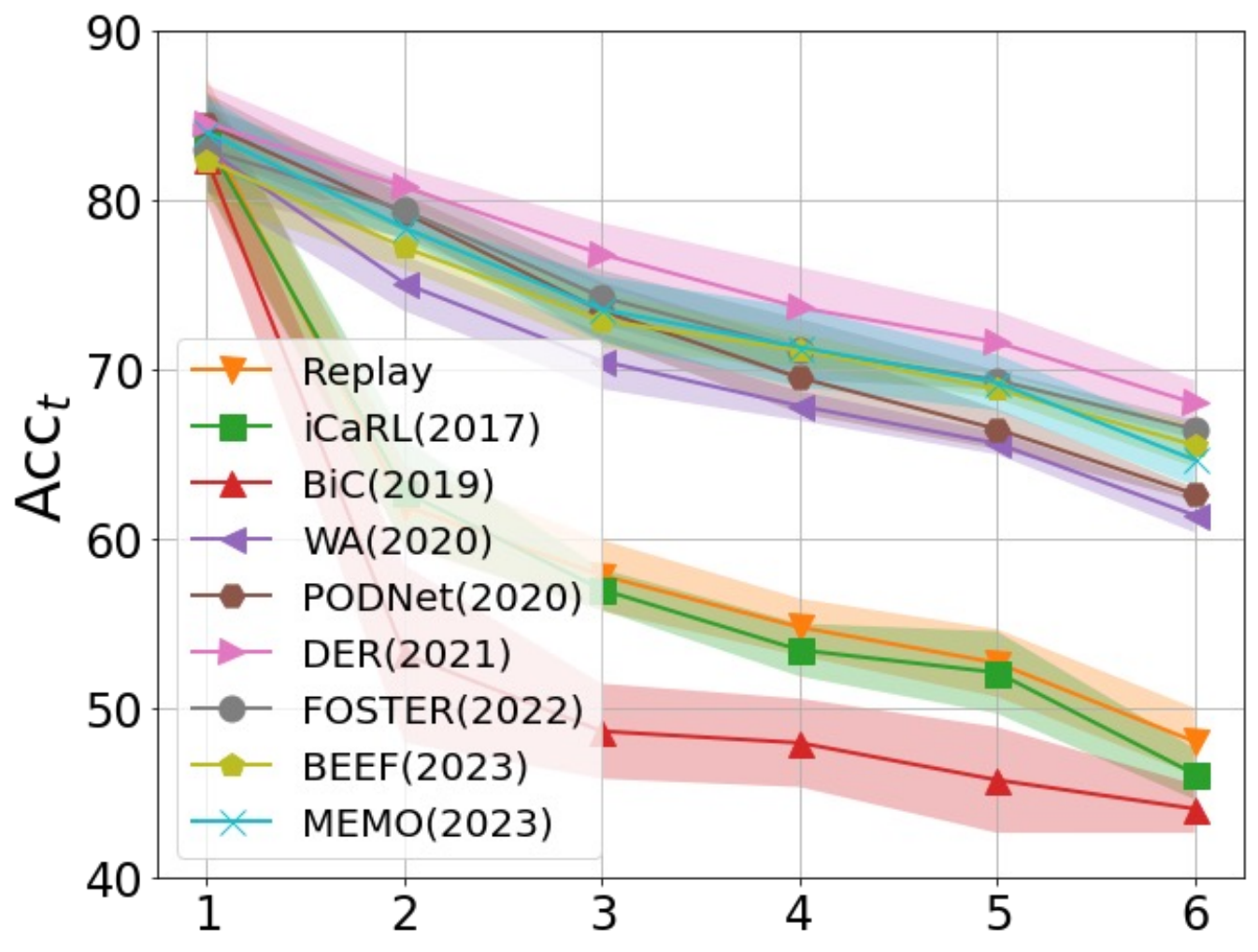}}
\vspace{-.1in}
\caption{Experimental results on the evaluation phase.}
\vspace{-.12in}
 \end{figure*}

\noindent\textbf{Class-IL without a pretrained model ($D^{HT} = \text{CIFAR50-1}, D^{E} = \text{ImageNet50-2}$)}

\begin{figure*}[!h]
\vspace{-.12in}
\centering 
\subfigure[10 tasks]
{\includegraphics[width=0.40\linewidth]{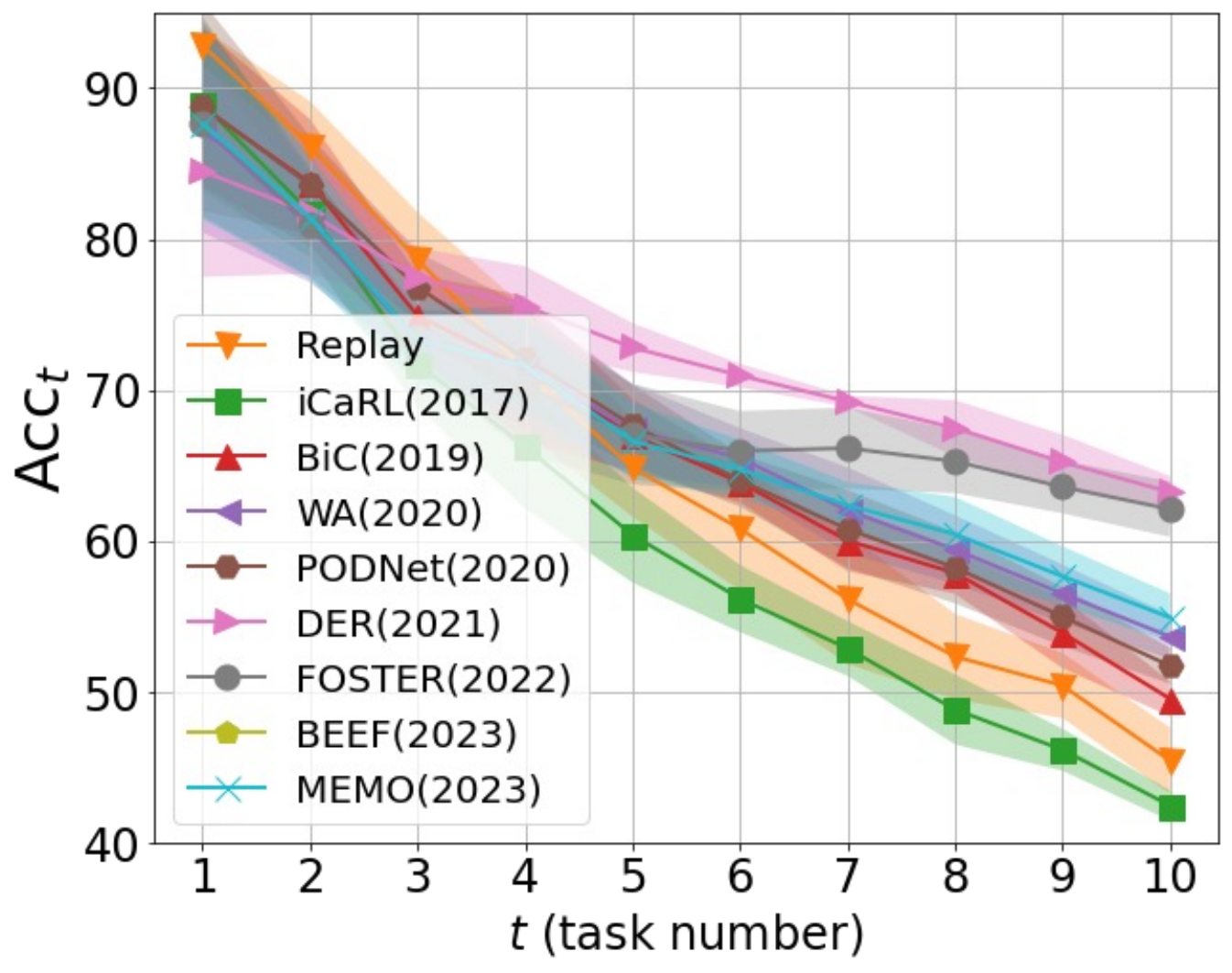}}
\subfigure[6 tasks]
{\includegraphics[width=0.40\linewidth]{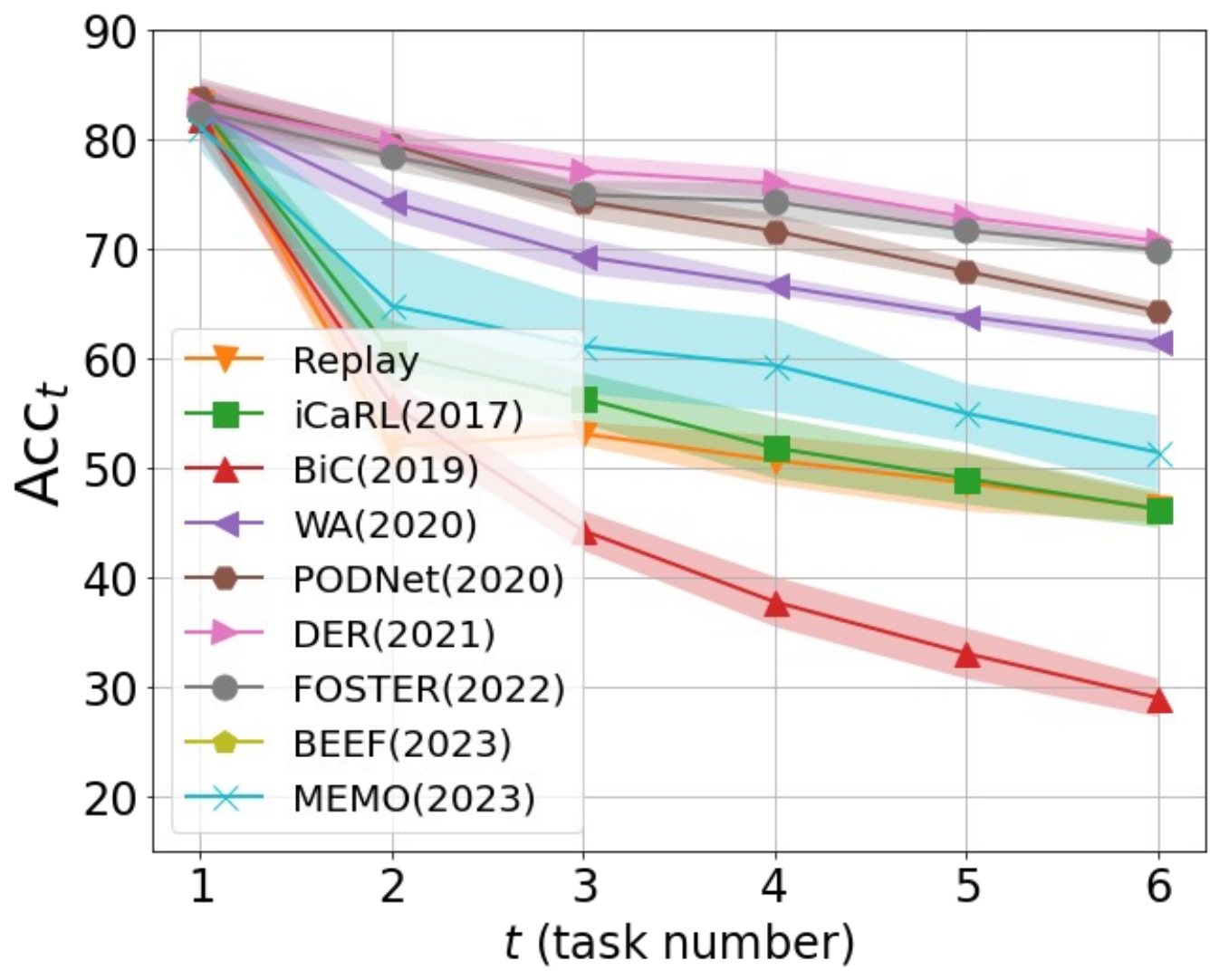}}
\vspace{-.1in}
\caption{Experimental results on the evaluation phase.}
\vspace{-.12in}
 \end{figure*}

\noindent\textbf{Class-IL without a pretrained model ($D^{HT} = \text{ImageNet50-1}, D^{E} = \text{ImageNet50-2}$)}

\begin{figure*}[!h]
\vspace{-.12in}
\centering 
\subfigure[10 tasks]
{\includegraphics[width=0.40\linewidth]{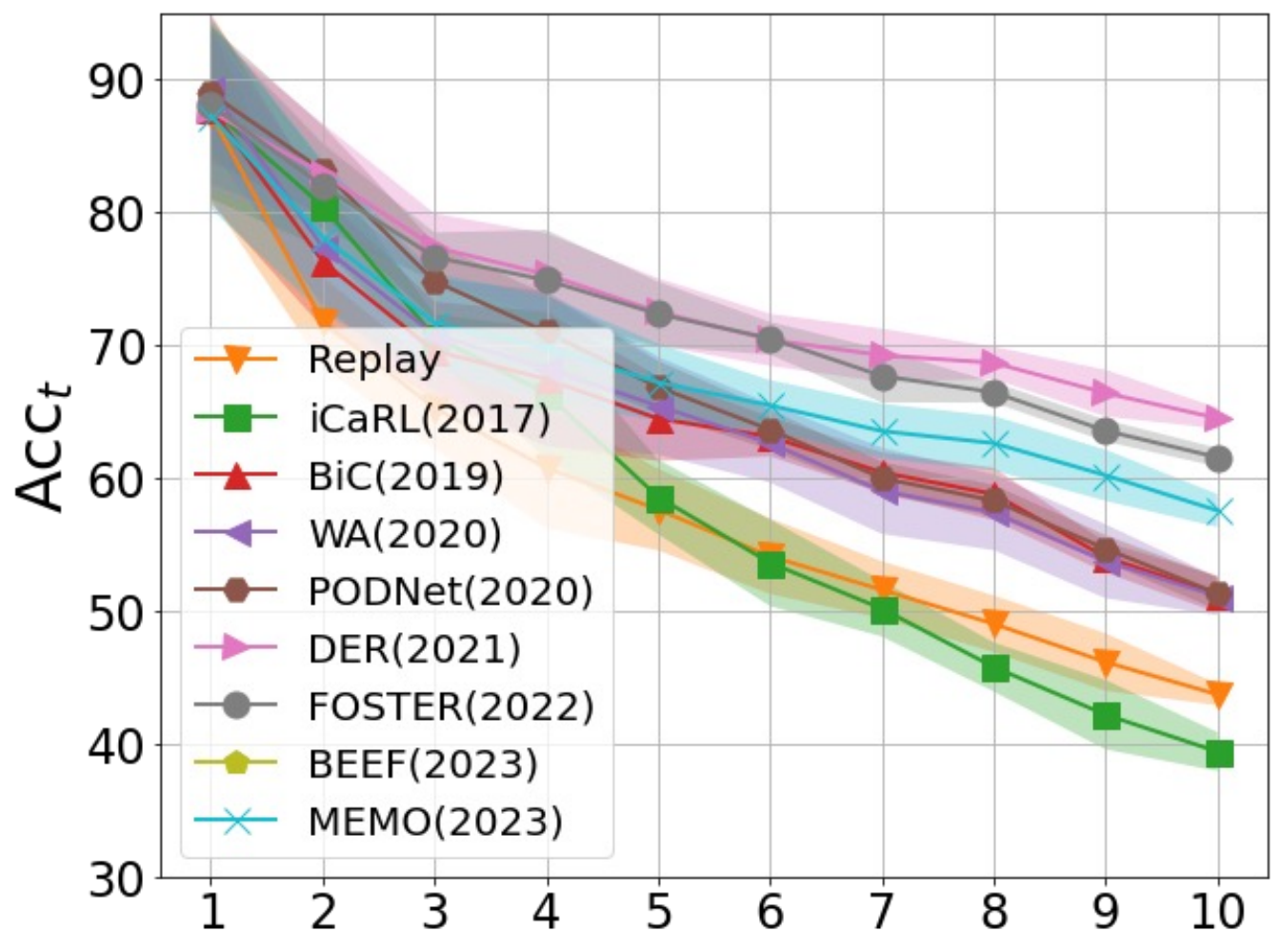}}
\subfigure[6 tasks]
{\includegraphics[width=0.40\linewidth]{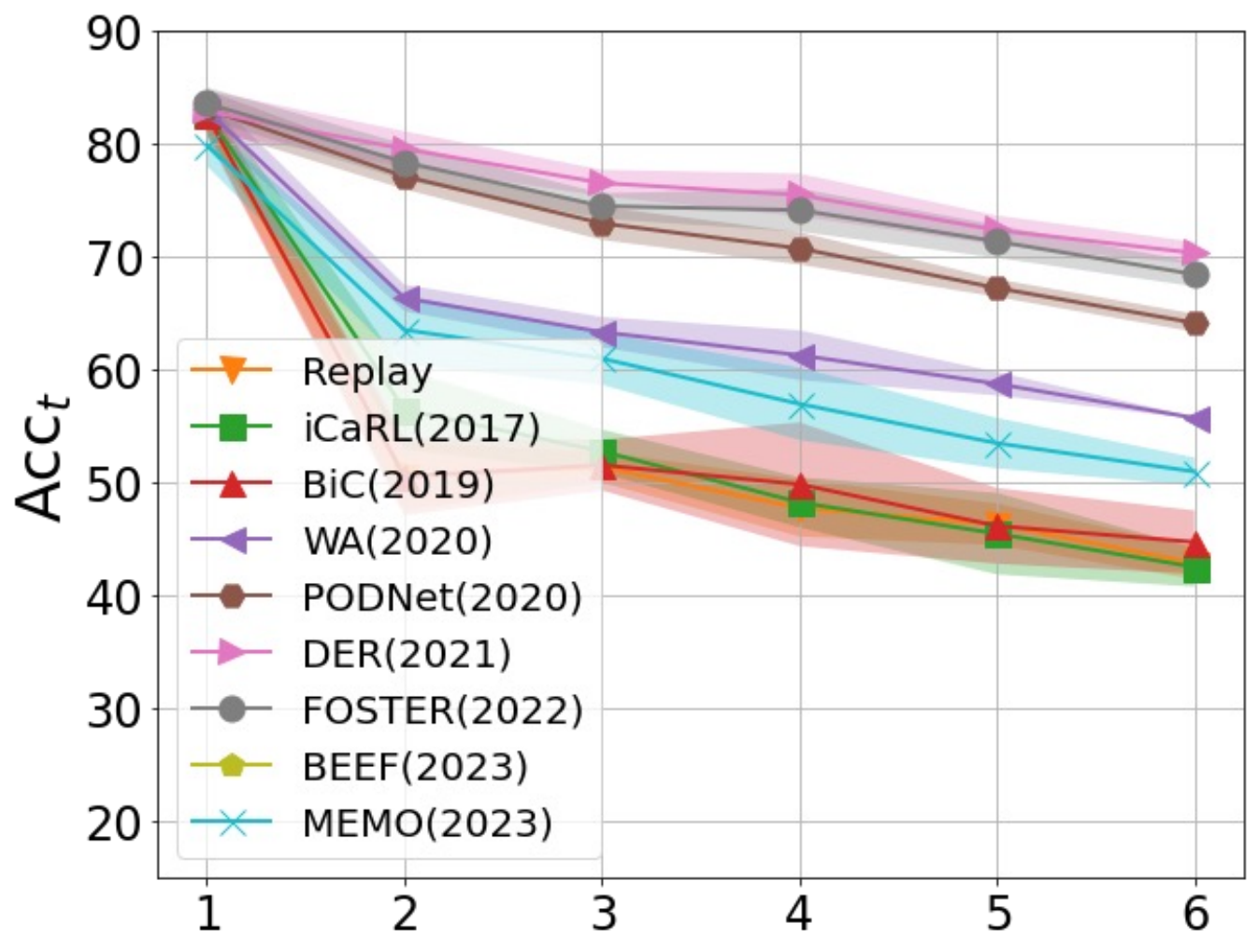}}
\vspace{-.1in}
\caption{Experimental results on the evaluation phase.}
\vspace{-.12in}
 \end{figure*}

 \newpage

\noindent\textbf{Class-IL without a pretrained model ($D^{HT} = \text{ImageNet50-1}, D^{E} = \text{CIFAR50-2}$)}

\begin{figure*}[!h]
\vspace{-.12in}
\centering 
\subfigure[10 tasks]
{\includegraphics[width=0.40\linewidth]{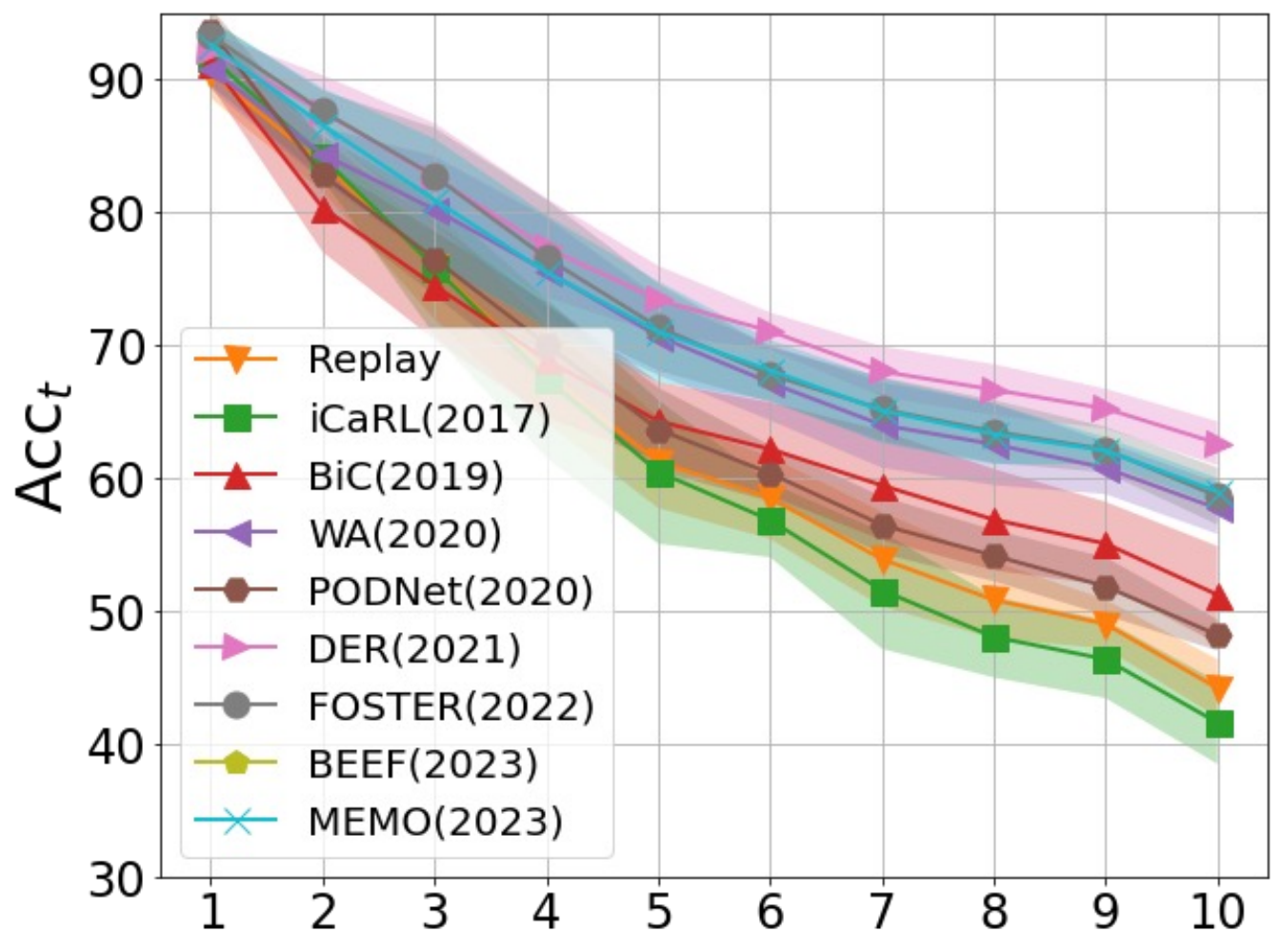}}
\subfigure[6 tasks]
{\includegraphics[width=0.40\linewidth]{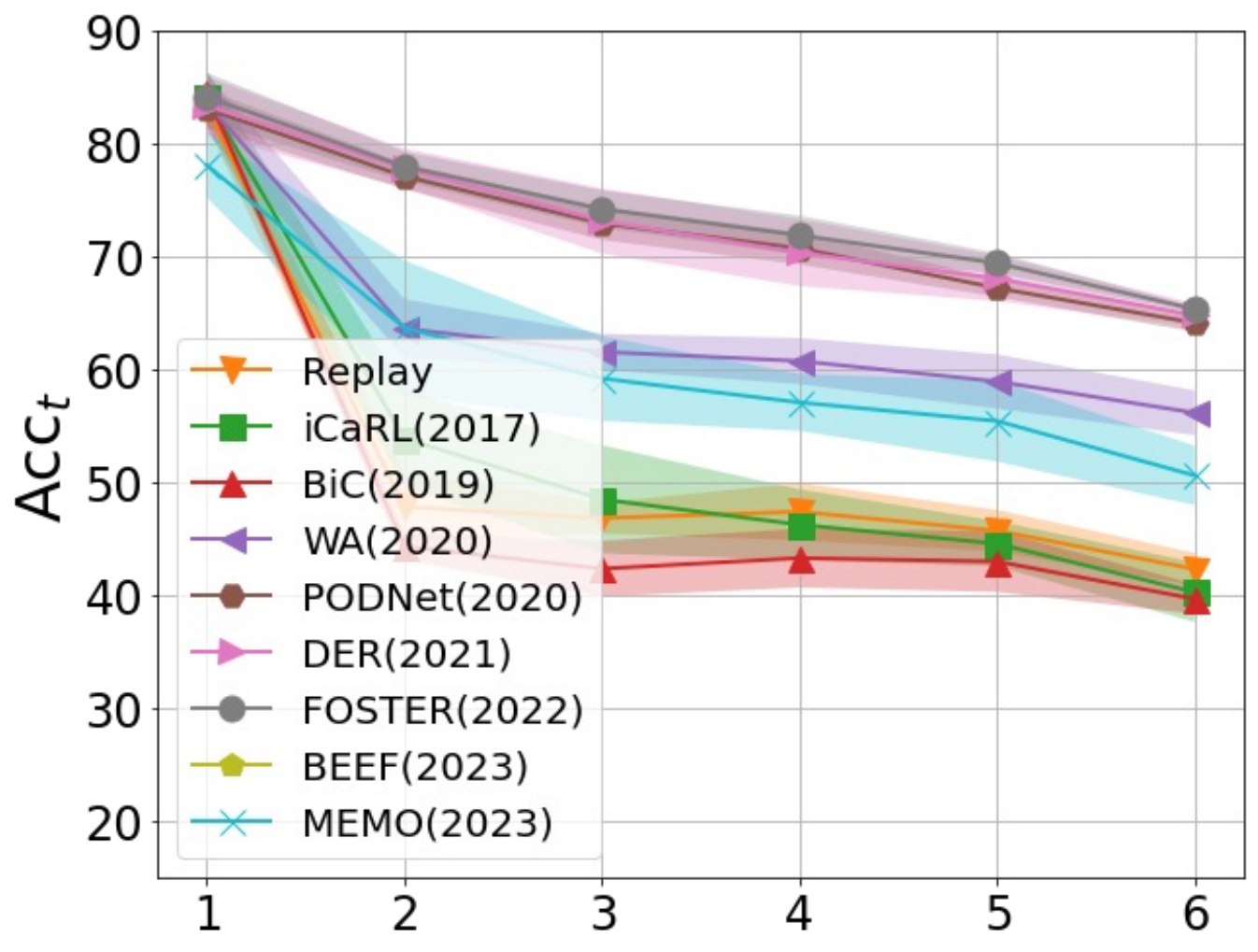}}
\vspace{-.1in}
\caption{Experimental results on the evaluation phase.}
\vspace{-.12in}
 \end{figure*}

\noindent\textbf{Class-IL with a pretrained model ($D^{HT} = \text{CUB100-1}, D^{E} = \text{CUB100-2}$)}

\begin{figure*}[!h]
\vspace{-.12in}
\centering 
\subfigure[20 tasks]
{\includegraphics[width=0.40\linewidth]{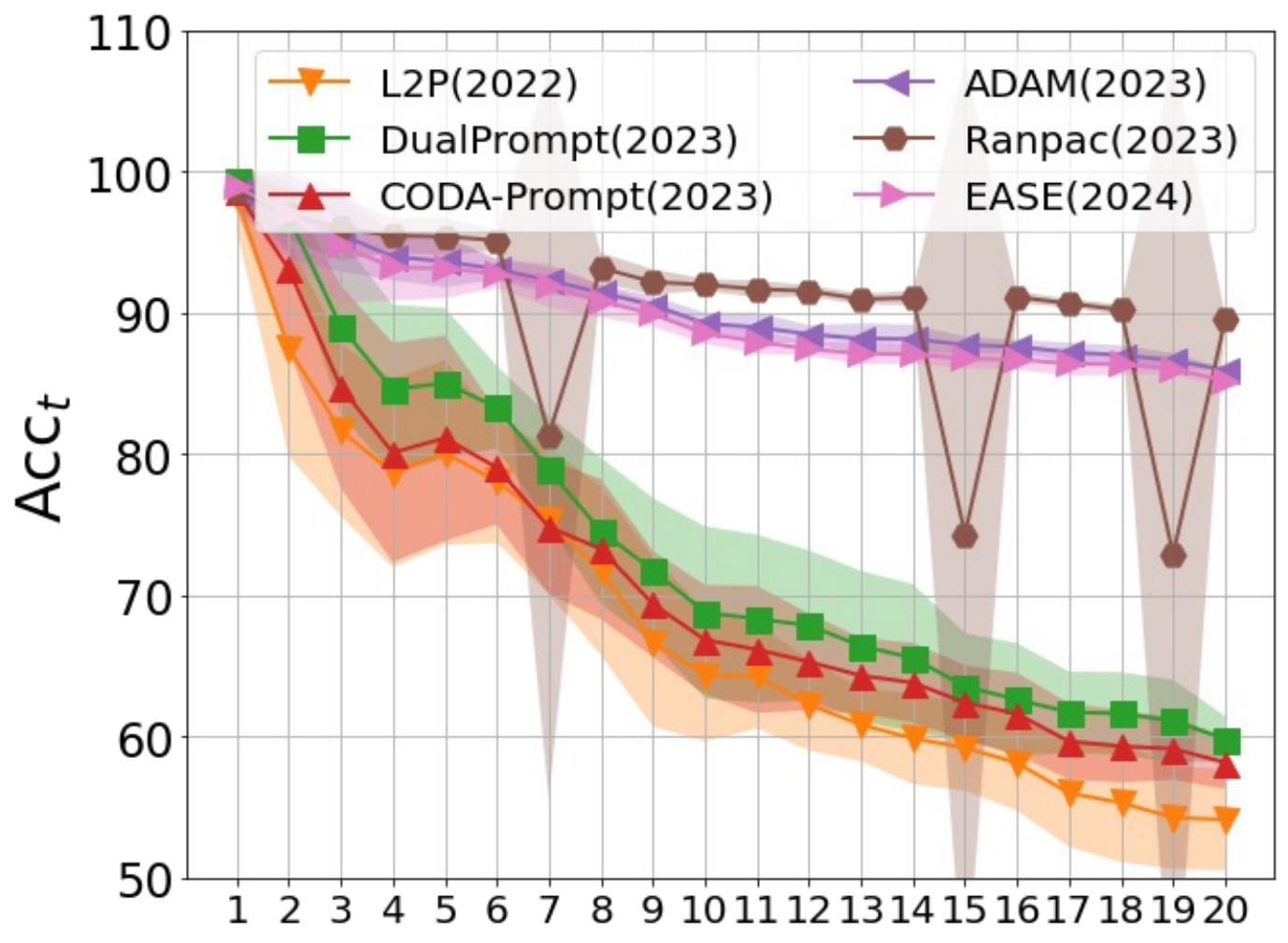}}
\subfigure[10 tasks]
{\includegraphics[width=0.40\linewidth]{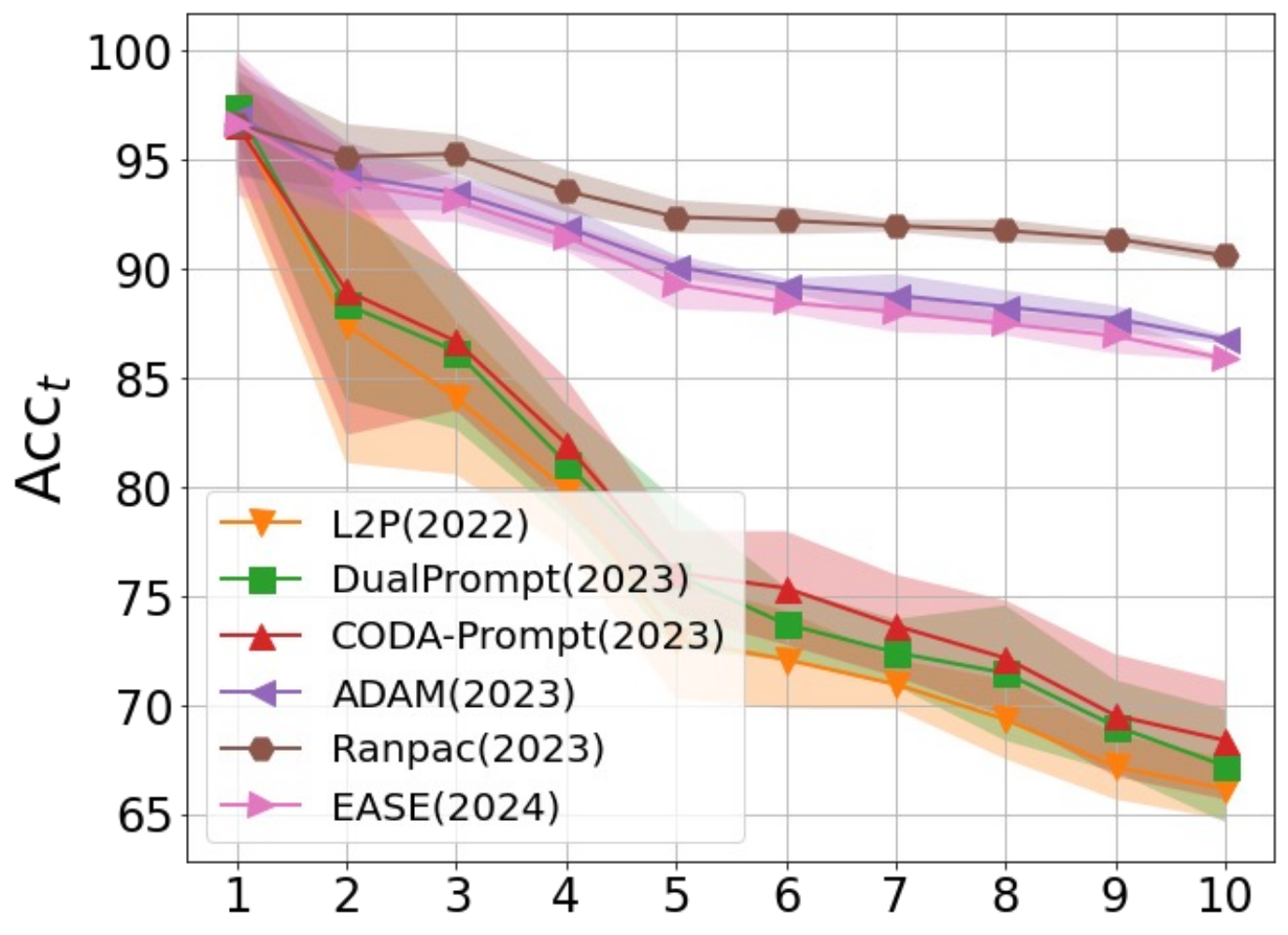}}
\vspace{-.1in}
\caption{Experimental results on the evaluation phase.}
\vspace{-.12in}
 \end{figure*}

\noindent\textbf{Class-IL with a pretrained model ($D^{HT} = \text{CUB100-1}, D^{E} = \text{ImageNet-R-2}$)}

\begin{figure*}[!h]
\vspace{-.12in}
\centering 
\subfigure[20 tasks]
{\includegraphics[width=0.40\linewidth]{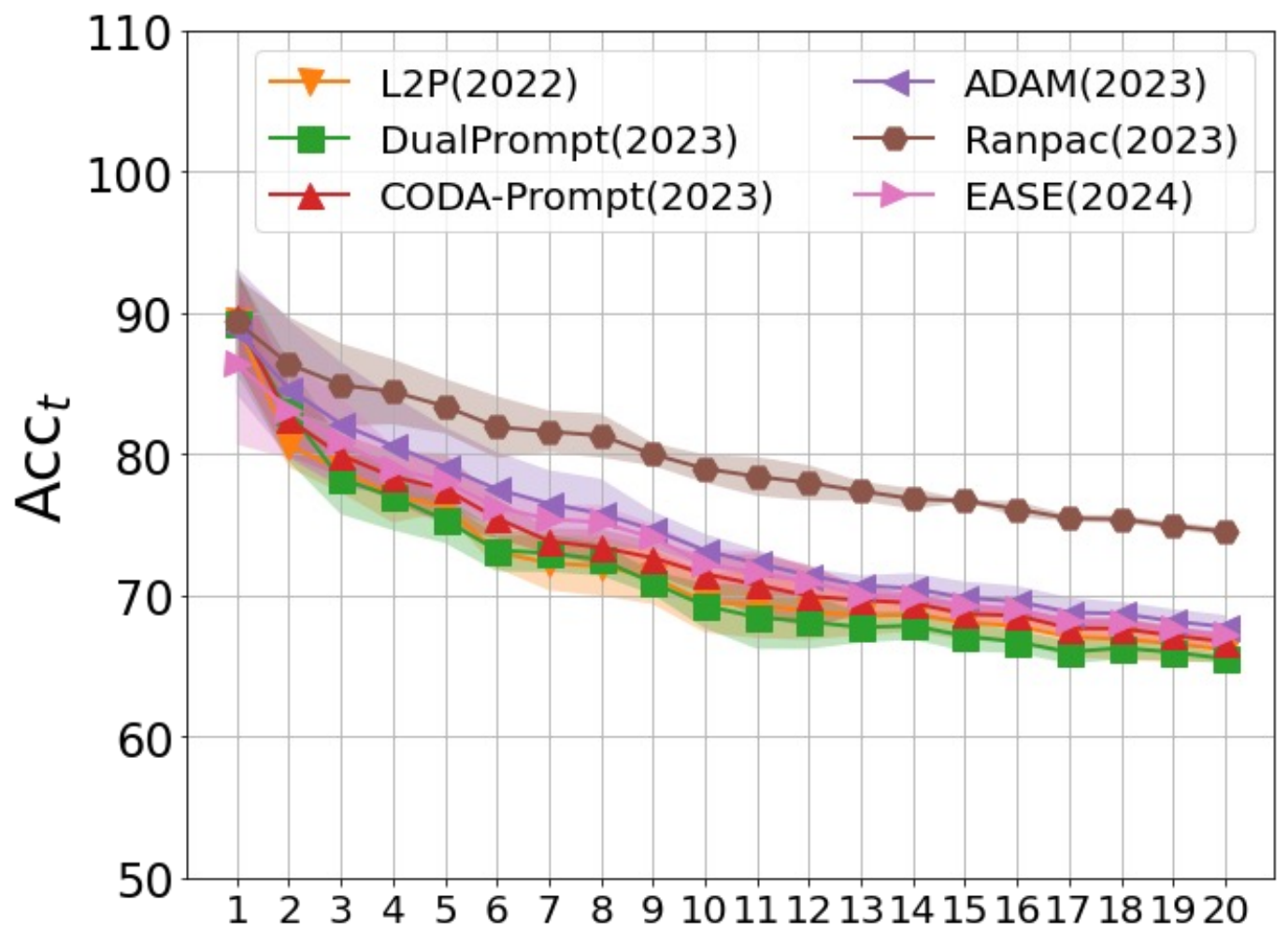}}
\subfigure[10 tasks]
{\includegraphics[width=0.40\linewidth]{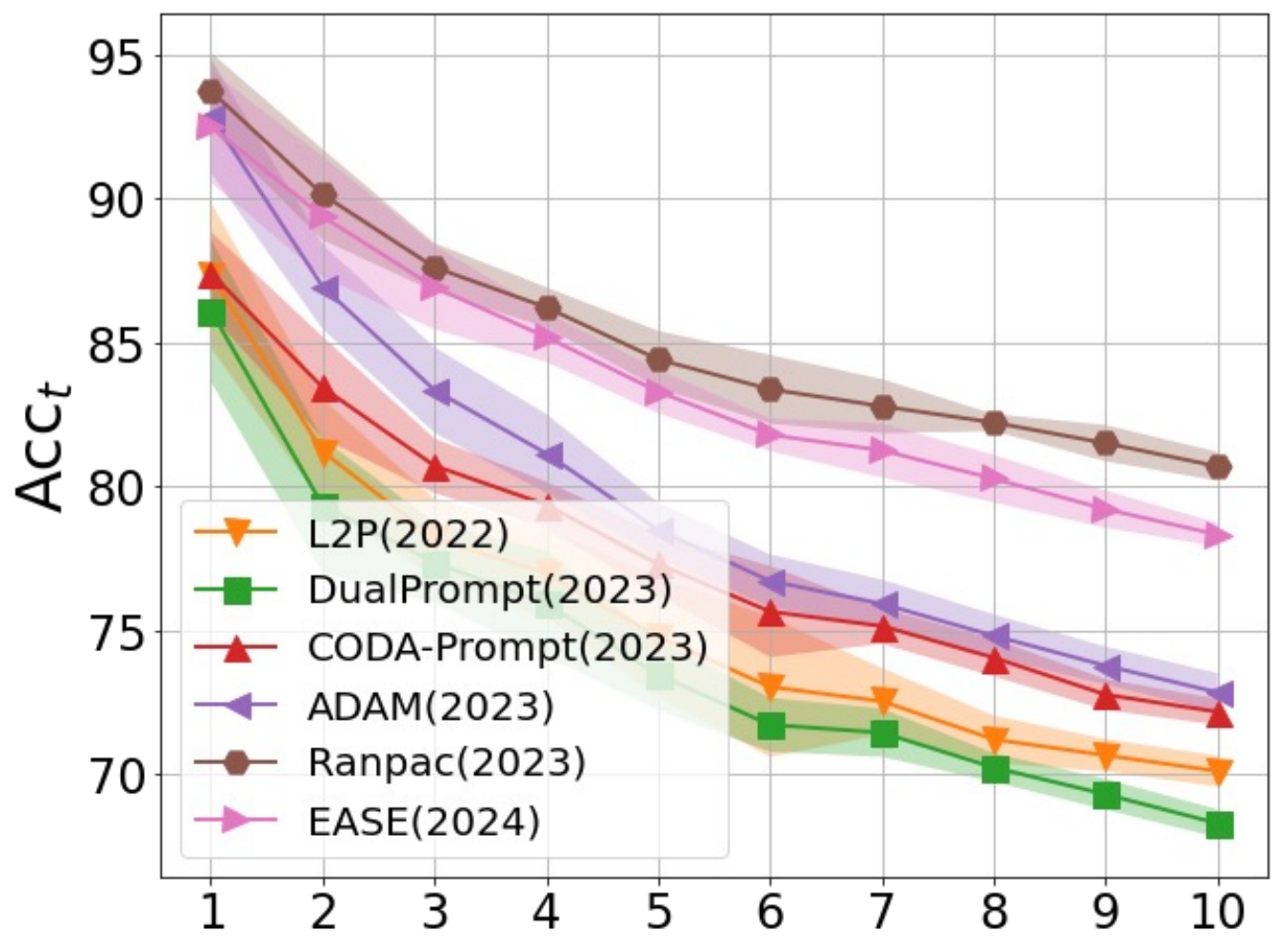}}
\vspace{-.1in}
\caption{Experimental results on the evaluation phase.}
\vspace{-.12in}
 \end{figure*}

 \newpage

\noindent\textbf{Class-IL with a pretrained model ($D^{HT} = \text{ImageNet-R-1}, D^{E} = \text{ImageNet-R-2}$)}

\begin{figure*}[!h]
\vspace{-.12in}
\centering 
\subfigure[20 tasks]
{\includegraphics[width=0.40\linewidth]{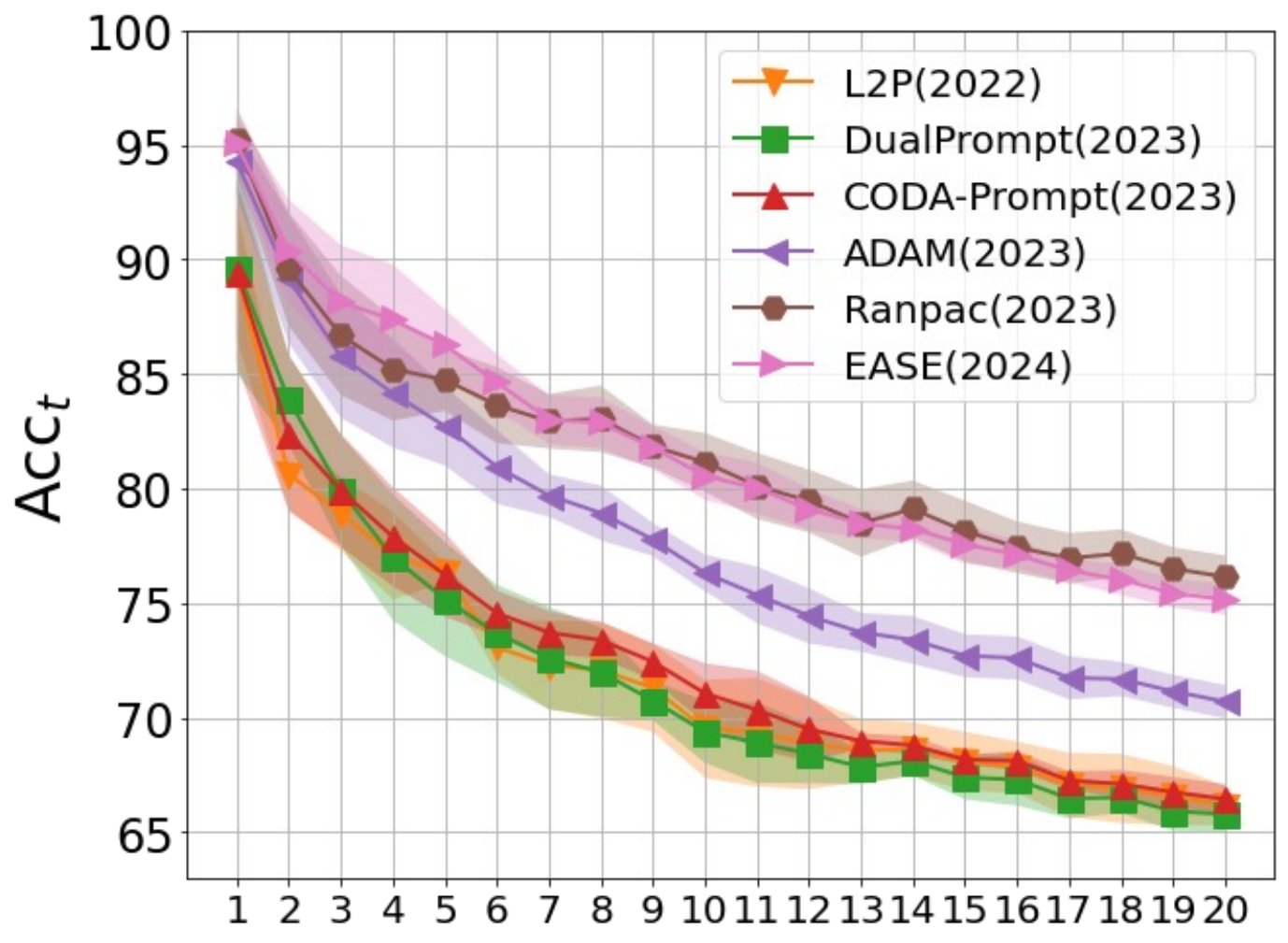}}
\subfigure[10 tasks]
{\includegraphics[width=0.40\linewidth]{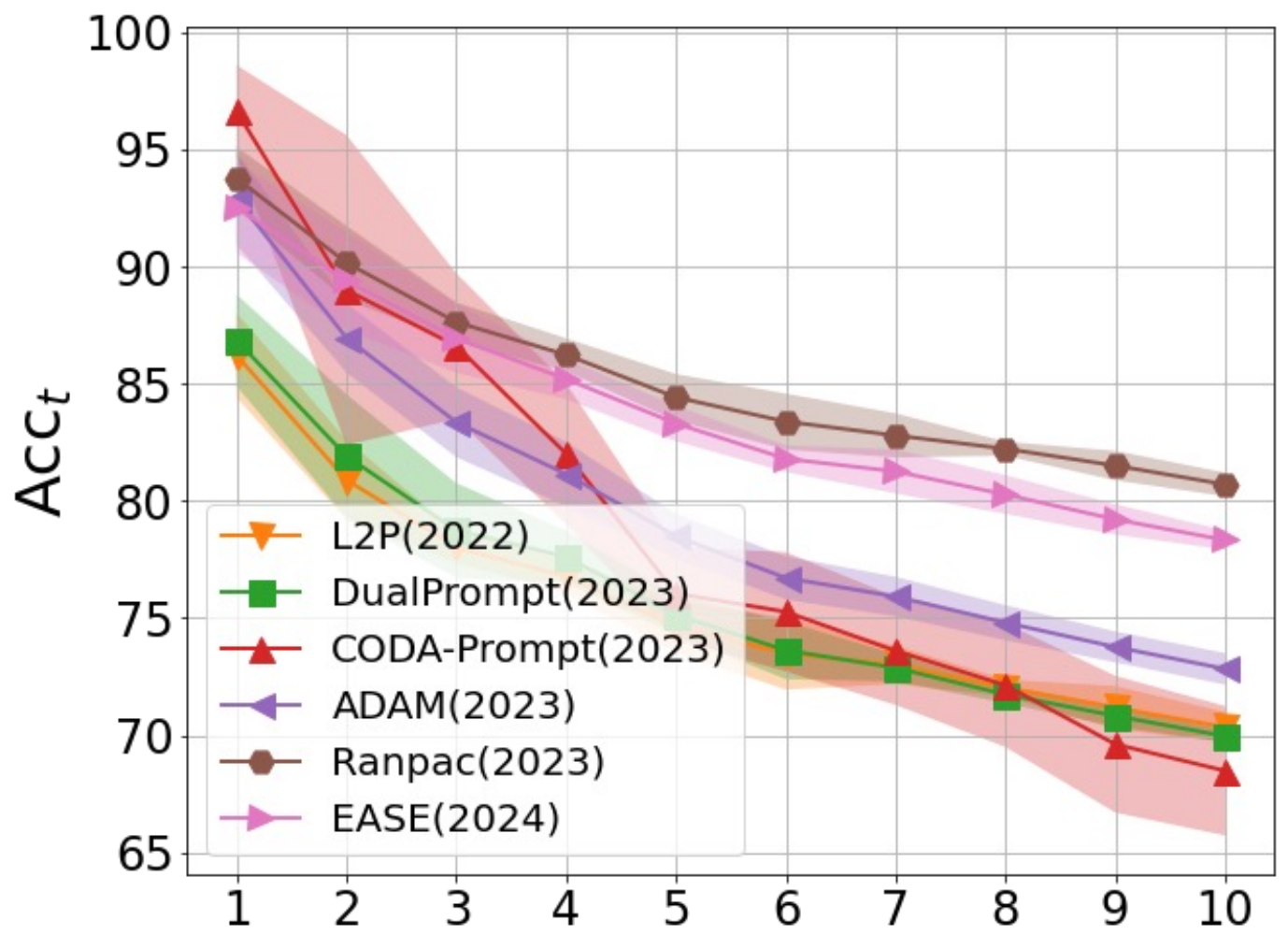}}
\vspace{-.1in}
\caption{Experimental results on the evaluation phase.}
\vspace{-.12in}
 \end{figure*}

\noindent\textbf{Class-IL with a pretrained model ($D^{HT} = \text{ImageNet-R-1}, D^{E} = \text{CUB100-2}$)}

\begin{figure*}[!h]
\vspace{-.12in}
\centering 
\subfigure[20 tasks]
{\includegraphics[width=0.40\linewidth]{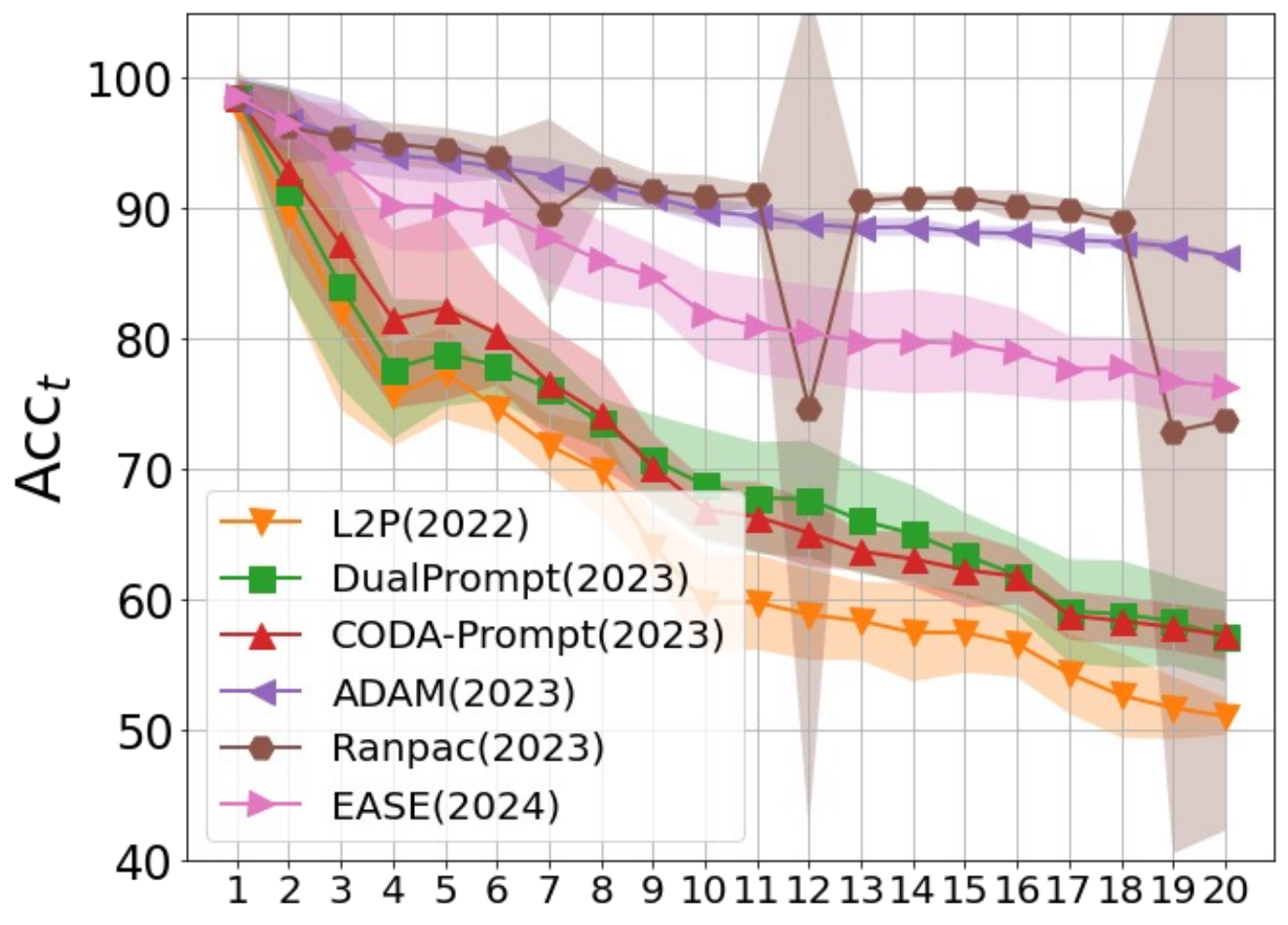}}
\subfigure[10 tasks]
{\includegraphics[width=0.40\linewidth]{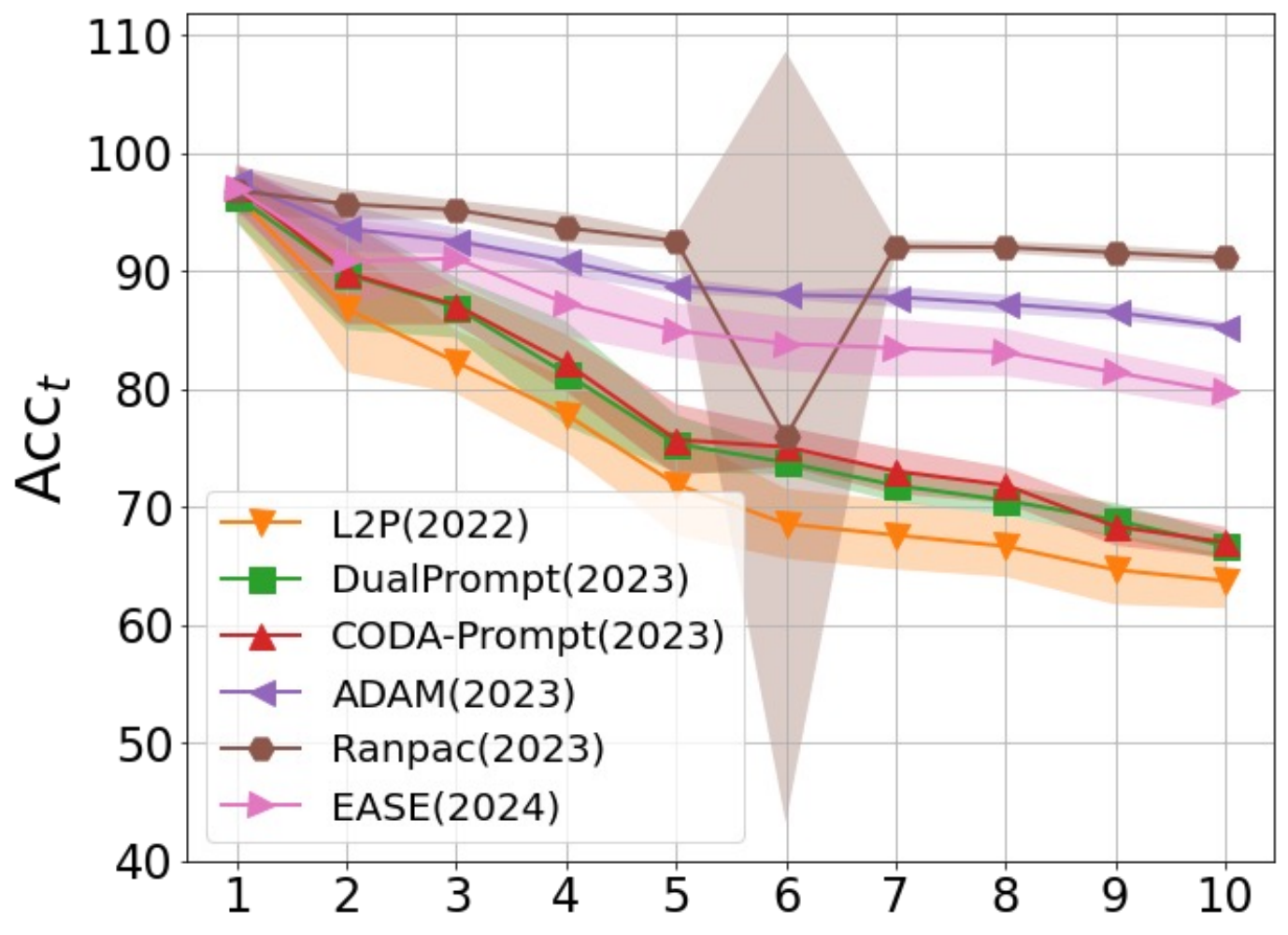}}
\vspace{-.1in}
\caption{Experimental results on the evaluation phase.}
\vspace{-.12in}
 \end{figure*}

\noindent\textbf{Class-IL with a pretrained model ($D^{HT} = \text{ImageNet-R-1}, D^{E} = \text{ImageNet-A-2}$)}

\begin{figure*}[!h]
\vspace{-.12in}
\centering 
\subfigure[20 tasks]
{\includegraphics[width=0.40\linewidth]{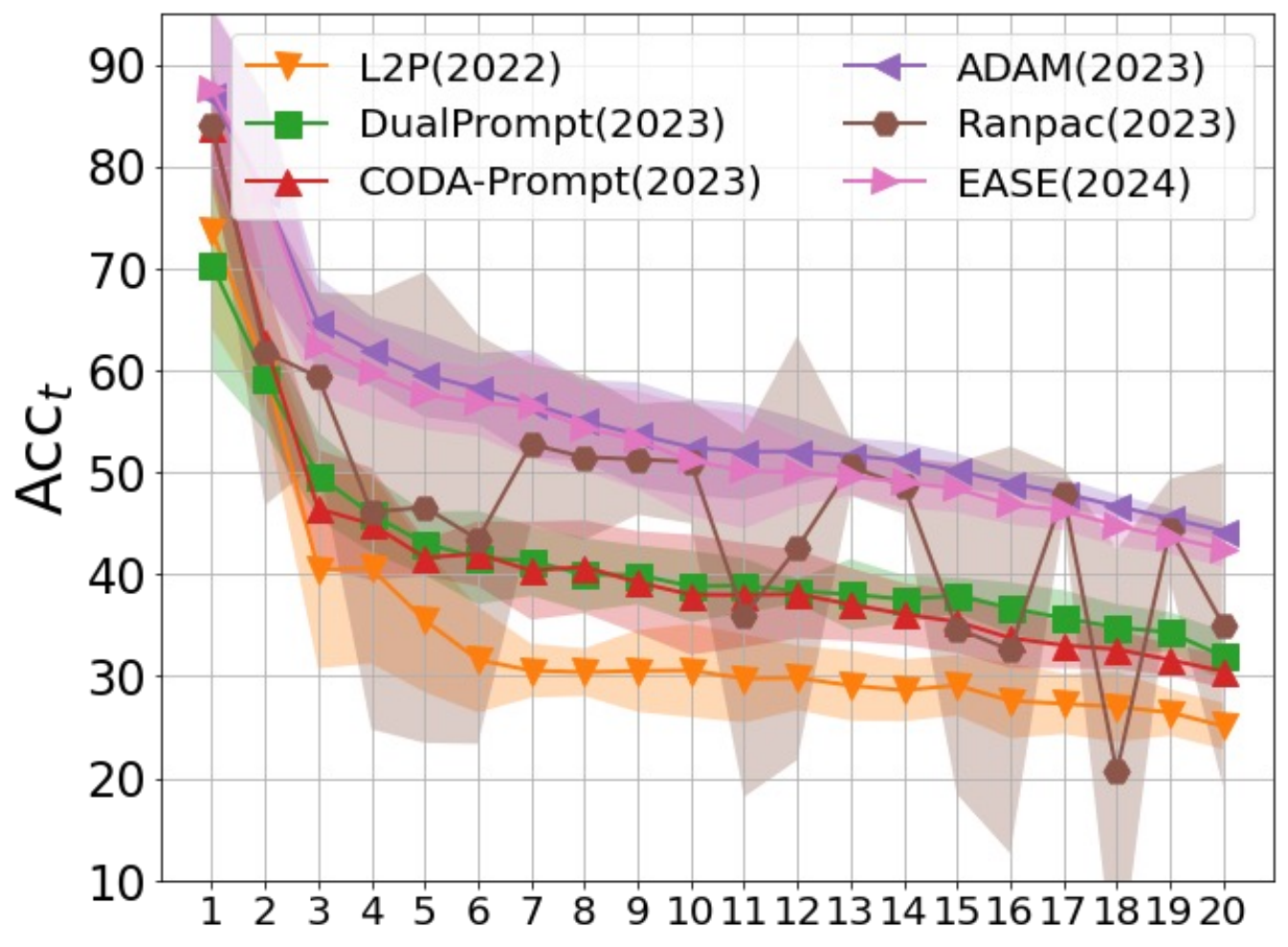}}
\subfigure[10 tasks]
{\includegraphics[width=0.40\linewidth]{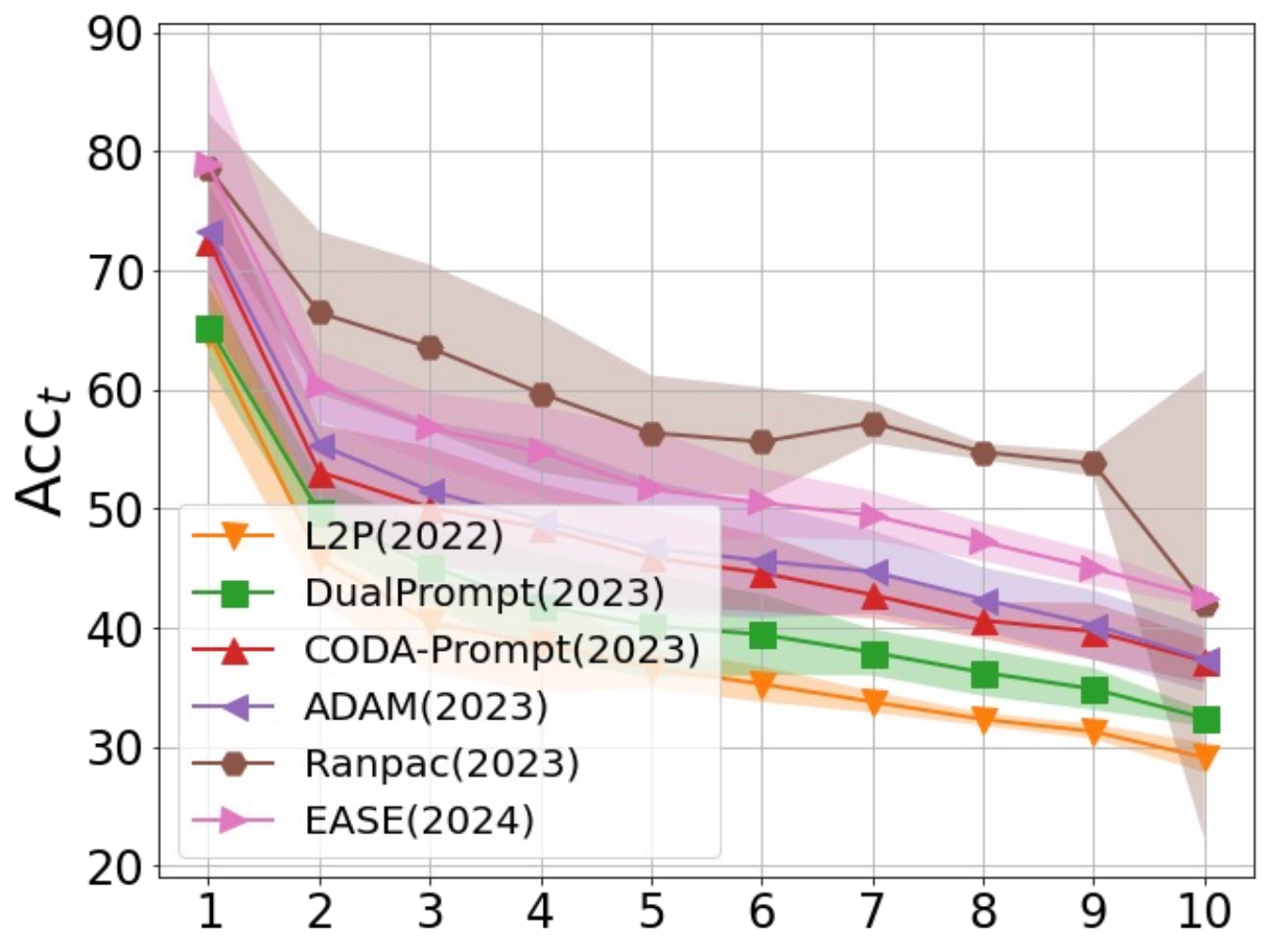}}
\vspace{-.1in}
\caption{Experimental results on the evaluation phase.}
\vspace{-.12in}
 \end{figure*}

 \newpage

\end{document}